\documentclass[preprint,12pt,letterpaper]{elsarticle}

\usepackage[ruled]{algorithm2e}
\usepackage{amsbsy}
\usepackage{amssymb}
\usepackage{amsthm}
\usepackage{amsmath}
\usepackage{bm}
\usepackage{graphicx}
\usepackage{multirow}
\usepackage{rotating}
\usepackage{subfigure}
\usepackage{url}
\usepackage{color}
\usepackage{natbib}
\usepackage{ifpdf}
\ifpdf
\usepackage{epstopdf} 
\fi
\usepackage[T1]{fontenc}

\usepackage[colorlinks=true, citecolor=black, filecolor=black, linkcolor=red, urlcolor=blue]{hyperref}

\usepackage[letterpaper]{geometry}
\renewcommand{\[}{\left[}
\renewcommand{\]}{\right]}
\renewcommand{\(}{\left(}
\renewcommand{\)}{\right)}

\newcommand{\order}[1]{{O}\(#1\)}

\newcommand{\ddt}[1]{\frac{d #1}{d t}}

\newcommand{\vvvert}{|\kern-1pt|\kern-1pt|}

\newcommand{\abs}[1]{\left|#1\right|}

\newcommand{\hU}{\hat{U}}

\newcommand{\barh}{\bar{h}}

\newcommand{\mb}[1]{\mathbf{#1}}

\newcommand{\bG}{\mathbf{G}}

\newcommand{\bd}{\mathbf{d}}

\newcommand{\bg}{\mathbf{g}}
\newcommand{\bk}{\mathbf{k}}
\newcommand{\bii}{\mathbf{i}}
\newcommand{\bj}{\mathbf{j}}

\newcommand{\by}{\mathbf{y}}

\newcommand{\bDelta}{\boldsymbol{\Delta}}

\newcommand{\bepsilon}{\boldsymbol{\epsilon}}

\newcommand{\btheta}{\boldsymbol{\theta}}

\newcommand{\bxi}{\boldsymbol{\xi}}
\newcommand{\bTheta}{\boldsymbol{\Theta}}
\newcommand{\bXi}{\boldsymbol{\Xi}}

\newcommand{\EE}{\mathbb{E}}

\newcommand{\NN}{\mathbb{N}}

\newcommand{\PP}{\mathbb{P}}
\newcommand{\RR}{\mathbb{R}}

\newcommand{\omegadot}{\dot{\omega}}

\newcommand{\CD}{\mathcal{D}}

\newcommand{\CF}{\mathcal{F}}

\newcommand{\CK}{\mathcal{K}}

\newcommand{\CN}{\mathcal{N}}

\newcommand{\CU}{\mathcal{U}}

\newcommand{\CY}{\mathcal{Y}}

\newcommand{\hhoo}{H$_2$O$_2$}
\newcommand{\hhoor}{H$_2$-O$_2$}

\newcommand{\etal}{\textit{et al.}}

\theoremstyle{definition}

\theoremstyle{definition}

\journal{Journal of Computational Physics}

\begin{document}

\begin{frontmatter}
  
   \title{Simulation-based optimal Bayesian experimental design for
    nonlinear systems}

 
  \author{Xun Huan}
  \ead{xunhuan@mit.edu}

  \author{Youssef M.~Marzouk}
  \ead{ymarz@mit.edu}

  \address{Massachusetts Institute of Technology, Cambridge, MA 02139, USA}

  \begin{abstract}
    
    The optimal selection of experimental conditions is essential to
    maximizing the value of data for inference and prediction,
    particularly in situations where experiments are time-consuming
    and expensive to conduct. We propose a general mathematical
    framework and an algorithmic approach for optimal experimental
    design with nonlinear simulation-based models; in particular, we
    focus on finding sets of experiments that provide the most
    information about targeted sets of parameters.

    Our framework employs a Bayesian statistical setting, which
    provides a foundation for inference from noisy, indirect, and
    incomplete data, and a natural mechanism for incorporating
    heterogeneous sources of information. An objective function is
    constructed from information theoretic measures, reflecting
    expected information gain from proposed combinations of
    experiments. Polynomial chaos approximations and a two-stage Monte
    Carlo sampling method are used to evaluate the expected
    information gain. Stochastic approximation algorithms are then
    used to make optimization feasible in computationally intensive
    and high-dimensional settings. These algorithms are demonstrated
    on model problems and on nonlinear parameter inference problems
    arising in detailed combustion kinetics.

  \end{abstract}

  \begin{keyword}
    
    Uncertainty quantification \sep Bayesian inference \sep Optimal
    experimental design \sep Nonlinear experimental design \sep
    Stochastic approximation \sep Shannon information \sep Chemical
    kinetics

  \end{keyword}
  
\end{frontmatter}

\section{Introduction}
\label{s:introduction}

%
%
%
%
%

Experimental data play an essential role in developing and refining
models of physical systems. For example, data may be used to update
knowledge of parameters in a model or to discriminate among competing
models. Whether obtained through field observations or laboratory
experiments, however, data may be difficult and expensive to
acquire. Even controlled experiments can be time-consuming or delicate
to perform. In this context, maximizing the value of experimental
data---designing experiments to be ``optimal'' by some appropriate
measure---can dramatically accelerate the modeling
process. Experimental design thus encompasses questions of where and
when to measure, which variables to interrogate, and what experimental
conditions to employ.

These questions have received much attention in the statistics
community and in many science and engineering applications. When
observables depend linearly on parameters of interest, common solution
criteria for the optimal experimental design problem are written as
functionals of the information matrix~\cite{atkinson:1992:oed}. These
criteria include the well-known `alphabetic optimality' conditions,
e.g., $A$-optimality to minimize the average variance of parameter
estimates, or $G$-optimality to minimize the maximum variance of model
predictions. Bayesian analogues of alphabetic optimality, reflecting
prior and posterior uncertainty in the model parameters, can be
derived from a decision-theoretic point of
view~\cite{chaloner:1995:bed}. For instance, Bayesian $D$-optimality
can be obtained from a utility function containing Shannon information
while Bayesian $A$-optimality may be derived from a squared error
loss. In the case of linear-Gaussian models, the criteria of Bayesian
alphabetic optimality reduce to mathematical forms that parallel their
non-Bayesian counterparts~\cite{chaloner:1995:bed}.

For nonlinear models, however, exact evaluation of optimal design
criteria is much more challenging.  More tractable design criteria can
be obtained by imposing additional assumptions, effectively changing
the form of the objective; these assumptions include linearizations of
the forward model, Gaussian approximations of the posterior
distribution, and additional assumptions on the marginal distribution
of the data~\cite{chaloner:1995:bed}. In the Bayesian setting, such
assumptions lead to design criteria that may be understood as
\textit{approximations of an expected utility}. Most of these involve
prior expectations of the Fisher information
matrix~\cite{chu:2008:ips}. Cruder ``locally optimal'' approximations
require selecting a ``best guess'' value of the unknown model
parameters and maximizing some functional of the Fisher information
evaluated at this point~\cite{ford:1989:ran}. None of these
approximations, though, is suitable when the parameter distribution is
broad or when it departs significantly from
normality~\cite{clyde:1993:bod}. A more general design framework, free
of these limiting assumptions, is preferred~\cite{muller:1998:sbo,
  guest:2009:ics}.


More rigorous information theoretic criteria have been proposed
throughout the literature. The seminal paper of
Lindley~\cite{lindley:1956:oam} suggests using expected gain in
Shannon information, from prior to posterior, as a measure of the
information provided by an experiment; the same objective can be
justified from a decision theoretic
perspective~\cite{lindley:1972:bsa,loredo:2010:rsa}. Sebastiani and
Wynn~\cite{sebastiani:2000:mes} propose selecting experiments for
which the marginal distribution of the data has maximum Shannon
entropy; this may be understood as a special case of Lindley's
criterion. Maximum entropy sampling (MES) has seen use in applications
ranging from astronomy~\cite{loredo:2003:bae} to
geophysics~\cite{vanDenBerg:2003:onb}, and is well suited to nonlinear
models. Reverting to Lindley's criterion, Ryan~\cite{ryan:2003:eei}
introduces a Monte Carlo estimator of expected information gain to
design experiments for a model of material fatigue. Terejanu
\etal~\cite{terejanu:2011:bed} use a kernel estimator of mutual
information (equivalent to expected information gain) to identify
parameters in chemical kinetic model. The latter two studies evaluate
their criteria on every element of a finite set of possible designs
(on the order of ten designs in these examples), and thus sidestep the
challenge of \textit{optimizing} the design criterion over general
design spaces. And both report significant limitations due to
computation expense; \cite{ryan:2003:eei}~concludes that ``full blown
search'' over the design space is infeasible, and that two
order-of-magnitude gains in computational efficiency would be required
even to discriminate among the enumerated designs.

The application of optimization methods to experimental design has
thus favored simpler design objectives. The chemical engineering
community, for example, has tended to use linearized and locally
optimal~\cite{mosbach:2011:iio} design criteria or other
objectives~\cite{russi:2008:sao} for which deterministic optimization
strategies are suitable.
But in the broader context of decision theoretic design formulations,
sampling is required. \cite{muller:1995:odv} proposes a curve fitting
scheme wherein the expected utility was fit with a regression model,
using Monte Carlo samples over the design space. This scheme relies on
problem-specific intuition about the character of the expected utility
surface.  Clyde \etal~\cite{clyde:1995:eeu} explore the joint design,
parameter, and data space with a Markov chain Monte Carlo (MCMC)
sampler; this strategy combines integration with optimization, such
that the marginal distribution of sampled designs is proportional to
the expected utility. This idea is extended with simulated annealing
in~\cite{muller:2004:obd} to achieve more efficient maximization of
the expected utility. \cite{clyde:1995:eeu,muller:2004:obd} use
expected utilities as design criteria but do not pursue information
theoretic design metrics. Indeed, direct optimization of information
theoretic metrics has seen much less development. Building on the
enumeration approaches of~\cite{vanDenBerg:2003:onb, ryan:2003:eei,
  terejanu:2011:bed} and the one-dimensional design space considered
in~\cite{loredo:2003:bae}, \cite{guest:2009:ics} iteratively finds MES
designs in multi-dimensional spaces by greedily choosing one component
of the design vector at a time. Hamada \etal~\cite{hamada:2001:fno}
also find ``near-optimal'' designs for linear and nonlinear regression
problems by maximizing expected information gain via genetic
algorithms. But the coupling of rigorous information theoretic design
criteria, complex physics-based models, and efficient optimization
strategies remains an open challenge.


This paper addresses exactly these issues. Our interest is in
physically realistic and hence \textit{computationally intensive}
models. We advance the state of the art by introducing flexible
approximation and optimization strategies that yield optimal
experimental designs for nonlinear systems, using a full information
theoretic formalism, efficiently and with few limiting assumptions.

In particular, we employ a Bayesian statistical approach and focus on
the case of parameter inference. Expected Shannon information gain is
taken as our design criterion; this objective naturally incorporates
prior information about the model parameters and accommodates very
general probabilistic relationships among the experimental
observables, model parameters, and design conditions. The need for
such generality is illustrated in the numerical examples
(Sections~\ref{s:applicationNonlinearModel}
and~\ref{s:applicationCombustionKinetics}).
To make evaluations of expected information gain computationally
tractable, we introduce a generalized polynomial chaos
surrogate~\cite{ghanem:1991:sfe,xiu:2002:twa} that captures smooth
dependence of the observables jointly on parameters and design
conditions. The surrogate carries no \textit{a priori} restrictions on
the degree of nonlinearity
and uses dimension-adaptive sparse quadrature~\cite{gerstner:2003:dat}
to identify and exploit anisotropic parameter and design dependencies
for efficiency in high dimensions. We link the surrogate with
stochastic approximation algorithms and use the resulting scheme to
maximize the design objective. This formulation allows us to plan
single experiments without discretizing the design space, and to
rigorously identify optimal ``batch'' designs of multiple experiments
over the product space of design conditions.

Figure~\ref{f:summary} shows the key components of our approach,
embedded in a flowchart describing a design--experimentation--model
improvement cycle. The upper boxes focus on experimental design: the
design criterion is formulated in Sections~\ref{s:experimentalGoals}
and \ref{s:expectedUtility}; estimation of the objective function is
described in Section~\ref{s:expectedUtilityNumericalMethods}; and
stochastic optimization approaches are described in
Section~\ref{s:stochasticOptimization}. The construction of polynomial
chaos surrogates for computationally intensive models is presented in
Section~\ref{s:polynomialChaos}. Section~\ref{s:BayesianParameterInference}
briefly reviews computational approaches for Bayesian parameter
inference, which come into play after the selected experiments have
been performed and data have been collected.
All of these tools are demonstrated on two example problems: a
simple nonlinear model in Section~\ref{s:applicationNonlinearModel}
and a shock tube autoignition experiment with detailed chemical
kinetics in Section~\ref{s:applicationCombustionKinetics}.

\section{Experimental design formulation}

\subsection{Experimental goals}
\label{s:experimentalGoals}

Optimal experimental design relies on the construction of a design
criterion, or objective function, that reflects how valuable or
relevant an experiment is expected to be. A fundamental consideration
in specifying this objective is the application of interest---i.e.,
what does the user intend to do with the results of the experiments?
For example, if one would like to estimate a particular physical
constant, then a good objective function might reflect the uncertainty
in the inferred values, or the error in a point estimate. On the other
hand, if one's ultimate goal is to make accurate model predictions,
then a more appropriate objective function should directly consider
the distribution of the model outputs conditioned on data. If one
would like to find the ``best'' model among a set of candidate models,
then the objective function should reflect how well the data are
expected to support each model, favoring experiments that maximize the
ability of the data to discriminate. These considerations motivate the
intuitive notion that an objective function should be based on
specific \textit{experimental goals}.

In this paper, we shall assume that the experimental goal is to infer
a finite number of model parameters of interest. Parameter inference
is of course an integral part of calibrating models from experimental
data~\cite{kennedy:2001:bco}. The expected utility framework developed
below can be generalized to other experimental goals, and we will
mention this where appropriate. Note that one could also augment the
objective function by adding a penalty that reflects experimental
effort or cost. More broadly, one can always add resource constraints
to the experimental design optimization problem. In the interest of
simplicity, and since costs and constraints are inevitably
problem-specific, we do not pursue such additions here.


\subsection{Design criterion and expected utility}
\label{s:expectedUtility}

We will formulate our experimental design criterion in a Bayesian
setting. Bayesian statistics offers a foundation for inference from
noisy, indirect, and incomplete data; a mechanism for incorporating
physical constraints and heterogeneous sources of information; and a
complete assessment of uncertainty in parameters, models, and
predictions. The Bayesian approach also provides natural links to
decision theory~\cite{berger:2010:sdt}, a framework we will exploit
below. (For discussions contrasting Bayesian and frequentist
approaches to statistics, see~\cite{gelman:2008:otb}
and~\cite{stark:2010:apo}.)


The Bayesian paradigm \cite{sivia:2006:daa} treats unknown parameters
as random variables. Let $(\Omega,\CF,\PP)$ be a probability space,
where $\Omega$ is a sample space, $\CF$ is a $\sigma$-field, and $\PP$
is a probability measure on $(\Omega, \CF)$.
Let the vector of real-valued random variables $\btheta: \Omega
\rightarrow \mathbb{R}^{n_{\theta}}$ denote the uncertain parameters
of interest, i.e., the parameters to be conditioned on experimental
data. $\btheta$ is associated with a measure $\mu$ on
$\mathbb{R}^{n_{\theta}}$, such that $\mu(A) = \PP\left
(\btheta^{-1}\left ( A \right ) \right ) $ for $A \in
\mathbb{R}^{n_{\theta}}$. We then define $p(\btheta) = d \mu / d
\btheta$ to be the density of $\btheta$ with respect to Lebesgue
measure.  For the present purposes, we will assume that such a density
always exists. Similarly we treat the data $\by$ as an
$\mathbb{R}^{n_y}$-valued random variable endowed with an appropriate
density. $\bd \in \mathbb{R}^{n_d}$ denotes the \textit{design
  variables} or experimental conditions. Hence $n_{\theta}$ is the
number of uncertain parameters, $n_y$ is the number of observations,
and $n_d$ is the number of design variables.
If one performs an experiment at conditions $\bd$ and observes a
realization of the data $\by$, then the change in one's state of
knowledge about the model parameters is given by Bayes' rule:
\begin{eqnarray}
  p(\btheta|\by,\bd) = \frac{p(\by|\btheta,\bd)p(\btheta|\bd)}
  {p(\by|\bd)}. \label{e:Bayes}
\end{eqnarray}
Here $p(\btheta|\bd)$ is the prior density, $p(\by|\btheta,\bd)$ is
the likelihood function, $p(\btheta|\by,\bd)$ is the posterior
density, and $p(\by|\bd)$ is the evidence. It is reasonable to assume
that prior knowledge on $\btheta$ does not vary with the experimental
design, leading to the simplification $p(\btheta|\bd) = p(\btheta)$.

Taking a decision theoretic approach, Lindley~\cite{lindley:1972:bsa}
suggests that an objective for experimental design should have the
following general form:
\begin{eqnarray}
  U(\bd) &=& \int_{\CY} \int_{\bTheta} u(\bd,\by,\btheta)
  \, p(\btheta,\by|\bd) \,d\btheta \, d\by \nonumber\\
  &=& \int_{\CY} \int_{\bTheta} u(\bd,\by,\btheta) \, p(\btheta
  |\by,\bd) \, p(\by|\bd) \, d\btheta \, d\by,
\end{eqnarray}
where $u(\bd,\by,\btheta)$ is a \textit{utility function}, $U(\bd)$ is
the \textit{expected} utility, $\bTheta$ is the support of
$p(\btheta)$, and $\CY$ is the support of $p(\by|\bd)$. The utility
function $u$ should be chosen to reflect the usefulness of an
experiment at conditions $\bd$, given a particular value of the
parameters $\btheta$ and a particular outcome $\by$. Since we do not
know the precise value of $\btheta$ and we cannot know the outcome of
the experiment before it is performed, we take the expectation of $u$
over the joint distribution of $\btheta$ and $\by$.

Our choice of utility function is rooted in information theory. In
particular, following~\cite{lindley:1956:oam}, we put
$u(\bd,\by,\btheta)$ equal to the relative entropy, or
Kullback-Leibler (KL) divergence, from the posterior to the prior. For
generic distributions $A$ and $B$, the KL divergence from $A$ to $B$ is
\begin{eqnarray}
  D_{\mathrm{KL}}( A || B) = \int_{\bTheta}
  p_A(\btheta)\ln\[\frac{p_A(\btheta)}{p_B(\btheta)}\]\,d\btheta =
  \mathbb{E}_A \left [ \ln \frac{p_A(\btheta)}{p_B(\btheta)} \right ]
\end{eqnarray}
where $p_A$ and $p_B$ are probability densities, $\bTheta$ is the
support of $p_B(\btheta)$, and $0\ln 0\equiv 0$. This quantity is
non-negative, non-symmetric, and reflects the difference in
information carried by the two distributions (in units of nats)
\cite{cover:2006:eoi,mackay:2006:eoi}. Specializing to the inference
problem at hand, the KL divergence from the posterior to the prior is
\begin{eqnarray}
  u(\bd,\by,\btheta) \equiv D_{\mathrm{KL}} \left ( p_{\btheta}(\cdot | \by, \bd)
    ||  p_{\btheta}( \cdot ) \right ) = 
  \int_{\bTheta}
  p(\tilde{\btheta}|\by,\bd)\ln\[\frac{p(\tilde{\btheta}|\by,\bd)} 
  {p(\tilde{\btheta})}\] \,d\tilde{\btheta} = u(\bd,\by).
  \label{e:utility}
\end{eqnarray}
Note that this choice of utility function involves an ``internal''
integration over the parameter space ($\tilde{\btheta}$ is a dummy
variable representing the parameters), therefore it is not a function
of the parameters $\btheta$. Thus we have
\begin{eqnarray}
  U(\bd) &=& \int_{\CY} \int_{\bTheta} u(\bd,\by)
  p(\btheta|\by,\bd)\,d\btheta \,p(\by|\bd) \,d\by \nonumber\\ & = & \int_{\CY}
  u(\bd,\by) \,p(\by|\bd) \,d\by \nonumber\\ &=& \int_{\CY}
  \int_{\bTheta} p(\tilde{\btheta}|\by,\bd)
  \ln\left[\frac{p(\tilde{\btheta}|\by,\bd)}{p(\tilde{\btheta})}\right]
  \,d\tilde{\btheta} \,p(\by|\bd) \,d\by .
    \label{e:expectedUtility1}
\end{eqnarray}
To simplify notation, $\tilde{\btheta}$ in
Equation~(\ref{e:expectedUtility1}) is replaced by $\btheta$, yielding
\begin{eqnarray}
  U(\bd) & = & \int_{\CY} \int_{\bTheta} p(\btheta|\by,\bd)
  \ln\left[\frac{p(\btheta|\by,\bd)}{p(\btheta)}\right] \,d\btheta
  \,p(\by|\bd) \,d\by \nonumber \\[8pt] & = & \mathbb{E}_{\by | \bd} \left
    [ D_{\mathrm{KL}} \left ( p ( \btheta |\by, \bd)
      ||  p ( \btheta ) \right )  \right ].
    \label{e:expectedUtility}
\end{eqnarray}
The expected utility $U$ is therefore the \textit{expected information
  gain} in $\btheta$. The intuition behind this expression is that a
large KL divergence from posterior to prior implies that the data
$\by$ decrease entropy in $\btheta$ by a large amount, and hence those
data are more informative for parameter inference. As we
have only a distribution for the data $\by | \bd$ that may be
observed, we are interested in maximizing information gain \textit{on
  average}. We also note that $U$ is equivalent to the \textit{mutual
  information} between the parameters $\btheta$ and the data
$\by$.\footnote{Using the definition of conditional probability, we
  have \begin{eqnarray*}U(\bd) &=& \int_{\CY} \int_{\bTheta}
    p(\btheta|\by,\bd)
    \ln\left[\frac{p(\btheta|\by,\bd)}{p(\btheta)}\right] \,d\btheta
    \,p(\by|\bd) \,d\by \\ &=& \int_{\CY} \int_{\bTheta}
    p(\btheta,\by|\bd)
    \ln\left[\frac{p(\btheta,\by|\bd)}{p(\btheta)p(\by|\bd)}\right]
    \,d\btheta \,d\by \\ &=& I(\btheta;\by|\bd),\end{eqnarray*} which
  is the mutual information between parameters and data, given the design.}  When
applied to a linear-Gaussian design problem, $U$ reduces to the
Bayesian $D$-optimality condition.\footnote{$D$-optimality maximizes
  the determinant of the information matrix in a linear design
  problem~\cite{atkinson:1992:oed}. Bayesian $D$-optimality, in a
  linear-Gaussian problem, maximizes the determinant of the sum of the
  information matrix and the prior covariance
  \cite{chaloner:1995:bed}.}

Finally, the expected utility must be maximized over the design space $\CD$ to find the optimal
experimental design
\begin{eqnarray}
  \bd^\ast = \mathrm{arg } \max_{\bd\in\CD}\, U(\bd).
  \label{e:optimization}
\end{eqnarray}

What if resources allow multiple (say $N > 1$) experiments to be
carried out, but time and setup constraints require them to be planned
(and perhaps performed) simultaneously? If $\bd^\ast$ is the optimal
design parameter for a single experiment, the best choice is not
necessarily to repeat the experiment at $\bd^\ast$ $N$ times; this
does not generally yield the optimal expected information gain from
all the experiments. (\ref{app:infoGain} shows that the expected
utility of two experiments is not, in general, equal to the sum of the
expected utilities of the individual
experiments. Section~\ref{s:applicationNonlinearModel} provides a
numerical example of this situation.) Instead, all of the experiments
should be incorporated into the likelihood function, where now
$\bd\in\RR^{N n_d}$, $\by\in\RR^{N n_y}$, and the data from the
different experiments are conditionally independent given $\btheta$
and the augmented $\bd$. The new optimal design $\bd^\ast \in\RR^{N
  n_d}$ then carries the $N$ sets of conditions for all the
experiments, maximizing the expected total information gain when these
experiments are simultaneously performed. It is interesting to note
that a simpler objective function often used in experimental
design---the predictive variance of the data $\by$---would always
suggest repeating all $N$ experiments at the single-experiment design
optimum.

If the $N$ experiments need not be carried out simultaneously, then
\textit{sequential} experimental design may be performed. In general,
a sequential design uses the results of one set of experiments (i.e.,
the $\by$ that are actually observed) to help plan the next set of
experiments. In one possible approach---in fact, a \textit{greedy}
approach---an optimal experiment is initially computed and carried
out, and its data are used to perform inference. The resulting
posterior $p(\btheta,\by|\bd_1)$ is then used as the prior in the
design of the next experiment $\bd_2$, and the process is
repeated. This approach is not necessarily optimal over a horizon of
many experiments, however. A more rigorous treatment would involve
formulating the sequential design problem as a dynamic programming
problem, but this is beyond the scope of the present
paper. Intuitively, a sequential experimental design should be at
least as good as a fixed design, due to the extra information gained
in the intermediate stages.

\subsection{Numerical evaluation of the expected utility}
\label{s:expectedUtilityNumericalMethods}


Typically, the expected utility in Equation~(\ref{e:expectedUtility})
has no closed form and must be approximated numerically. One approach
is to rewrite $U(\bd)$ as
\begin{eqnarray}
  U(\bd) &=& \int_{\CY} \int_{\bTheta}
  p(\btheta|\by,\bd)
  \ln\left[\frac{p(\btheta|\by,\bd)}{p(\btheta)}\right]
  \,d\btheta \,p(\by|\bd) \, d\by \nonumber\\
  &=& \int_{\CY} \int_{\bTheta}
  \ln\left[\frac{p(\by|\btheta,\bd)}{p(\by|\bd)}
  \right]p(\by|\btheta,\bd) \, 
  p(\btheta)\,d\btheta \,d\by \nonumber\\
  &=& \int_{\CY} \int_{\bTheta}
  \left\{\ln\[p(\by|\btheta,\bd)\]-\ln\[p(\by|\bd)\]\right\}
  p(\by|\btheta,\bd) \, p(\btheta)\,d\btheta \,d\by,
  \label{e:EUAlternative}
\end{eqnarray}
where the second equality is due to the application of Bayes' theorem
to the quantities both inside and outside the logarithm. 
In the special case where the Shannon entropy of $p(\by, \btheta | \bd)$
is independent of the design variables $\bd$, the first term in
Equation~(\ref{e:EUAlternative}) becomes constant for all designs \cite{shewry:1987:mes} and
can be dropped from the objective function. Maximizing the remaining
term---which is the entropy of $p(\by|\bd)$---is then equivalent to the
maximum entropy sampling approach of Sebastiani and
Wynn~\cite{sebastiani:2000:mes}. Here we retain the more general
formulation of Equation~(\ref{e:EUAlternative}) in order to
accommodate, for example, likelihood functions containing a
measurement error whose magnitude depends on $\by$ or $\bd$.

Monte Carlo sampling can then be used to estimate the integral in
Equation~(\ref{e:EUAlternative})
\begin{eqnarray}
  U(\bd) \approx \frac{1}{n_{\mathrm{out}}} \sum_{i=1}^{n_{\mathrm{out}}}
  \left\{\ln\[p(\by^{(i)}|\btheta^{(i)},\bd)\] 
    -\ln\[p(\by^{(i)}|\bd)\]\right\},
  \label{e:expectedUtilityMC}
\end{eqnarray}
where $\btheta^{(i)}$ are drawn from the prior $p(\btheta)$;
$\by^{(i)}$ are drawn from the conditional distribution
$p(\by|\btheta=\btheta^{(i)}, \bd )$ (i.e., the likelihood); and
$n_{\mathrm{out}}$ is the number of samples in this ``outer'' Monte
Carlo estimate. The evidence evaluated at $\by^{(i)}$,
$p(\by^{(i)}|\bd)$, typically does not have an analytical form, but it
can be approximated using yet another importance sampling estimate:
\begin{eqnarray}
  p(\by^{(i)}|\bd) =
  \int_{\bTheta}p(\by^{(i)}|\btheta,\bd)p(\btheta)
  \,d\btheta \approx  \frac{1}{n_{\mathrm{in}}} \sum_{j=1}^{n_{\mathrm{in}}}
  p(\by^{(i)}|\btheta^{(i,j)},\bd),
  \label{e:evidenceMC}
\end{eqnarray}
where $\btheta^{(\cdot,j)}$ are drawn from the prior $p(\btheta)$ and
$n_{\mathrm{in}}$ is the number of samples in this ``inner'' Monte
Carlo sum. The combination of Equations~(\ref{e:expectedUtilityMC}) and
(\ref{e:evidenceMC}) yields a biased estimator $\hat{U}(\bd)$ of
$U(\bd)$~\cite{ryan:2003:eei}. The variance of this estimator is
proportional to
${A(\bd)}/{n_{\mathrm{out}}}+{B(\bd)}/{n_{\mathrm{out}}n_{\mathrm{in}}}$,
where $A$ and $B$ are terms that depend only on the distributions at
hand. The bias is proportional to
${C(\bd)}/{n_{\mathrm{in}}}$~\cite{ryan:2003:eei}. Hence
$n_{\mathrm{in}}$ controls the bias while $n_{\mathrm{out}}$ controls
the variance.

Evaluating and sampling from the likelihood for each new sample of
$\btheta$ constitutes the most significant computational cost above
(see Section~\ref{s:polynomialChaos}). In order to mitigate the cost
of the nested Monte Carlo estimator, we draw a fresh batch of prior
samples $\{\btheta^{\(k\)}\}_{k=1}^{n_{\mathrm{out}}}$ for every
$\bd$, and use this set for both the outer Monte Carlo sum (i.e.,
$\btheta^{\(i\)}=\btheta^{\(k\)}$) and all the inner Monte Carlo
estimates at that $\bd$ (i.e.,
$\btheta^{\(\cdot,j\)}=\btheta^{\(k\)}$, and consequently
$n_{\mathrm{out}} = n_{\mathrm{in}}$). This treatment reduces the
computational cost for a fixed $\bd$ from
$\order{n_{\mathrm{out}}n_{\mathrm{in}}}$ to
$\order{n_{\mathrm{out}}}$. In practice, sample reuse also avoids
producing near-zero evidence estimates (and hence infinite values for
the expected utility) at small sample sizes. The reuse of samples
contributes to the bias of the estimator, but this effect is very
small~\cite{huan:2010:abe}. See \ref{app:reuseBias} for a numerical
study of the bias.

\subsection{Stochastic optimization}
\label{s:stochasticOptimization}

Now that the expected utility $U(\bd)$ can be estimated at any value
of the design variables, we turn to the optimization problem
(\ref{e:optimization}). Maximizing $U$ via a grid search over $\CD$ is
clearly impractical, since the number of grid points grows
exponentially with dimension. Since only a Monte Carlo estimate
$\hat{U}(\bd)$ of the objective function is available, another
na\"{i}ve approach would be to use a large sample size $\left (
  n_{\mathrm{out}}, n_{\mathrm{in}} \right )$ at each $\bd$ and then
apply a deterministic optimization algorithm, but this is still too
expensive. (And even with large sample sizes, $\hat{U}(\bd)$ is
effectively non-smooth.) Instead, we would like to use only a
\textit{few} Monte Carlo samples to evaluate the objective at any
given $\bd$, and thus we need algorithms suited to noisy objective
functions. Two such algorithms are simultaneous perturbation
stochastic approximation (SPSA) and Nelder-Mead nonlinear simplex
(NMNS).

SPSA, proposed by Spall~\cite{spall:1998:aoo,spall:1998:iot}, is a
stochastic approximation method that has received considerable
attention~\cite{spall:2008:sps}. The method is similar to a
steepest-descent method using finite difference estimates of the
gradient, except that SPSA only uses two random perturbations to
estimate the gradient regardless of the problem's dimension:
\begin{eqnarray}
  \bd_{k+1} &=& \bd_k-a_k\bg_k(\bd_k) \label{e:step}\\
  \bg_k(\bd_k) &=&
  \frac{\hat{U}(\bd_k+c_k\bDelta_k)-\hat{U}(\bd_k-c_k\bDelta_k)}{2c_k} 
  \[ \begin{array}{c} \Delta_{k,1}^{-1} \\ \Delta_{k,2}^{-1} \\ \vdots
    \\ \Delta_{k,n_d}^{-1} \end{array} \],
  \label{e:gradient}
\end{eqnarray}
where $k$ is the iteration number,
\begin{eqnarray}
  a_k = \frac{a}{\( A+k+1 \)^{\alpha}}, & &
  c_k = \frac{c}{\( k+1 \)^{\gamma}},
\end{eqnarray}
and $a$, $A$, $\alpha$, $c$, and $\gamma$ are algorithm parameters
with recommended values available, e.g.,
in~\cite{spall:1998:iot}. $\bDelta_k$ is a random vector whose entries
are i.i.d.\ draws from a symmetric distribution with finite inverse
moments~\cite{spall:1998:aoo}; here, we choose $\Delta_{k,i} \sim
\operatorname{Bernoulli}\(0.5\)$. Common random numbers are also used
to evaluate each pair of estimates $\hat{U}(\bd_k+c_k\bDelta_k)$ and
$\hat{U}(\bd_k-c_k\bDelta_k)$ at a given $\bd_k$, in order to reduce
variance in estimating the gradient~\cite{kleinman:1999:sbo}.

An intuitive justification for SPSA is that error in the gradient
``averages out'' over a large number of
iterations~\cite{spall:1998:aoo}. Convergence proofs with varying
conditions and assumptions can be found in~\cite{spall:1988:asa,
  spall:1992:msa, he:2003:csp}. Randomness introduced through the
noisy objective $\hat{U}$ and the finite-difference-like perturbations
allows for a global convergence property~\cite{maryak:2004:gro}.
Constraints in SPSA are handled by projection: if the current position
does not remain feasible under \textit{all} possible random
perturbations, then it is projected to the nearest point that does
satisfy this condition.

The NMNS algorithm~\cite{nelder:1965:asm} has been well studied and is
widely used for deterministic optimization. The details of the
algorithm are thus omitted from this discussion but can be found,
e.g., in~\cite{nelder:1965:asm, barton:1996:nms, spall:2003:its}.
This algorithm has a natural advantage in dealing with noisy objective
functions because it requires only a \textit{relative ordering} of
function values, rather than the magnitudes of differences (as in
estimating gradients). Minor modifications to the algorithm parameters
can improve optimization performance for noisy
functions~\cite{barton:1996:nms}. Constraints in NMNS are handled
simply by projecting from the infeasible point to the nearest feasible
point.

There are advantages and disadvantages to both algorithms. SPSA is a
gradient-based approach, taking advantage of any regularity in the
underlying objective function while requiring only two function
evaluations per step to estimate the gradient instead of $2n_d$
evaluations, as with a full finite-difference scheme. However, a
very high noise level can lead to slow convergence and cause the
algorithm to stagnate in local optima. NMNS is relatively less
sensitive to the noise level, but the simplex can be unfavorably
distorted due to the projection treatment of constraints, leading to
slow or false convergence.

Using either of the algorithms described in this section, we can
approximately solve the stochastic optimization problem posed in
Equation~(\ref{e:optimization}) and obtain the best experimental
design. In a sense, this completes the experimental design phase of
Figure~\ref{f:summary}. But a remaining difficulty is one of
computational cost. Even with an effective Monte Carlo estimator of
the expected utility, and with efficient algorithms for stochastic
optimization, the complex physical model embedded in
Equation~(\ref{e:expectedUtilityMC}) still must be evaluated repeatedly,
over many values of the model parameters and design variables. Methods
for making this task more tractable are discussed in the next section.

\section{Polynomial chaos surrogate}
\label{s:polynomialChaos}

Expensive physical models can render the evaluation and maximization
of expected information gain impractical. Models enter the formulation
through the likelihood function $p(\by|\btheta,\bd)$. For example, a
simple likelihood function might allow for an additive discrepancy
between experimental observations and model predictions:
\begin{eqnarray}
\by = \bG\left ( \btheta, \bd \right ) + \bepsilon.
\end{eqnarray}
Here, $\bepsilon$ is a random variable with density $p_{\bepsilon}$;
we leave the form of this density non-specific for now. The ``forward
model'' of the experiment is $\bG: \RR^{n_{\theta}} \times \RR^{n_{d}}
\longrightarrow \RR^{n_y}$; it maps both the design variables and the
parameters into the data space. Drawing a realization from
$p(\by|\btheta,\bd)$ thus requires evaluating $\bG$ at a particular
$(\btheta, \bd)$. Evaluating the density $p(\by|\btheta,\bd) =
p_{\bepsilon} \left (\by - \bG ( \btheta, \bd ) \right )$ again requires
evaluating $\bG$.

To make these calculations tractable, one would like to replace $\bG$
with a cheaper ``surrogate'' model that is accurate over the entire
prior support $\bTheta$ and the entire design space $\CD$. Many
options exist, ranging from projection-based model
reduction~\cite{buithanh:2007:mrl,frangos:2010:srm} to spectral
methods based on polynomial chaos (PC)
expansions~\cite{wiener:1938:thc,ghanem:1991:sfe, xiu:2002:twa,
  debusschere:2004:nci, najm:2009:uqa, xiu:2009:fnm,
  lemaitre:2010:smf}. The latter approaches do not reduce the internal
state of a deterministic model; rather, they explicitly exploit
regularity in the dependence of model outputs on uncertain input
parameters. Polynomial chaos has seen extensive use in a range of
engineering applications (e.g.,~\cite{hosder:2006:ani,
  reagan:2003:uqi, walters:2003:tsf, xiu:2003:ans}) including
parameter estimation and inverse problems
(e.g.,~\cite{marzouk:2007:ssm,marzouk:2009:asc,marzouk:2009:dra}), and
this is the approach we shall use.

Let $\xi_i: \Omega \rightarrow \mathbb{R}$ be i.i.d.\ random variables
defined on a probability space $(\Omega,\CF,\PP)$, where $\Omega$ is
the sample space, $\CF$ is the $\sigma$-field generated by all the
$\xi_i$, and $\PP$ is the probability measure. Then any random
variable $\theta:\Omega\rightarrow \mathbb{R}$, measurable with
respect to $(\Omega,\CF)$ and possessing finite variance, $\theta\in
L^2(\Omega,\PP)$, can be represented as follows:
\begin{eqnarray}
  \theta(\omega) = 
  \sum_{|\bii|=0}^{\infty} \theta_{\bii}
  \Psi_{\bii}(\xi_1(\omega),\xi_2(\omega),\ldots),
  \label{e:PCEForm2}
\end{eqnarray}
where $\omega\in\Omega$ is an element of the sample space;
$\bii=\(i_1,i_2,\ldots\),\,i_j\in\NN_0$, is an
infinite-dimensional multi-index; $|\bii|=i_1+i_2+\ldots$ is the $l_1$
norm; $\theta_{\bii} \in \RR$ are the expansion coefficients; and
\begin{eqnarray}
  \Psi_{\bii}(\xi_1,\xi_2,\ldots) = \prod_{j=1}^{\infty}
  \psi_{i_j}(\xi_j)
\end{eqnarray}
are multivariate polynomial basis functions~\cite{xiu:2002:twa}. Here
$\psi_{i_j}$ is an orthogonal polynomial of order $i_j$ in the
variable $\xi_j$, where orthogonality is with respect to the
distribution of $\xi_j$,
\begin{eqnarray}
  \EE_{\xi}\[\psi_m\psi_n\] = \int_{\Xi}
  \psi_m(\xi)\psi_n(\xi)p(\xi)\,d\xi =
  \delta_{m,n}\EE_{\xi}\[\psi_m^2\],
  \label{e:orthogonality}
\end{eqnarray}
and $\Xi$ is the support of $p(\xi)$. The
expansion~(\ref{e:PCEForm2}) is convergent in the mean-square
sense~\cite{cameron:1947:tod}. For computational purposes, the
infinite sum and infinite dimension must be truncated to some finite
stochastic dimension $n_s$ and polynomial order. A common choice is
the ``total-order'' truncation $|\bii|\leq p$:
\begin{eqnarray}
  \theta(\omega) &\approx& 
  \sum_{|\bii|\leq p} \theta_{\mb{i}}
 \Psi_{\bii}(\xi_1,\xi_2,\ldots,\xi_{n_s}) 
  \label{e:truncatedPC} \\
  \Psi_{\bii}(\xi_1,\xi_2,\ldots,\xi_{n_s}) &=& \prod_{j=1}^{n_s}
  \psi_{i_j}(\xi_j).
  \label{e:truncatedPsi}
\end{eqnarray}
The total number of terms in this expansion is
\begin{eqnarray}
  n_{\mathrm{PC}} = \(\begin{array}{c}n_s+p \\ p \end{array} \) =
  \frac{(n_s+p)!}{n_s!p!}.
  \label{e:npc}
\end{eqnarray}
The choice of $p$ is influenced by the degree of nonlinearity in the
relationship between $\theta$ and $\xi_j$, and the choice of $n_s$
reflects the degrees of freedom needed to capture the stochasticity of
the system. These choices might also be constrained by the
availability of computational resources, as $n_{\mathrm{PC}}$ grows quickly
when these numerical parameters are increased.

\subsection{Joint expansion for design variables}
\label{ss:polynomialChaosDesignVariables}

In the Bayesian setting, the model parameters $\btheta$ are random
variables, for which PC expansions are easily applied. But the model
outputs also depend on the design conditions, and constructing a
separate PC expansion at each value of $\bd$ required during
optimization would be impractical. Instead, we can construct a
\textit{single} PC expansion for each component of $\bG$, depending
jointly on $\btheta$ and $\bd$. (Similar suggestions have recently
appeared in the context of robust design~\cite{eldred:2011:duu}.) To
proceed, we increase the stochastic dimension by the number of design
dimensions, putting $n_s = n_\theta + n_d$, where we have assigned one
stochastic dimension to each component of $\btheta$ and one to each
component of $\bd$ for simplicity. Further, we assume an affine
transformation between each component of $\bd$ and the corresponding
$\{ \xi_i \}_{i=n_\theta + 1}^{n_s}$; any value of $\bd$ can thus be
uniquely associated with a vector of these $\xi_i$. Since the design
parameters will usually be supported on a bounded domain (e.g., inside
some hyper-rectangle) the corresponding $\xi_i$ are given uniform
distributions. (The corresponding univariate $\psi_i$ are thus
Legendre polynomials.)  These distributions effectively define a
uniform weight function over the design space $\CD$ that governs where
the $L^2$-convergent PC expansions should be accurate.


\subsection{Pseudospectral projection}

Constructing the PC expansion involves computing the coefficients
$\theta_{\bii}$; this generally can proceed via two alternative
approaches, intrusive and non-intrusive. The intrusive approach
results in a new system of equations that is larger than the original
deterministic system, but it needs be solved only once. The difficulty
of this latter step depends strongly on the character of the original
equations, however, and may be prohibitive for arbitrary nonlinear
systems. The non-intrusive approach computes the expansion
coefficients by directly projecting the quantity of interest (e.g.,
the model output) onto the basis functions $\{ \Psi_{\bii} \}$. One
advantage of this method is that the deterministic solver can be
reused and treated as a black box. The deterministic problem needs to
be solved many times, but typically at carefully chosen parameter
values. The non-intrusive approach also offers flexibility in choosing
arbitrary functionals of the state trajectory as observables; these
functionals may depend smoothly on $\bxi$ even when the state itself
has a less regular dependence. (The combustion model in
Section~\ref{s:applicationCombustionKinetics} provides an example of
such a situation.)

Taking advantage of orthogonality, the PC coefficients are simply:
\begin{eqnarray}
  G_{c,\bii} = \frac{\EE_{\bxi}\[G_c \Psi_{\bii}\]}{\EE_{\bxi}\[\Psi_{\bii}^2 \]} =
  \frac{\int_{\bXi} G_c(\btheta(\bxi),\bd(\bxi)) \Psi_{\bii}(\bxi) p(\bxi) \,d\bxi}{\int_{\bXi}
    \Psi_{\bii}^2(\bxi) p(\bxi) \,d\bxi},\,c=1\ldots n_y,
  \label{e:NISP}
\end{eqnarray}
where $G_{c,\bii}$ is the PC coefficient with multi-index $\bii$ for
the $c$th observable.\footnote{Here we are equating the dimension of
  the forward model output with the number of observables $n_y$. If
  the data contain repeated observations of the same quantity, for
  instance, in the case of multiple experiments, then the same PC
  approximation can be used for all model-based predictions of that
  quantity.} Analytical expressions are available for the denominators
$\EE_{\bxi}\[\Psi_{\bii}^2\]$, but the numerators must be evaluated
via numerical quadrature, because of the forward model $\bG$. The
resulting approach is termed pseudospectral projection, or
non-intrusive spectral projection (NISP). When the evaluations of the
integrand (and hence the forward model) are expensive and $n_s$ is
large, an efficient method for high-dimensional integration is
essential.

\subsection{Dimension-adaptive sparse quadrature}
\label{ss:PCNumericalIntegration}

A host of useful methods are available for numerical
integration~\cite{morokoff:1995:qmc, sobol:1967:otd,smolyak:1963:qai,
  barthelmann:2000:hdp,gerstner:1998:niu, bungartz:2004:hoq}.
In the present context, we seek a method that can evaluate the
numerator of Equation~(\ref{e:NISP}) efficiently in high dimensions,
i.e., with a minimal number of integrand evaluations, taking advantage
of regularity and anisotropy in the dependence of $\bG$ on $\btheta$
and $\bd$. We thus employ the dimension-adaptive sparse quadrature
(DASQ) algorithm of Gerstner and Griebel~\cite{gerstner:2003:dat}, an
efficient extension of Smolyak sparse quadrature that
\textit{adaptively} tensorizes quadrature rules in each coordinate
direction. It has a weak dependence on dimension, making it an
excellent candidate for problems of moderate size (e.g., $n_s<100$).
Its formulation is briefly described below.

Let
\begin{eqnarray}
  Q_l^{(1)}f = \sum_{i=1}^{n_l}w_{i}f(\bxi_{i})
\end{eqnarray}
be the $l$th level (with $n_l$ quadrature points) of some univariate
quadrature rule, where $w_i$ are the weights, $\bxi_i$ are the
abscissae, and $f(\bxi)$ is the integrand.
The level is usually defined to take advantage of any nestedness in
the quadrature rule and to reduce the overall computational cost. We
have chosen to use Clenshaw-Curtis (CC) quadrature for $\bxi$'s with
compact support, with the following level definition
\begin{eqnarray}
  n_l = 2^{l-1}+1,\,l\geq 2, &&  n_1 = 1.
  \label{e:CCLevels}
\end{eqnarray}
The CC rule is especially appealing because it is
accurate,\footnote{Although Gauss-Legendre quadrature (which is not
  nested) has a higher degree of polynomial exactness,
  \cite{trefethen:2008:igq} notes that
\begin{quote}
  \textit{``the Clenshaw-Curtis and Gauss formulas have essentially
    the same accuracy unless $f$ is analytic in a sizable neighborhood
    of the interval of the integration---in which case both methods
    converge so fast that the difference hardly
    matters.''}
\end{quote}} nested, and easy to construct.\footnote{The abscissae are
simply $x_i = \cos\(\frac{i\pi}{n}\)$, and the weights can be computed
very efficiently via
FFT~\cite{gentleman:1972:icc1,gentleman:1972:icc2}, requiring only
$\order{n\log n}$ time and introducing very little roundoff error.}

The difference formulas, defined by
\begin{eqnarray}
  \Delta_kf = (Q_k^{(1)}-Q_{k-1}^{(1)})f,\qquad
  Q_0^{(1)}f = 0,
\end{eqnarray}
are the differences between 1D quadrature rules at two consecutive
levels. The subtraction is carried out by subtracting the weights at
the quadrature points of the lower level.  Then, for $\bk\in \NN_1^d$
(where each entry of the multi-index $\bk$ represents the level in
that dimension, with a total of $d$ dimensions), the multivariate
quadrature rule is defined to be
\begin{eqnarray}
  Qf = \sum_{\bk \in \CK} (\Delta_{k_1}\otimes
  \cdots \otimes \Delta_{k_d})f, \label{e:smolyak}
\end{eqnarray}
where $\CK$ is some set determined by the adaptation algorithm, to be
described below. For example, $\CK:|\bk|_1 \leq L+d-1$ where $L$ is
some user-defined level, corresponds to the Smolyak sparse quadrature,
while $\CK:|\bk|_{\infty} \leq N$ corresponds to a tensor-product
quadrature.

The original DASQ algorithm can be found
in~\cite{gerstner:2003:dat}. The idea is to divide all the
multi-indices $\bk$ into two sets: an old set and an active
set. A member of the active set is able to propose a new candidates by
increasing the level in any dimension by 1.
However, the candidate can only be accepted if all its backward
neighbors
are in the old set; this so-called admissibility condition ensures the
validity of the telescoping expansion of the general sparse
quadrature formulas via the differences $\Delta_{k}$. Finally, each
multi-index has an error indicator, which is proportional to its
corresponding summand value in Equation~(\ref{e:smolyak}). Intuitively,
if this term contributes little to the overall integral estimate, then
integration error due to this term should be small. New candidates are
proposed from the multi-index corresponding to the highest error
estimate. The process iterates until the sum of error indicators for
the active set members falls below some user-specified tolerance. More
details, including proposed data structures for this algorithm, can be
found in~\cite{gerstner:2003:dat}. One drawback of DASQ is that
parallelization can only be implemented within the evaluation of each
$\bk$, which is not as efficient as the parallelization in
non-adaptive methods. The original DASQ algorithm also does not
address how integrand evaluations at nested quadrature points can
easily be identified and reused as adaptation
proceeds. Huan~\cite{huan:2010:abe} proposes an algorithm to solve
this problem, taking advantage of the specific quadrature
structure.

The ultimate goal of quadrature is to compute the polynomial chaos
coefficients of the model outputs in Equation~(\ref{e:NISP}). There are
a total of $n_{\mathrm{PC}}$ (see Equation~(\ref{e:npc}))
coefficients for each output variable, and a total of $n_y$ model
outputs, yielding a total of $n_{\mathrm{coef}}=n_{\mathrm{PC}}n_y$
integrals. To simplify notation, let the PC coefficients
$G_{c,\bii},\,c=1 \ldots n_y, \,|\bii| \leq p$, be re-indexed by
%
%
$G_{r},\,r=1 \ldots n_{\mathrm{coef}}$. It would be very inefficient
to compute each integral from scratch, since the corresponding
quadrature points will surely overlap and any evaluations of
$\bG(\btheta(\bxi),\bd(\bxi))$ ought to be reused. To realize these
computational savings, the original DASQ algorithm is altered to
integrate for all the coefficients $G_{r}$ simultaneously. We guide
all the integrations via a single adaptation route, which uses a
``total effect'' local error indicator $\barh_{\bk}$ that reflects all
the local error indicators $h_{r,\bk}$ from the integrals. The total
effect indicator at a given $\bk$ may be defined as the max or 2-norm
of the local error indicators $\{ h_{r,\bk}
\}_{r=1}^{n_{\mathrm{coef}}}$. The new algorithm is presented as
Algorithm~\ref{a:DASQinNISP}.

Lastly, compensated summation (the Kahan
algorithm~\cite{kahan:1965:fro}) is used throughout our
implementation, as it significantly reduces numerical error when
summing long sequences of finite-precision floating point numbers as
required above.



\section{Bayesian parameter inference}
\label{s:BayesianParameterInference}

Once data are collected by performing an optimal experiment, they can
be used in the manner specified by the original experimental goal. In
the present case, the goal is to infer the model parameters $\btheta$
by exploring or characterizing the posterior distribution in
Equation~(\ref{e:Bayes}). Ideally the data will lead to a narrow
posterior such that, with high probability, the parameters can only
take on small range of values.

The posterior can be evaluated pointwise up to a constant factor, but
computing it on a grid is immediately impractical as the number of
dimensions increases. A more economical method is to generate
independent samples from the posterior, but given the arbitrary form
of this distribution (particularly for nonlinear $\bG$), direct Monte
Carlo sampling is seldom possible.
Instead, one must resort to Markov chain Monte Carlo (MCMC) sampling.
Using only pointwise evaluations of the \textit{unnormalized}
posterior density, MCMC constructs a Markov chain whose stationary and limiting
distribution is the posterior. Samples generated in this way are
correlated, such that the effective sample size is smaller than
the number of MCMC steps. Nonetheless, a well-tuned MCMC algorithm can
be reasonably efficient. The resulting samples can then be used in
various ways---to evaluate marginal posterior densities, for instance,
or to approximate posterior expectations
\begin{eqnarray}
  \EE_{\btheta|\by,\bd}\[f(\btheta)\] =
  \int_{\bTheta}f(\btheta)p(\btheta|\by,\bd) \,d\btheta
\end{eqnarray}
with the $n_M$-sample average
\begin{eqnarray}
  \bar{f}_{n_M} = \frac{1}{n_{M}} \sum_{t=1}^{n_{M}} f(\btheta^{(t)}),
\end{eqnarray}
where $\btheta^{(t)}$ are samples extracted from the chain (perhaps
after burn-in or thinning). For example, the minimum mean square error
(MMSE) estimator is simply the mean of the posterior, while the
corresponding Bayes risk is the posterior variance, both of which can
be estimated using MCMC.

A very simple and powerful MCMC method is the Metropolis-Hastings (MH)
algorithm, first proposed by Metropolis
\etal~\cite{metropolis:1953:eos}, and later generalized by
Hastings~\cite{hastings:1970:mcs}; details of the algorithm can be
found
in~\cite{tierney:1994:mcf,gilks:1996:mcm,andrieu:2003:ait,robert:2004:mcs}. Two
useful improvements to MH are the concepts of delayed rejection
(DR)~\cite{green:2001:dri,mira:2001:omh} and adaptive Metropolis
(AM)~\cite{haario:2001:aam}; combining these lead to the DRAM
algorithm of Haario \etal~\cite{haario:2006:dra}. While countless
other MCMC algorithms exist or are under active development, some
involving derivatives (e.g., Langevin) or even Hessians of the
posterior density, DRAM offers a good balance of simplicity and
efficiency in the present context.


Even with efficient proposals, MCMC typically requires a tremendous
number of samples (tens of thousands or even millions) to compute
posterior estimates with acceptable accuracy. Since each MCMC step
requires evaluation of the posterior density, which in turn requires
evaluation of the likelihood and thus the forward model $\bG$,
surrogate models for the dependence of $\bG$ on $\btheta$ can offer
tremendous computational savings. Polynomial chaos surrogates, as
described in Section~\ref{s:polynomialChaos}, can be quite helpful in
this
context~\cite{marzouk:2007:ssm,marzouk:2009:asc,marzouk:2009:dra}.

\section{Application: nonlinear model}
\label{s:applicationNonlinearModel}

We first illustrate the optimal design of experiments using a simple
algebraic model, nonlinear in both the parameters and the design
variables. Since the model is inexpensive to evaluate, we use it to
illustrate features of the core formulation---estimating expected
information gain, designing single and multiple experiments, and the
role of prior information---leaving demonstrations of stochastic
optimization and polynomial chaos surrogates to the next section.

\subsection{Design of a single experiment}
\label{s:applicationNonlinearModelSingle}

Consider a simple nonlinear model with a scalar observable $y$, one
uncertain parameter $\theta$, and one design variable $d$:
\begin{eqnarray}
  y(\theta,d) &=& G(\theta,d) + \epsilon \nonumber\\ 
  &=& \theta^3 d^2 + \theta \exp(-|0.2-d|) + \epsilon, 
  \label{e:simpleDesign}
\end{eqnarray}
where $G(\theta,d)$ denotes the model output (without noise) and
$\epsilon\sim\CN(0,10^{-4})$ is an additive Gaussian measurement
error. Let the prior be $\theta\sim\mathcal{U}(0,1)$ and the design
space be $d\in[0,1]$.

Suppose our \textit{experimental goal} is to infer the uncertain
parameter $\theta$ based on a single measurement $y$. The expected
utility $U(d)$ in Equation~(\ref{e:expectedUtility}) and its estimate
$\hat{U}(d)$ in Equation~(\ref{e:expectedUtilityMC}) are appropriate
choices, and our ultimate goal is to maximize
$U(d)$. Figure~\ref{f:simpleDesign1Exp_theta0to1} shows estimates of
the expected utility, using $n_{\mathrm{out}}=n_{\mathrm{in}}=10^{5}$,
plotted along a $101$-node uniform grid spanning the entire design
space.  Local maxima appear at $d=0.2$ and $d=1.0$, a pattern which
can be understood by examining Equation~(\ref{e:simpleDesign}). A $d$
value away from 0.2 or 1.0 (such as $d=0$) would lead to an
observation $y$ that is dominated by the noise $\epsilon$, which is
not useful for inferring the uncertain parameter $\theta$. But if $d$
is chosen close to 0.2 or 1.0, such that the noise is insignificant
compared to the first or second term of the equation, then $y$ would
be very informative for $\theta$.

\subsection{Design of two experiments}
\label{s:nonlinearDesignOfTwoExperiments}

Consider the ``batch'' or fixed design of \textit{two} experiments
(where the results of one experiment cannot be used to design the other,
as described in Section~\ref{s:expectedUtility}). Moreover, assume that
both experiments are described by the same model; this is not a
requirement, but an assumption adopted here for simplicity. Then the
overall algebraic model is simply extended to
\begin{eqnarray}
  \by(\theta,\bd) &=& \bG(\theta,\bd) + \bepsilon \nonumber\\[6pt]
  \[\begin{array}{c}y_1(\theta,d_1) \\
    y_2(\theta,d_2) \end{array}\]
  &=& \left[\begin{array}{c} \theta^3
      d_1^2 + \theta
      \exp(-|0.2-d_1|) \\ \theta^3 d_2^2 + \theta
      \exp(-|0.2-d_2|) \end{array}\] +
    \left[\begin{array}{c} \epsilon_1 \\ \epsilon_2 \end{array}\], 
  \label{e:simpleDesign2}
\end{eqnarray}
where the subscripts $\cdot_1$ and $\cdot_2$ denote variables
associated with experiments 1 and 2, respectively. Note that there is
still a single common parameter $\theta$. The errors $\epsilon_1$ and
$\epsilon_2$ are i.i.d.\ with $\epsilon_1, \epsilon_2 \sim \CN(0,10^{-4})$.

\begin{figure}[htb]
  \centering
  \subfigure[Prior $\theta\sim \CU(0,1)$]
  {
    \includegraphics[width=0.47\textwidth]{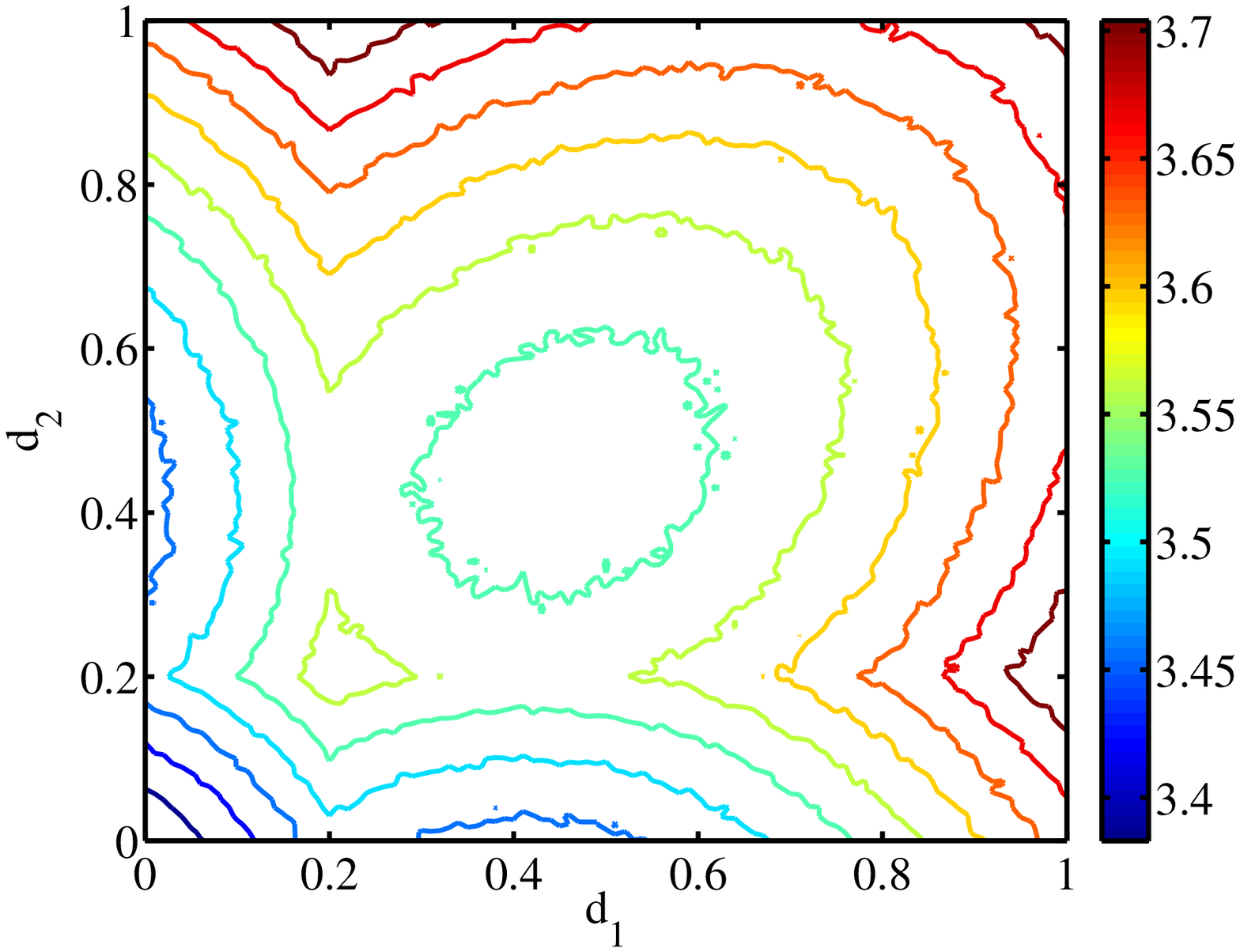}
    \label{f:simpleDesign2Exp_theta0to1}
  }\\
  \subfigure[Prior $\theta\sim \CU(0,\theta_e)$]
  {
    \includegraphics[width=0.47\textwidth]{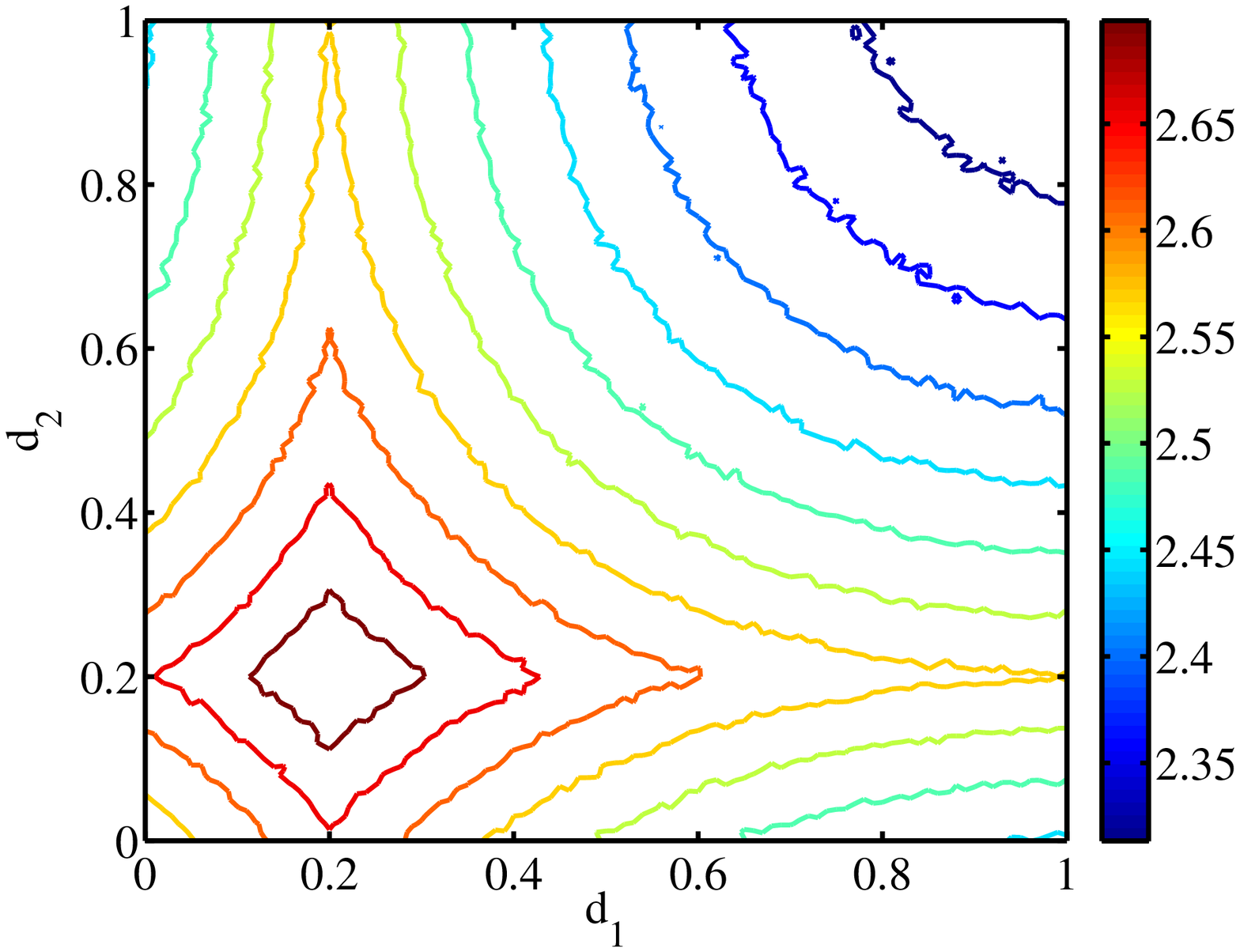}
    \label{f:simpleDesign2Exp_theta0to0_4373}
  }
  \subfigure[Prior $\theta\sim \CU(\theta_e,1)$]
  {
    \includegraphics[width=0.47\textwidth]{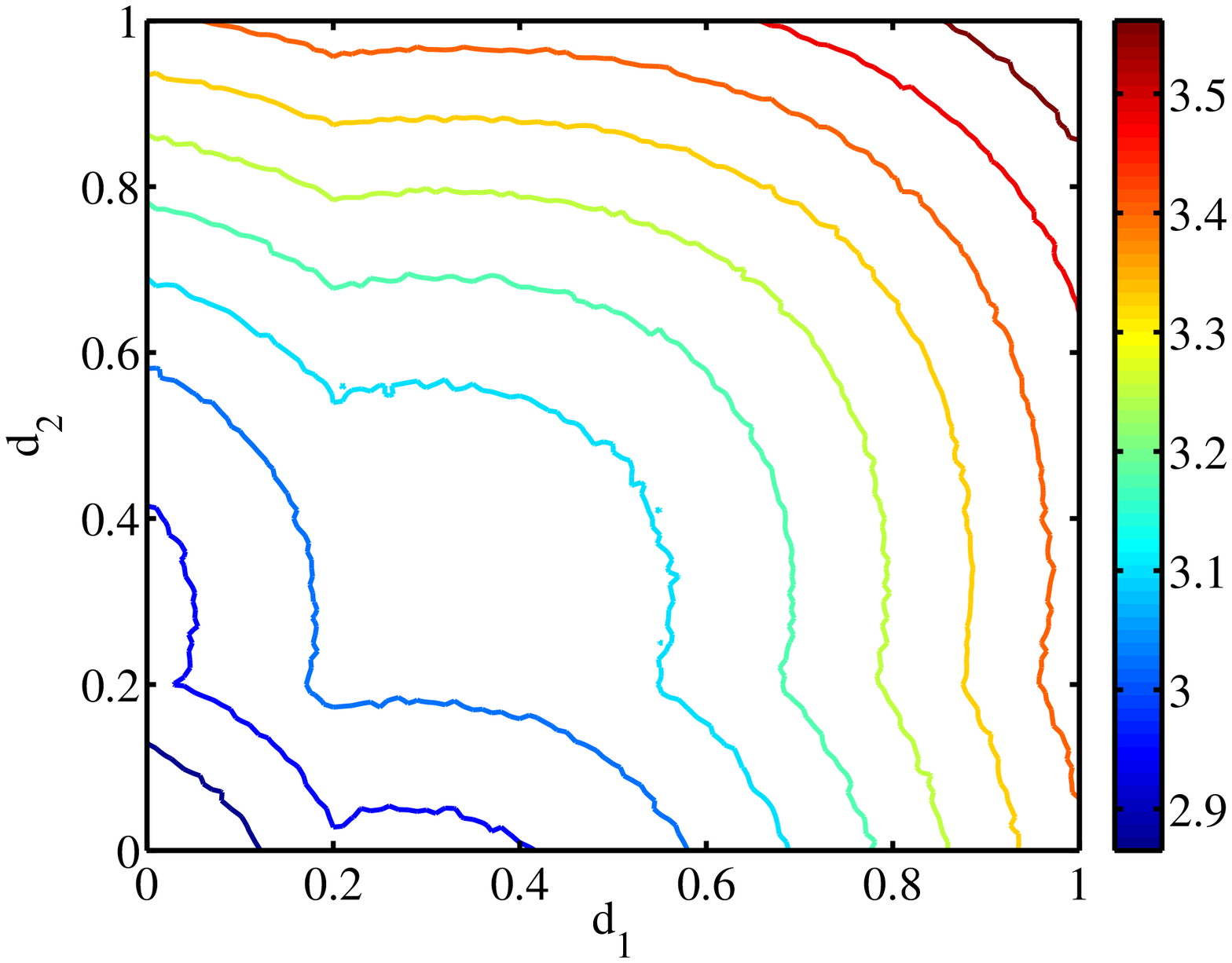}
    \label{f:simpleDesign2Exp_theta0_4373to1}
  }
  \caption{Estimated expected utility for the design of \textbf{two}
    experiments with the simple nonlinear model, under different
    priors, where $\theta_e=\sqrt{\frac{1-e^{-0.8}}{2.88}} \approx
    0.4373$.}
  \label{f:simpleDesign2Exp}
\end{figure}

\begin{figure}[htb]
  \centering
  \includegraphics[width=0.47\textwidth]{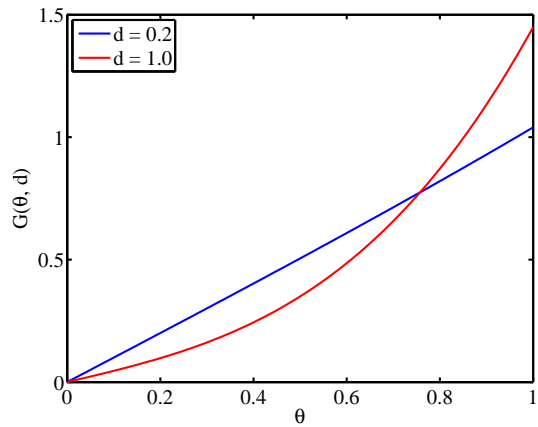}
  \caption{Noiseless output of the simple nonlinear model
    $G(\theta,d)$ as a function of the uncertain parameter $\theta$,
    at two designs: $d=0.2$ and $d=1$.}
  \label{f:simpleDesign_modelOutputAt_d0_2_and_d1}
\end{figure}

Again using a $\CU(0,1)$ prior on $\theta$, the expected utility is
plotted in Figure~\ref{f:simpleDesign2Exp_theta0to1}. First, note the
symmetry in the contours along the $d_1=d_2$ line, which is expected
since the two experiments have identical structure. Second, the
optimal pair of experiments is \textit{not} just a repeat of the
optimal single-experiment design: $\bd^\ast \neq (1,1)$, and instead
we have $\bd^\ast = (0.2,1.0)$ or $(1.0,0.2)$. Some insight can be
obtained by examining
Figure~\ref{f:simpleDesign_modelOutputAt_d0_2_and_d1}, which plots the
single-experiment model output $G(\theta,d)$ as a function of the
uncertain parameter $\theta$ at the two locally optimal designs:
$d=0.2$ and $d=1$. Intuitively, a high slope of $G$ should be more
informative for the inference of $\theta$, as the output is then more
sensitive to variations in the input. The plot shows that neither
design has a greater slope over the entire range of the prior $\theta
\sim \CU(0, 1)$. Instead, the slope is greater for $\theta\in
[0,\theta_e]$ with design $d=0.2$, and greater for $\theta\in
[\theta_e, 1.0]$ with design $d=1.0$, where
$\theta_e=\sqrt{\frac{1-e^{-0.8}}{2.88}} \approx 0.4373$.

Let us then examine the cases of ``restricted'' priors $\theta \sim
\CU(0, \theta_e)$ and $\theta \sim \CU(\theta_e, 1)$. Expected
utilities for a single experiment, under either of these priors, are
shown in Figures~\ref{f:simpleDesign1Exp_theta0to0_4373}
and~\ref{f:simpleDesign1Exp_theta0_4373to1}. The optimal design for
$\CU(0,\theta_e)$ is at 0.2 and for $\CU(\theta_e,1)$ it is at 1.0,
supporting intuition from the analysis of slopes. Next, the expected
utilities of two experiments, under the restricted priors, are shown
in Figures~\ref{f:simpleDesign2Exp_theta0to0_4373}
and~\ref{f:simpleDesign2Exp_theta0_4373to1}. Since in both cases, a
single design point can give $G$ maximum slope over the entire
\textit{restricted} prior range of $\theta$, it is not surprising that
the optimal \textit{pair} of experiments involves repeating the respective
single-experiment optima. In contrast, the lack of a ``clear winner''
over the \textit{full} prior $\CU(0,1)$ intuitively explains why a
combination of different design conditions may yield more informative
data overall. Note that we have only focused on the two local optima
$d=0.2$ and $d=1$ from the original $\theta\sim \CU(0,1)$ analysis,
but it is possible that new local or globally optimal design points
could emerge as the prior is changed.



\section{Application: combustion kinetics}
\label{s:applicationCombustionKinetics}

Experimental diagnostics play an essential role in the development and
refinement of chemical kinetic models for
combustion~\cite{frenklach:2007:tdk,davidson:2004:ist}. Available
diagnostics are often indirect, imprecise, and incomplete, leaving
significant uncertainty in relevant rate parameters and thermophysical
properties~\cite{baulch:1994:ekd1,baulch:2005:ekd2,reagan:2003:uqi,phenix:1998:ipu}. Uncertainties
are particularly acute when developing kinetic models for new
combustion regimes or for fuels derived from new feedstocks, such as
biofuels. Questions of experimental design---e.g., which species to
interrogate and under what conditions---are thus of great practical
importance in this context.

\subsection{Model description}

We demonstrate our optimal experimental design framework on shock tube
ignition experiments, which are a canonical source of kinetic
data. In a shock tube experiment, the mixture behind the reflected
shock experiences a sharp rise in tempeature and pressure; if
conditions are suitable, this mixture then ignites after some time,
known as the ignition delay time. Ignition delays and other
observables extracted from the experiment carry indirect information
about the elementary chemical kinetic processes occurring in the
mixture. These experiments are well described by the dynamics of a
spatially homogeneous, adiabatic, constant-pressure chemical mixture.

We model the evolution of the mixture using ordinary differential
equations (ODEs) expressing conservation of energy and of individual
chemical species. Governing equations are detailed in
\ref{app:combustionGovEqns}. We consider an initial mixture of
hydrogen and oxygen. (Note that \hhoor\ kinetics are a key subset of
the reaction mechanisms associated with the combustion of complex
hydrocarbons.) Our baseline kinetic model is a 19-reaction mechanism
proposed in~\cite{yetter:1991:acr}, reproduced in
Table~\ref{t:hoMech}. Detailed chemical kinetics lead to a stiff set
of nonlinear ODEs, with state variables consisting of temperature and
species mass or molar fractions. The initial condition of the system
is specified by the initial temperature $T_0$ and the fuel-oxidizer
equivalence ratio $\phi$. Species production rates depend on the
mixture conditions and on a set of kinetic parameters: pre-exponential
factors $A_{m}$, temperature exponents $b_{m}$, and activation energies
$E_{a,m}$, where $m$ is the reaction number in
Table~\ref{t:hoMech}. These parameters are important in determining
combustion characteristics and are of great interest in
practice. Thermodynamic parameters and reaction rates in the governing
equations are evaluated with the help of Cantera
1.7.0~\cite{goodwin:2002:ccu,cantera1.7.0:2009:c1w}, an open-source
chemical kinetics software package. ODEs are solved implicitly, using
the variable-order backwards differentiation formulas implemented in
CVODE~\cite{cohen:1996:cas}.

\subsection{Experimental goals}
\label{s:combustionExperimentalGoals}

In this study, the experimental goal is to infer selected kinetic
parameters ($A_{m}$, $b_{m}$, and $E_{a,m}$) associated with the
elementary reactions in Table~\ref{t:hoMech}. For demonstration, we
let the kinetic parameters of interest be $A_1$ and
$E_{a,3}$. Reaction 1 is a chain-branching reaction, leading to a net
increase in the number of radical species in the system. Reaction 3 is
a chain-propagating reaction, exchanging one radical for another, but
nonetheless relevant to the overall dynamics.\footnote{As the
  methodology explored here is quite general, we have the freedom to
  select any parameters appearing in the mechanism. The selection
  reflects the particular goals of the experimentalist or investigator.
  We also note that the ``evaluated'' combustion kinetic data
  in~\cite{baulch:1994:ekd1,baulch:2005:ekd2} can help select
  parameters to target for inference \textit{and} help define their
  prior ranges. }
We infer $\ln(A_1/A_1^0)$ rather than $A_1$ directly, where $A_1^0$ is
the nominal value of $A_1$ in \cite{yetter:1991:acr}; this
transformation ensures positivity and lets us easily impose a
log-uniform prior on $A_1$, which is appropriate since the
pre-exponential is a multiplicative factor. The design variables are
the initial temperature $T_0$ and equivalence ratio $\phi$.

We once again use the expected utility defined in
Equation~(\ref{e:expectedUtility}) (and its estimator in
Equation~(\ref{e:expectedUtilityMC})) as the objective to be maximized
for optimal design. Unlike the algebraic model in
Section~\ref{s:applicationNonlinearModel}, however, this combustion
problem offers many possible choices of observable. Some observables
are more informative than others; we explore this choice in the next
section.

\subsection{Observables and likelihood function}
\label{s:combustionSelectionOfObservables}

\begin{figure}[htb]
  \centering
  \subfigure[Temperature]{
    \includegraphics[width=0.47\textwidth]{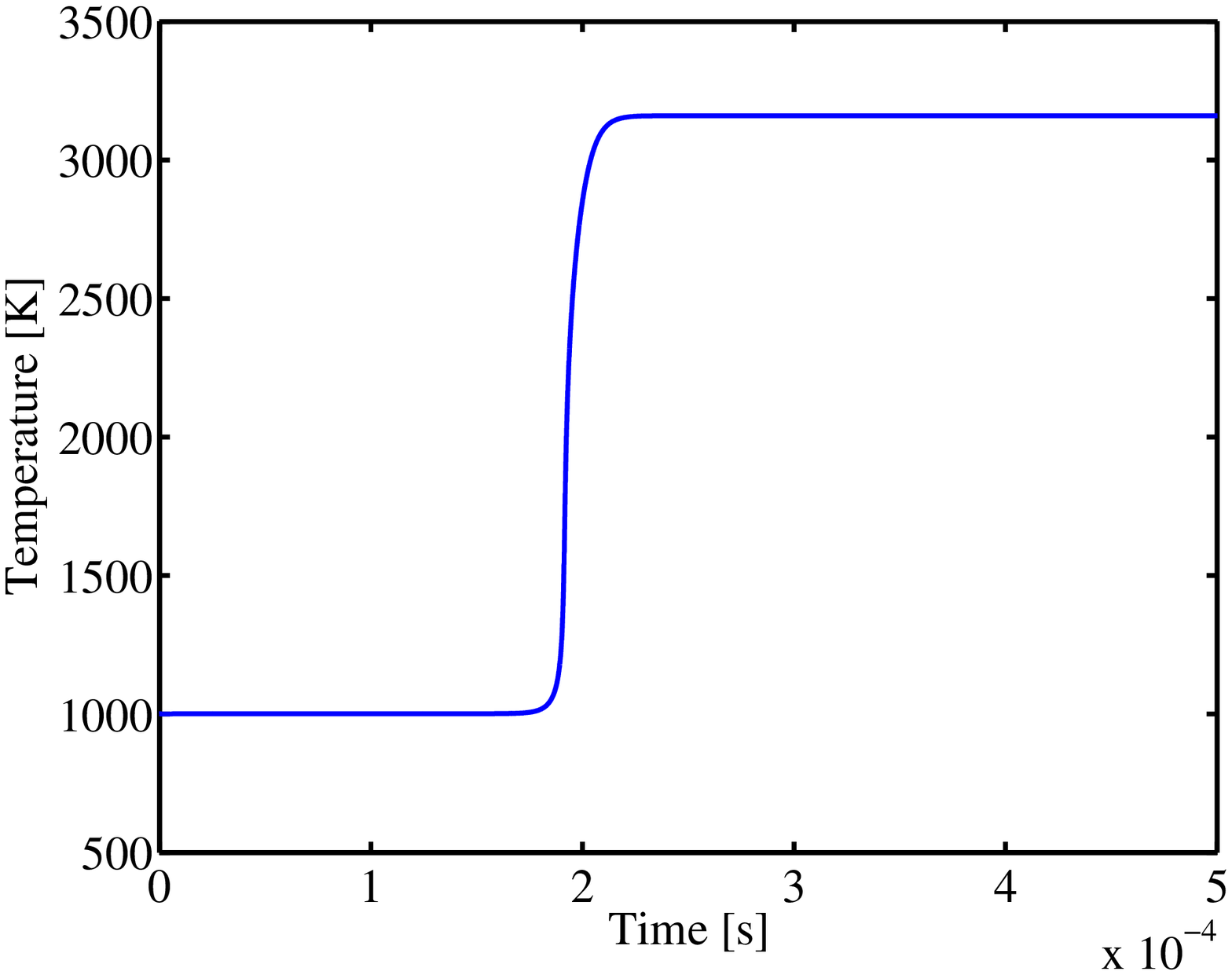}
  }
  \subfigure[Species molar fractions]{
    \includegraphics[width=0.47\textwidth]{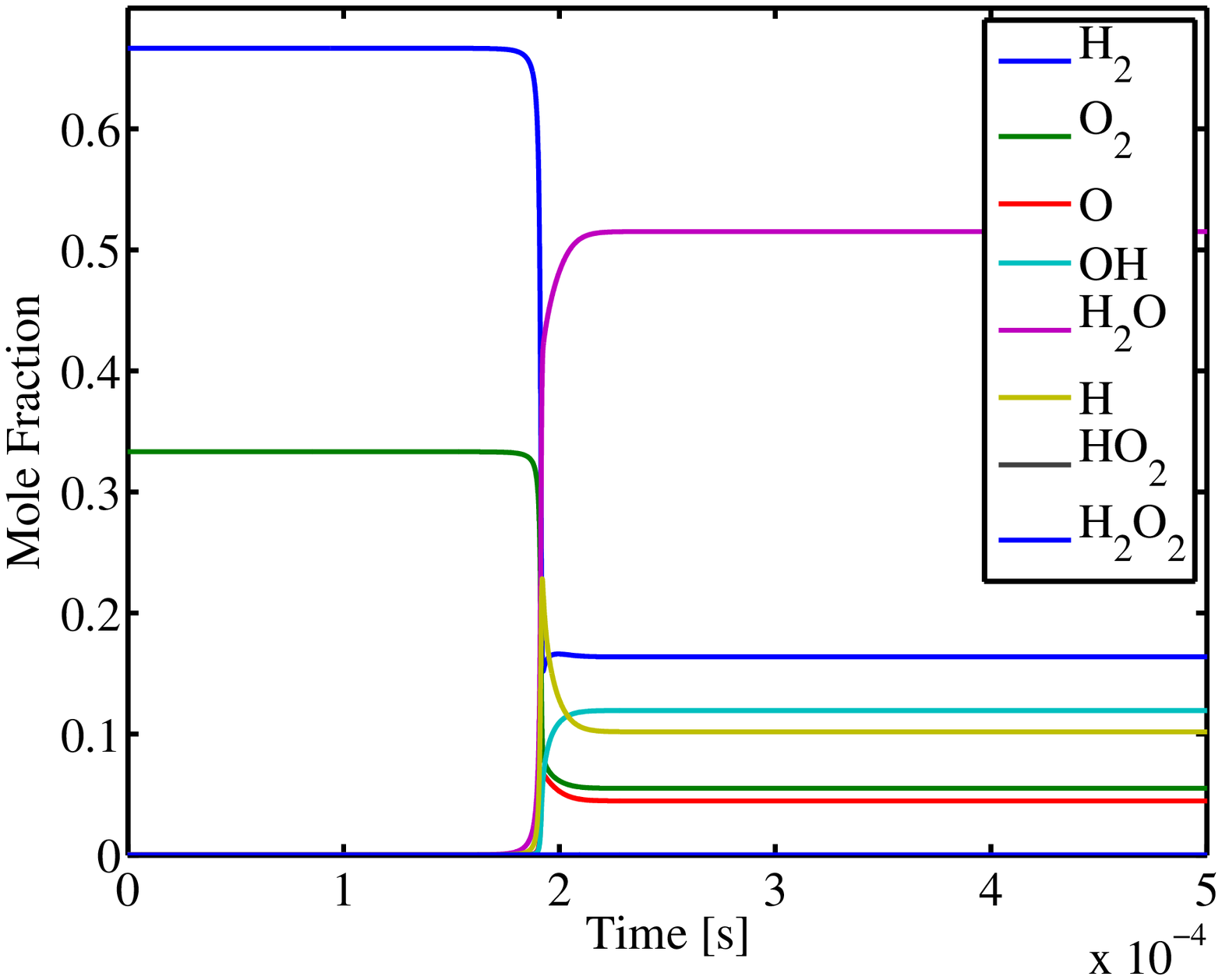}
  }
  \caption{Typical time-evolution of temperature and species molar
    fractions in \hhoor\, ignition.}
  \label{f:typicalComb}
\end{figure}

Typical trajectories of the state variables are shown in
Figure~\ref{f:typicalComb}. The temperature rises suddenly upon
ignition; reactant species are rapidly consumed and product species
are produced as the mixture comes to equilibrium. 
%
%
%
The most complete and detailed set of system observables are the state
variables as a function of time. One could simply discretize the time
domain to produce a finite-dimensional data vector $\by$. Too few
discretization points might fail to capture the state behaviour,
however. And because the kinetic parameters affect ignition delay, the
state at any given time may have a nearly discontinuous dependence on
the parameters. (This is due to the sharpness of the ignition front;
at a fixed time, the state is most probably either pre-ignition or
post-ignition.) Such a dependence makes construction of a polynomial
chaos surrogate far more challenging~\cite{najm:2009:uqc}. It is
desirable to transform the state into alternate observables that
somehow ``compress'' the information and depend relatively smoothly on
the kinetic parameters, while retaining features that are relevant to
the experimental goals. We would also like to select observables that
are easy to obtain experimentally.

\begin{figure}[htb]
  \centering
  \subfigure[$\tau_{ign}$]{
    \includegraphics[width=0.47\textwidth]{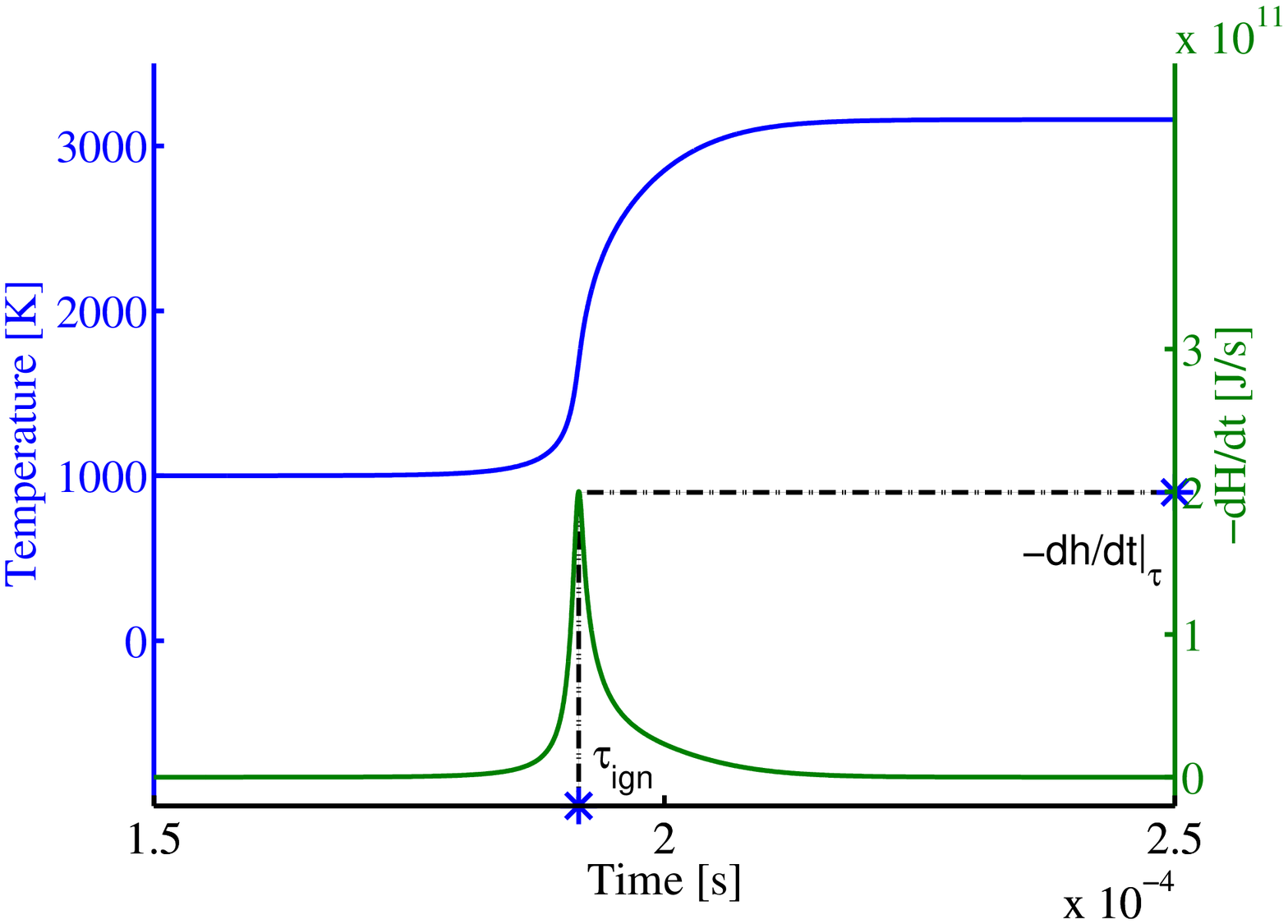}
  }
  \subfigure[$\tau_{H}$ and $X_{H,\tau}$]{
    \includegraphics[width=0.47\textwidth]{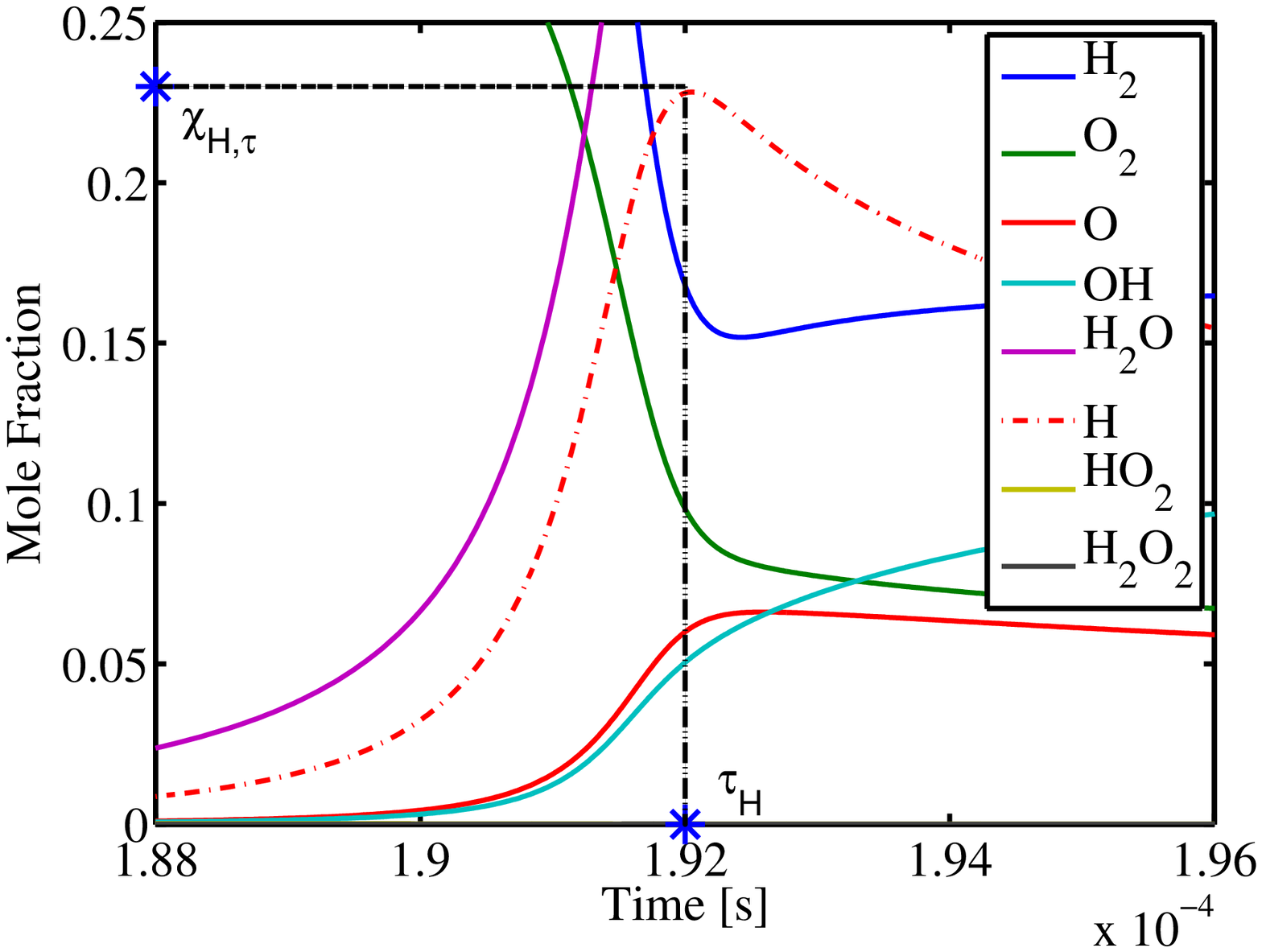}
  }
  \caption{Illustration of the observables $\tau_{ign}$, $\tau_{H}$, and $X_{H,\tau}$
    in the combustion problem.}
  \label{f:sampleObs}
\end{figure}

Taking the above factors into consideration, we will use the
observables in Table~\ref{t:obs}. The observables are the \textit{peak
  value} of the heat release rate, the \textit{peak concentrations} of
various intermediate chemical species (O, H, HO$_2$, \hhoo), and the
\textit{times} at which these peak values occur.
%
Examples of $\tau_{ign}$, $\tau_{H}$,
$\left.\frac{dh}{dt}\right|_{\tau}$, and $X_{H,\tau}$ are shown in
Figure~\ref{f:sampleObs}. The time of peak heat release coincides with
the time at which temperature rises most rapidly. We thus take it as
our definition of ignition delay, $\tau_{ign}$. We use the logarithm
of all the characteristic time variables in our actual implementation,
as the times are positive and vary over several orders of magnitude as
a function of the kinetic parameters and design variables.

The likelihood is defined using the ODE model predictions and
independent additive Gaussian measurement errors: $\by = \bG(\btheta,
\bd) + \bepsilon$, with components $\epsilon_c \sim \CN(0,
\sigma^2_c)$. For the concentration observables, the standard
deviation of the measurement error is taken to be 10\% of the value of
the corresponding signal:
\begin{eqnarray}
  \sigma^X_{c} = 0.1 G_c(\btheta,\bd).
\end{eqnarray}
For the characteristic-time observables, we add a small constant
$\alpha = 10^{-5}$ s to the standard deviation, reflecting the minimum
resolution of the timing technology:
\begin{eqnarray}
  \sigma^{\tau}_c = 0.1 G_c(\btheta,\bd) + \alpha.
\end{eqnarray}
Note that the noise magnitude depends implicitly on both the kinetic
parameters and the design variables. Both terms contributing to the
expected information gain in Equation~(\ref{e:EUAlternative}) are
therefore influential, and one would expect a maximum entropy sampling
approach to yield different results than the present experimental
design methodology.



\subsection{Polynomial chaos construction}

Each solve of the ODE system defining $\bG(\btheta, \bd)$ is
expensive, and thus we employ a polynomial chaos surrogate. In
practice, since non-intrusive construction of the surrogate requires
many forward model evaluations, the surrogate is only worth forming if
the total number of model evaluations required for optimization of the
expected utility exceeds the number required for surrogate
construction. A detailed analysis of this tradeoff and the potential
computational gains can be found in Section~\ref{ss:PCWorthIt}.

Uniform priors are assigned to the model parameters $\btheta \equiv
\left [ \ln(A_1/A_1^0), E_{a,3} \right ]$ and uniform input
distributions are assumed for the design variables $ \left [ T_0, \phi
\right ]$ (see Section~\ref{ss:polynomialChaosDesignVariables}), with
the supports given in Table~\ref{t:priorSupport}. The polynomial chaos
expansions thus use Legendre polynomials, with
$\xi_1,\xi_2,\xi_3,\xi_4 \sim \CU(-1,1)$. Our goal now is to construct
PC expansions for the model outputs $\bG(\btheta,\bd) =
(\ln\tau_{ign},\ln\tau_{O},\ln\tau_{H},\ln\tau_{HO_2},\ln\tau_{H_2O_2},$
$\left.\ddt{h}\right|_{\tau},X_{O,\tau},X_{H,\tau},X_{HO_2,\tau},X_{H_2O_2,\tau})$,
using the projection given in Equation~(\ref{e:NISP}). For each desired
PC coefficient, the numerator in that equation is evaluated using the
modified DASQ algorithm described in
Section~\ref{ss:PCNumericalIntegration}. The expansions are truncated
at a total order of $p=12$, and DASQ is stopped once a total of
$n_{\mathrm{quad}}=10^6$ function evaluations have been exceeded. The
degree of this expansion is admittedly (and deliberately) chosen
rather high. The performance of lower-order expansions is examined
below and in Section~\ref{ss:PCWorthIt}.
%
%
%

Indeed, the accuracy of the PC surrogate can be analyzed more
rigorously by evaluating its \textit{relative} $L^2$ \textit{error}
over a range of $p$ and $n_{\mathrm{quad}}$ values:
\begin{eqnarray}
  e_c = \frac{\int_{\bXi}
    \abs{G_c(\btheta(\bxi),\bd(\bxi))-G_c^{p,n_{\mathrm{quad}}}(\bxi)}^2p(\bxi)\,d\bxi}
  {\int_{\bXi}
    \abs{G_c(\btheta(\bxi),\bd(\bxi))}^2p(\bxi)\,d\bxi},\qquad c=1
  \ldots n_{y}.
  \label{e:L2}
\end{eqnarray}
For the $c$th observable, $G_c$ is the output of the original ODE
model and $G_c^{p,n_{\mathrm{quad}}}$ is the corresponding PC
surrogate. $\btheta$ and $\bd$ are affine functions of $\bxi$ (the PC
expansions of the model inputs). Accurately evaluating the $L^2$ error
is expensive, certainly more expensive than computing
$G_c^{p,n_{\mathrm{quad}}}$ in the first place. But additional
integration error must be minimized, and the integrals in
Equation~(\ref{e:L2}) are thus evaluated using a level-15 isotropic
Clenshaw-Curtis sparse quadrature rule, containing 3,502,081 distinct
abscissae.

Figure~\ref{f:L2ErrorsExample} shows contours of $\log_{10}$ of the
$L^2$ error over a range of $p$ and $n_{\mathrm{quad}}$, for the PC
expansion of the peak enthalpy release rate. The $n_{\mathrm{quad}}$
values are approximate, as DASQ is terminated at the end of the
iteration that exceeds $n_{\mathrm{quad}}$. When $n_{\mathrm{quad}}$
is too small, the error is dominated by aliasing (integration) error
and increases with $p$. When a sufficiently large $n_{\mathrm{quad}}$
is used such that truncation error dominates, exponential convergence
with respect to $p$ can be observed, as expected for smooth
functions. Ideally, $n_{\mathrm{quad}}$ and $p$ should be selected at
the ``knees'' of these contour plots, since little accuracy can be
gained when $n_{\mathrm{quad}}$ is increased any further, but these
locations can be difficult to pinpoint \textit{a priori}.

\subsection{Design of a single experiment}

\begin{figure}[htb]
  \centering
  \subfigure[Full ODE model]
  {
    \includegraphics[width=0.47\textwidth]{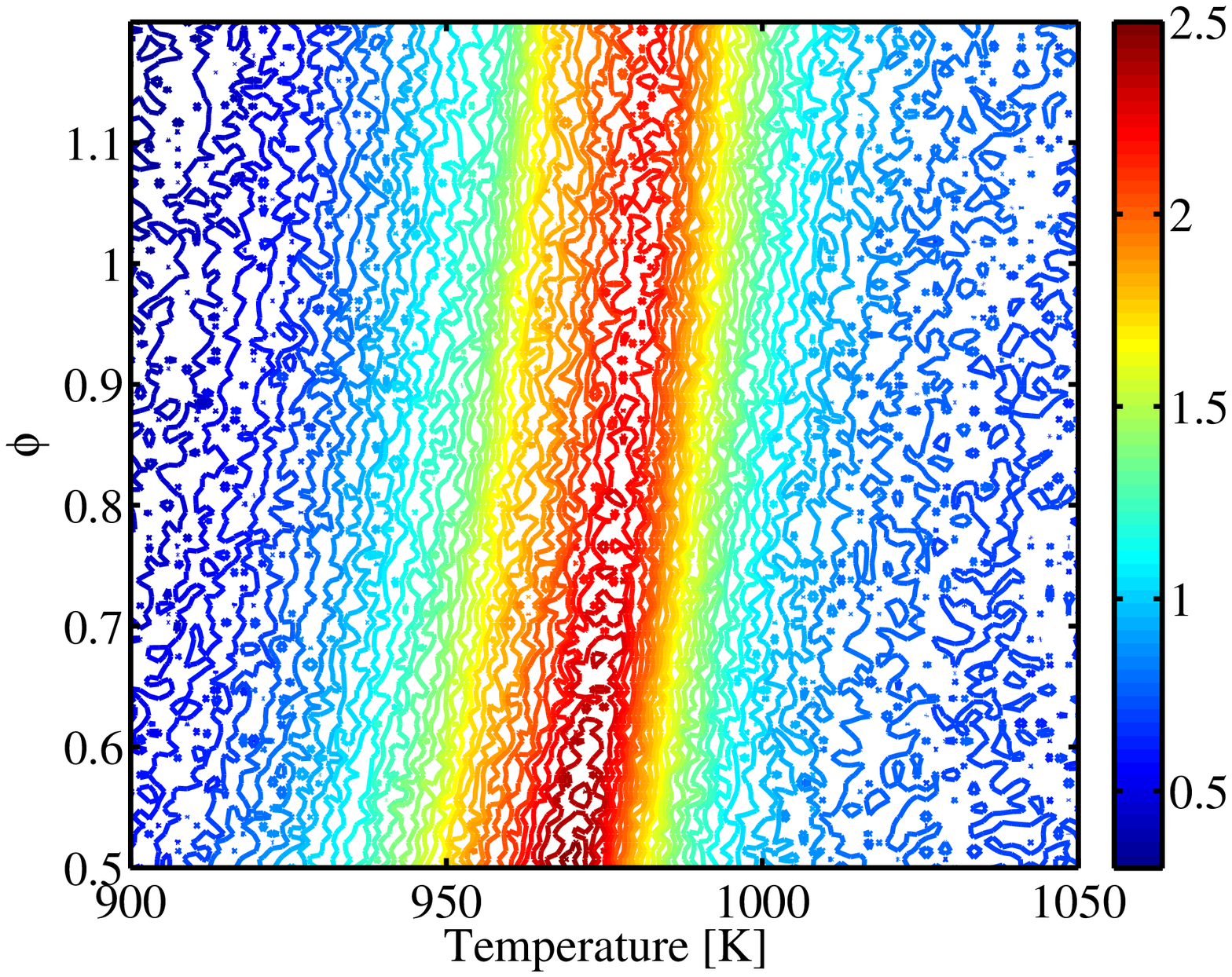}
    \label{f:canteraDesignAllObs}
  }
  \subfigure[PC surrogate]
  {
    \includegraphics[width=0.47\textwidth]{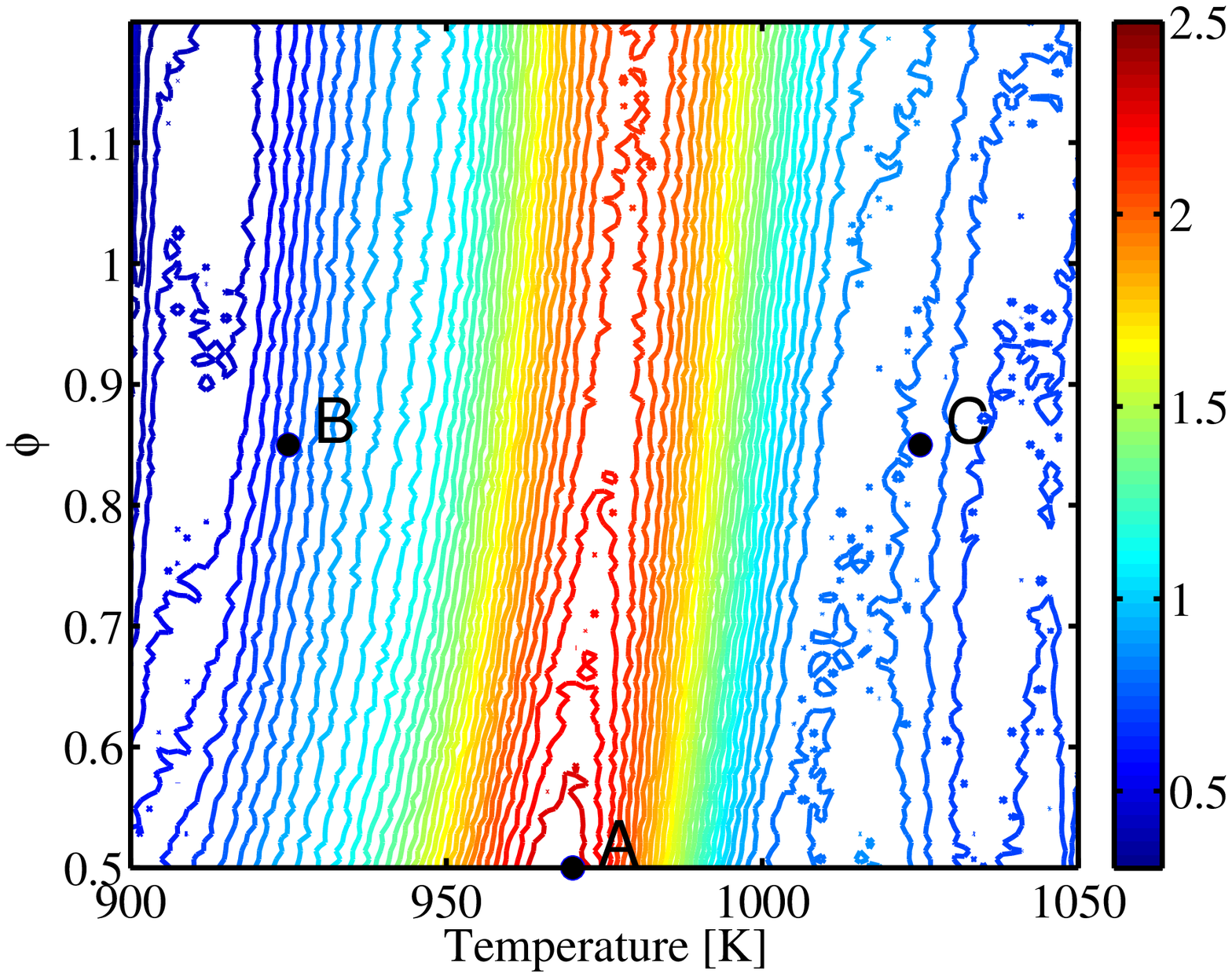}
    \label{f:p12n1000000DesignAllObs}
  }
  \caption{Estimated expected utility contours in the single-experiment
    combustion design problem, with design variables $T_0$ and
    $\phi$, using the full ODE model and the PC surrogate with
    $p=12$ and $n_{\mathrm{quad}}=10^6$. Inference problems are then solved
    at experimental conditions $A$, $B$, and $C$ to validate the experimental
    design procedure.}
  \label{f:1ExpOptimaDesign}
\end{figure}

Figures~\ref{f:canteraDesignAllObs}
and~\ref{f:p12n1000000DesignAllObs} show contours of the expected
utility estimates, $\hat{U}(\bd)$, in the two-dimensional design
space, constructed using the full ODE model (with estimator parameters
$n_{\mathrm{in}}=n_{\mathrm{out}}=10^3$) and the PC surrogate (with
estimator parameters $n_{\mathrm{in}}=n_{\mathrm{out}}=10^4$),
respectively. Contours from the PC surrogate are very similar to those
from the full model, though the former have less variability due to
the larger number of Monte Carlo samples used to compute
$\hat{U}$. Most importantly, both plots yield the same optimal
experimental design at around $(T_0^\ast,\phi^\ast) = (975,0.5)$.

\begin{figure}[htb]
  \centering
  \subfigure[Design $A$ full ODE model]
  {
    \includegraphics[width=0.47\textwidth]{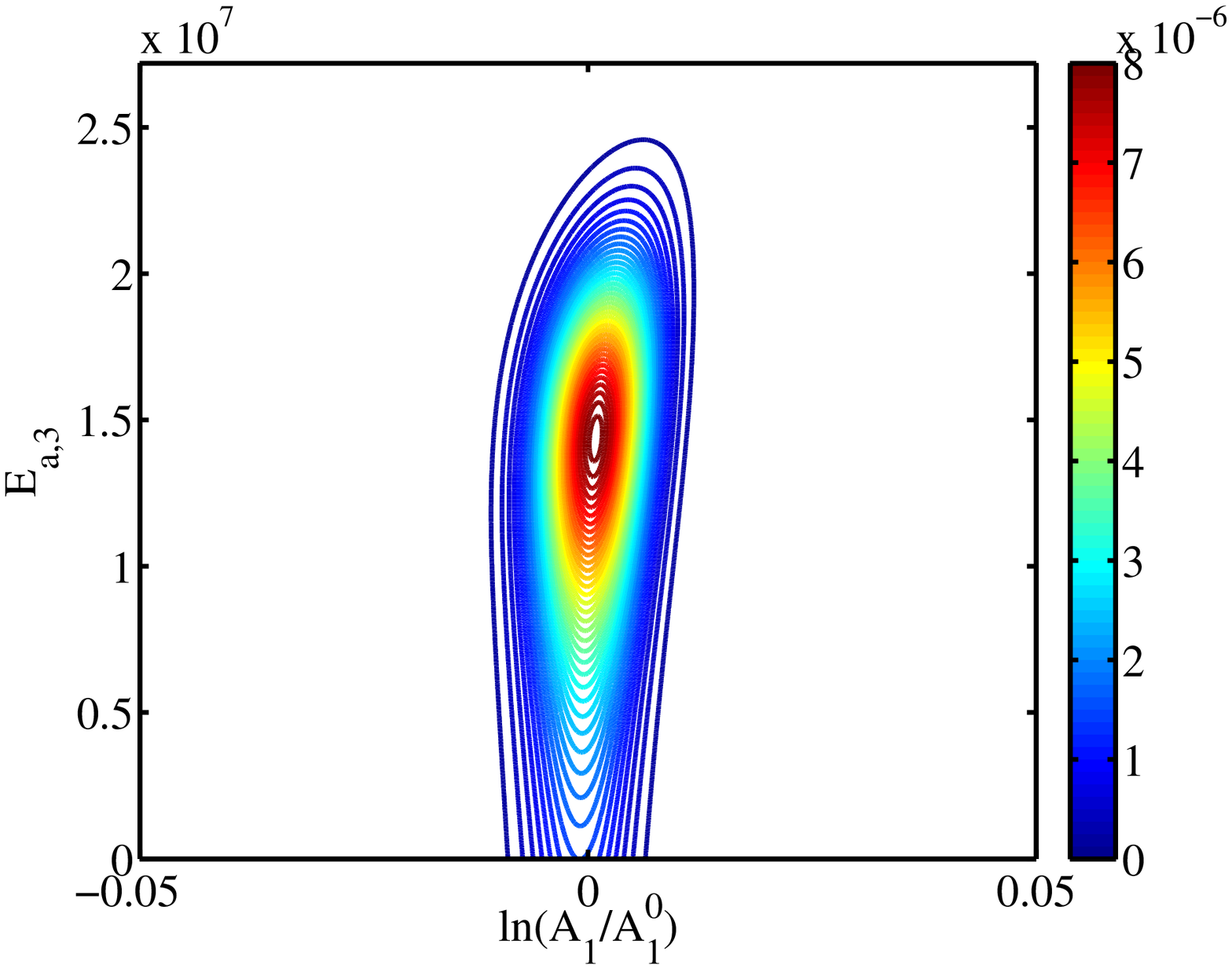}
  }
  \subfigure[Design $A$ PC surrogate]
  {
    \includegraphics[width=0.47\textwidth]{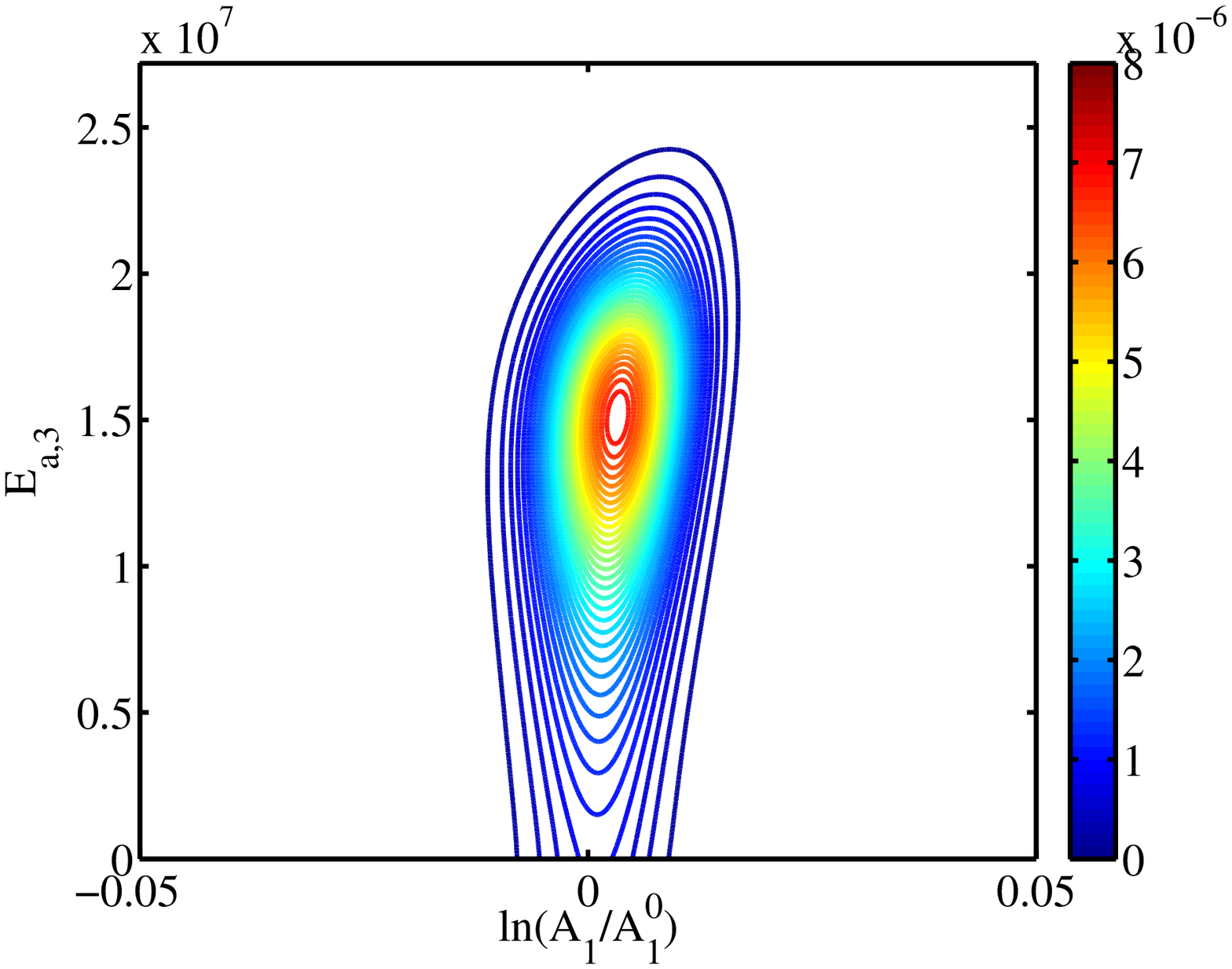}
  }
  \subfigure[Design $B$ full ODE model]
  {
    \includegraphics[width=0.47\textwidth]{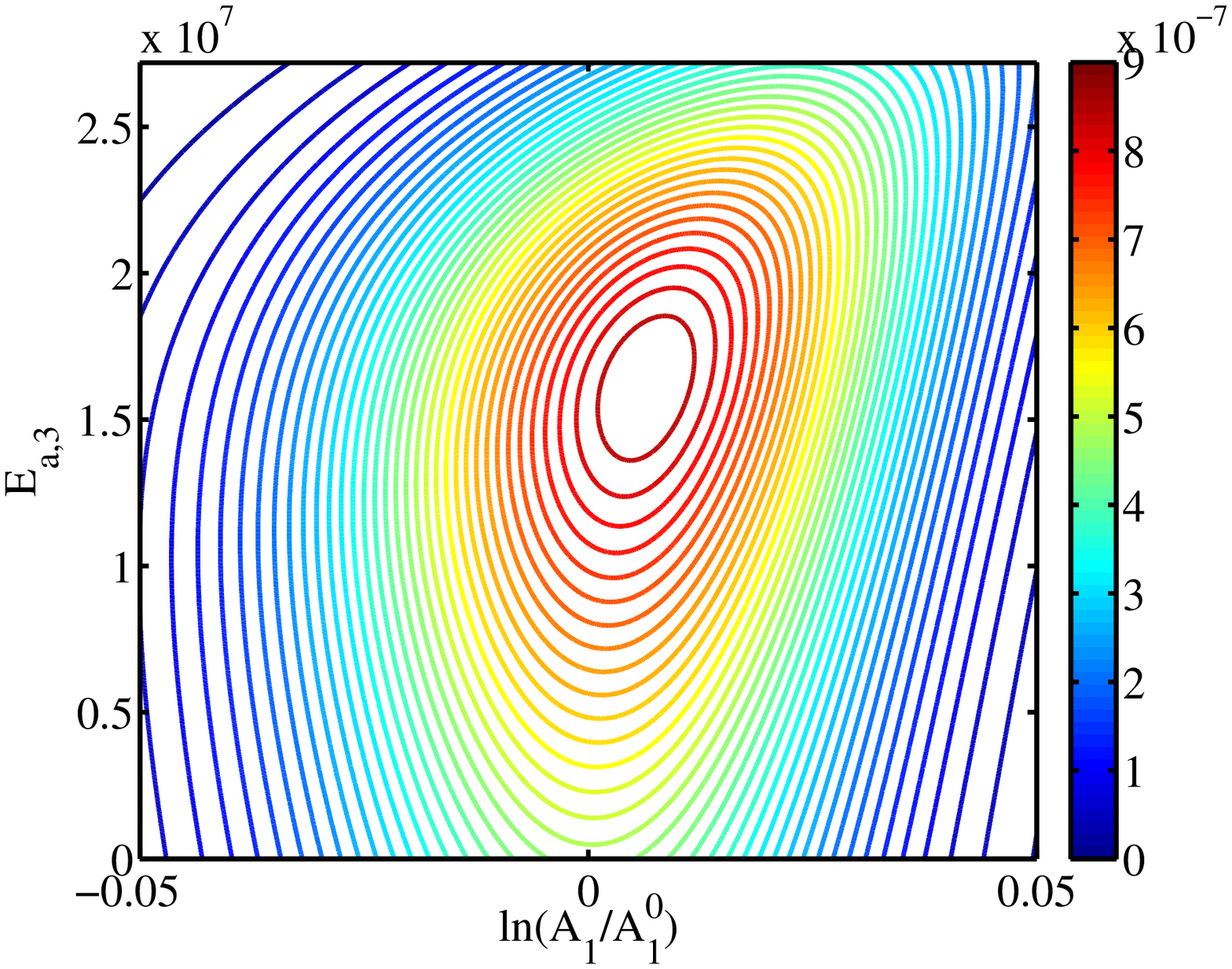}
  }
  \subfigure[Design $B$ PC surrogate]
  {
    \includegraphics[width=0.47\textwidth]{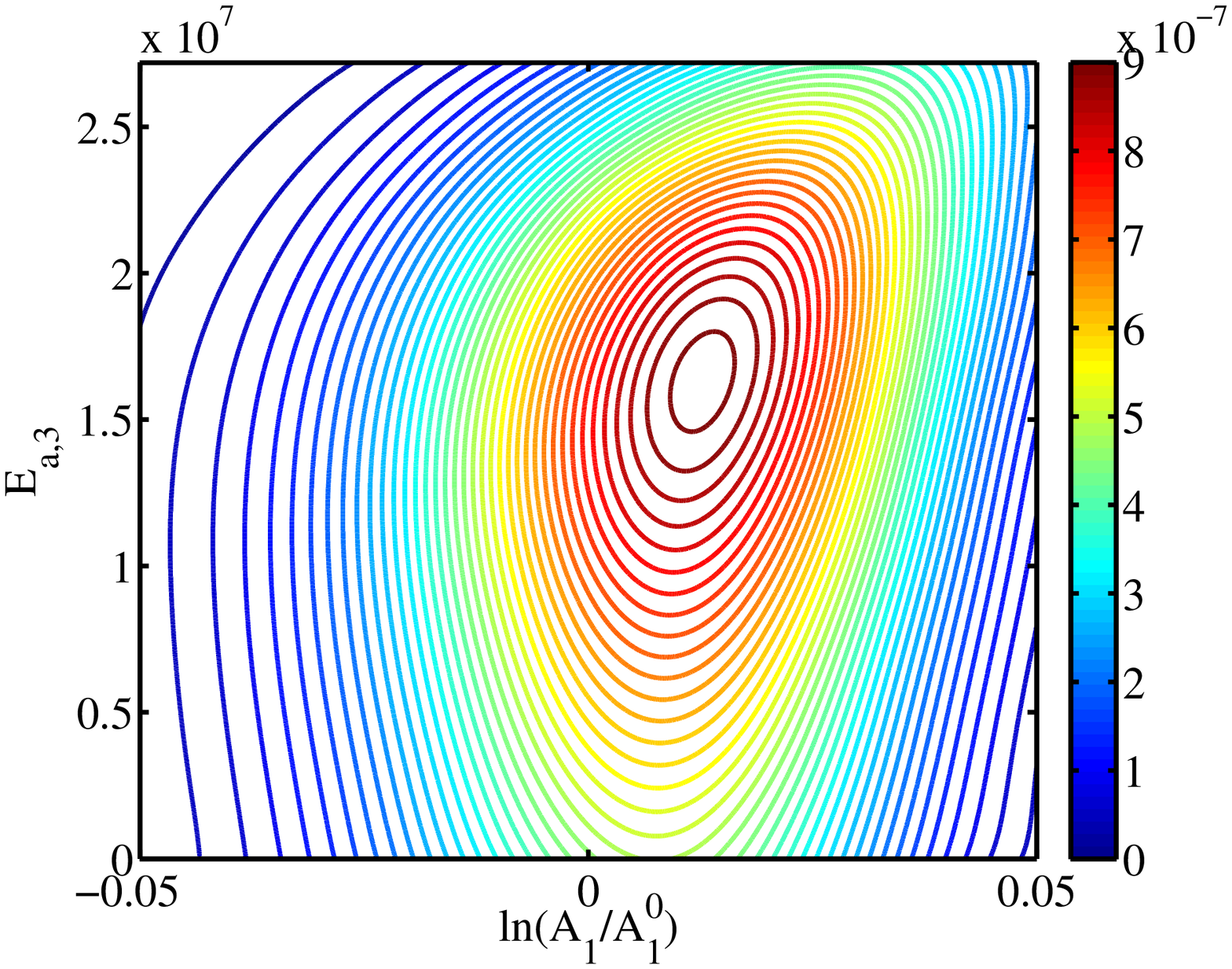}
  }
  \subfigure[Design $C$ full ODE model]
  {
    \includegraphics[width=0.47\textwidth]{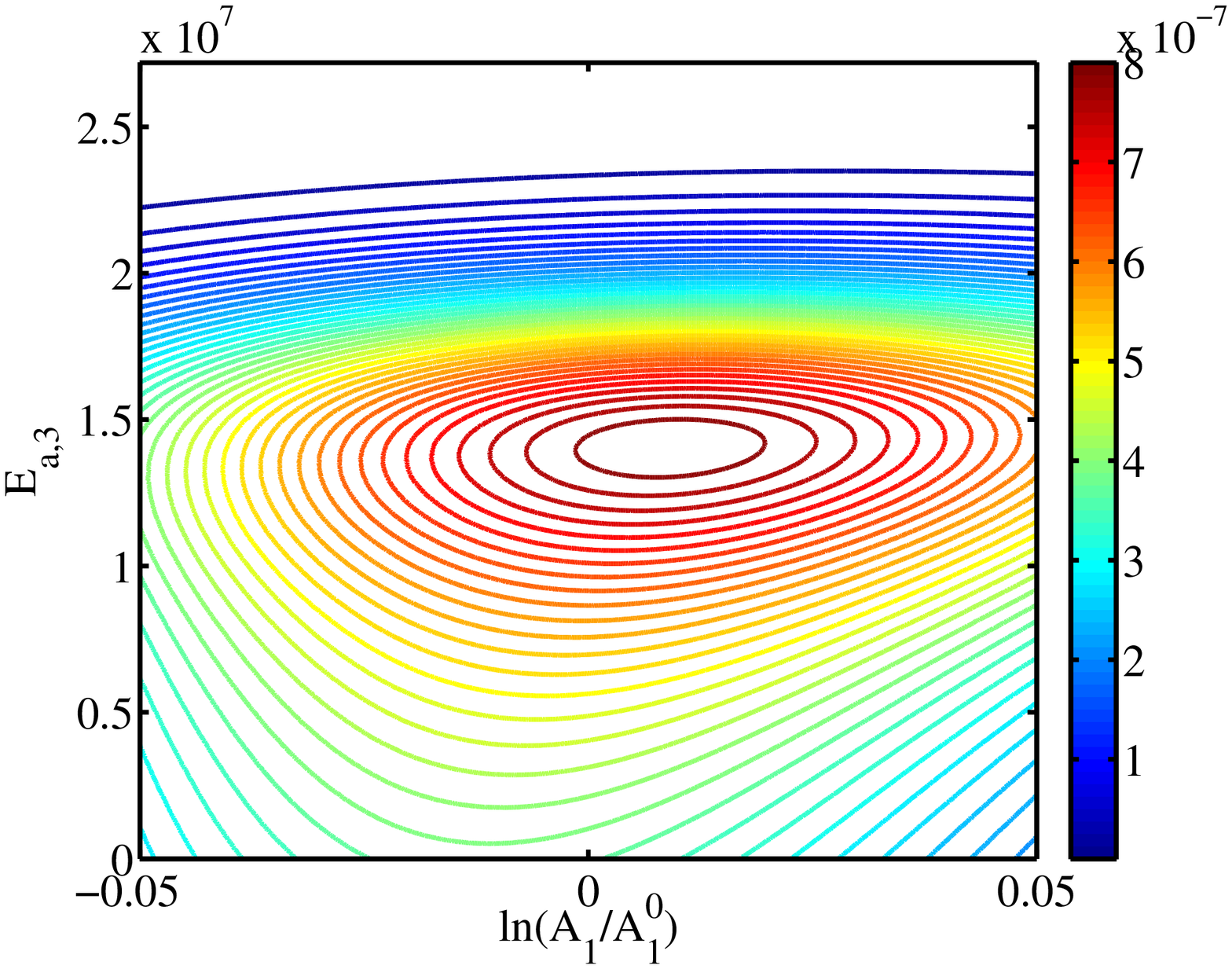}
  }
  \subfigure[Design $C$ PC surrogate]
  {
    \includegraphics[width=0.47\textwidth]{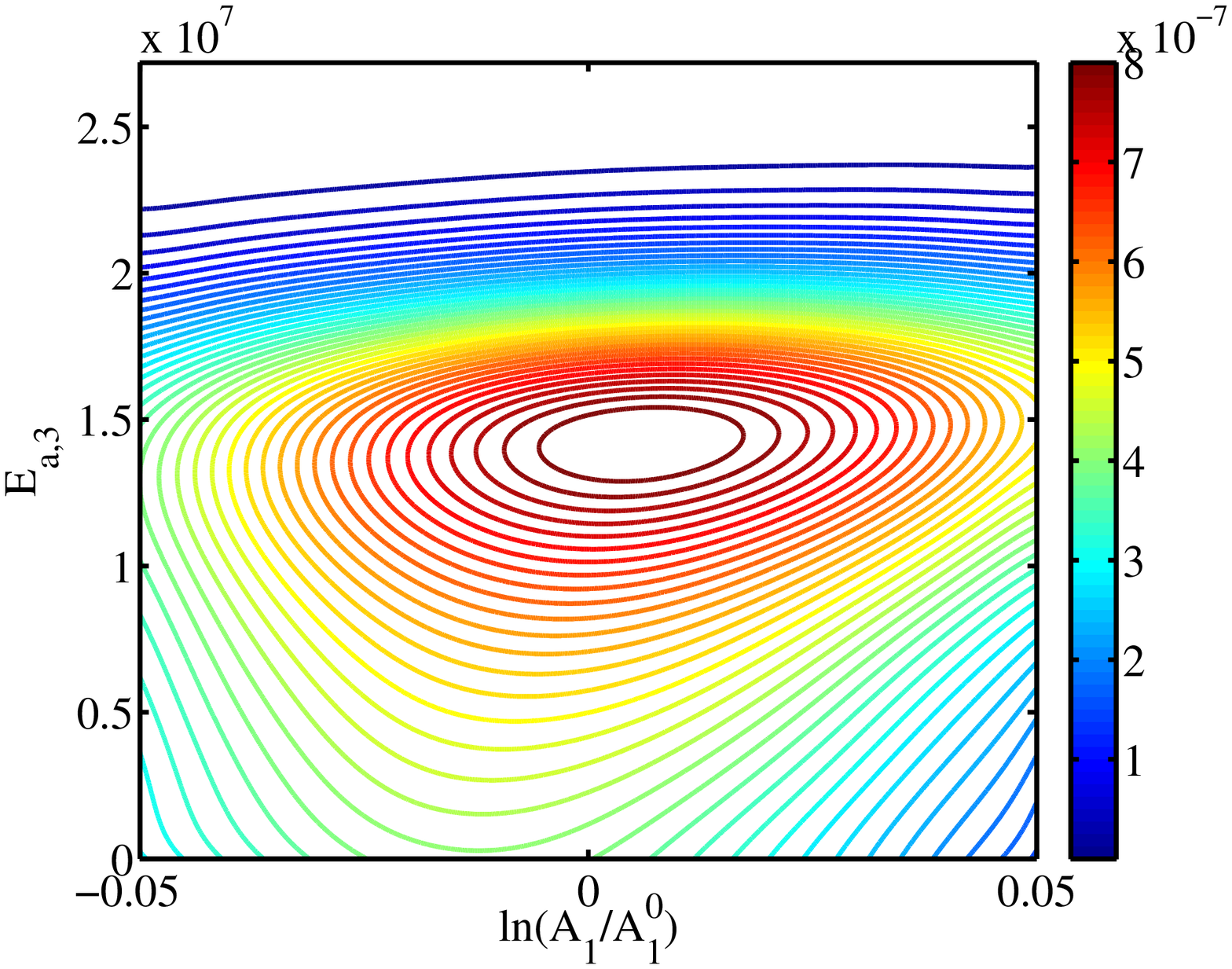}
  }
  \caption{Contours of posterior density of the kinetic parameters,
    showing the results of inference with data obtained at three
    different experimental conditions (designs $A$, $B$, and
    $C$). Left column: posteriors constructed using the full ODE
    model; right column: posteriors constructed via the PC surrogate
    with $p=12$, $n_{\mathrm{quad}}=10^6$.}
  \label{f:inference}
\end{figure}

To test how well the expected information gain anticipates the
performance of an experiment, the inference problem is solved at three
different design points $A$, $B$, and $C$, listed in
Table~\ref{t:inferenceConditions} and illustrated in
Figure~\ref{f:p12n1000000DesignAllObs}. Since the expected utility is
highest at design $A$, then the posterior is \textit{expected} to
reflect the largest information gain at that experimental
condition. We use the full ODE model to generate artificial data at
each of the three design conditions, then perform inference. Contours
of posterior density are shown in Figure~\ref{f:inference}, using the
full ODE model and the PC surrogate. The posteriors of the full model
and PC surrogate match very well; hence the PC surrogate is suitable
not only for experimental design, but also for inference. As expected
from the expected utility plots, the posterior distribution of the
kinetic parameters is tightest at design $A$; this was the most
informative of the three experimental conditions. Posterior modes of
the ODE model and PC results are not precisely the same, however, due
to the modeling error associated with the PC surrogate. Also, the
posterior modes obtained with the full ODE model do not exactly match
the values used to generate the artificial data, due to the noise in
the likelihood model.

\begin{figure}[htb]
  \centering
  \subfigure[All observables (same as Figure~\ref{f:p12n1000000DesignAllObs})]
  {
    \includegraphics[width=0.47\textwidth]{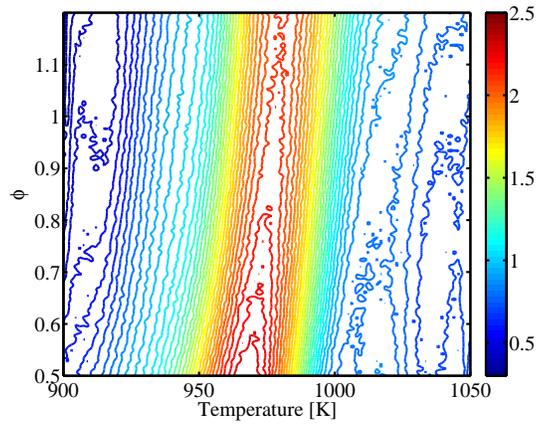}
  }\\
  \subfigure[Characteristic time observables]
  {
    \includegraphics[width=0.47\textwidth]{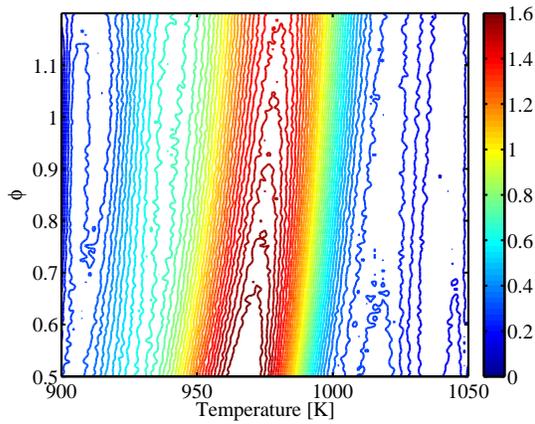}
  }
  \subfigure[Peak value observables]
  {
    \includegraphics[width=0.47\textwidth]{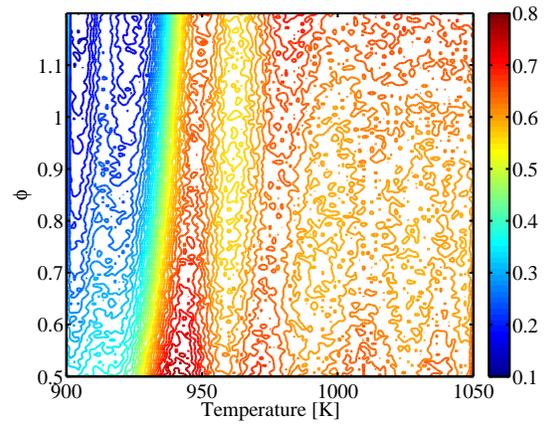}
  }
  \caption{Estimated expected utility contours in the
    single-experiment combustion design problem using the PC surrogate
    with $p=12$ and $n_{\mathrm{quad}}=10^6$, but with different sets
    of observables. }
  \label{f:1ExpOptimaDesignSubsetObs}
\end{figure}

What if a different set of observables are used? Two cases are
explored: first, using only the characteristic time observables (i.e.,
the first five rows of Table~\ref{t:obs}); and second, using only the
peak value observables (i.e., the last five rows of
Table~\ref{t:obs}). The corresponding expected utility plots are given
in Figure~\ref{f:1ExpOptimaDesignSubsetObs} (using the PC surrogate
only). Several remarks can be made. First, the characteristic time
observables are more informative than the peak value observables, as
demonstrated by the higher expected utility values in
Figure~\ref{f:1ExpOptimaDesignSubsetObs}(b) than
Figure~\ref{f:1ExpOptimaDesignSubsetObs}(c). Second, the choice of
observables can greatly influence the optimal value of the design
parameters. Third, even though the observables from the two cases form
a partition of the full observable set, their expected utility values
do not simply sum to that of the full-observable case. (This is a
special case of the analysis in \ref{app:infoGain}.)  The lesson is
that the selection of appropriate observables is a very important part
of the design procedure, especially if one is forced to select only a
few modes of observation. This selection could be made into an
argument of the objective function, augmenting $\bd$ and leading to a
mixed-integer optimization problem.

\subsection{Design of two experiments: stochastic optimization}

Now we perform a study analogous to that in
Section~\ref{s:nonlinearDesignOfTwoExperiments}, designing two
ignition experiments (of the same structure) simultaneously. The
experimental goal of inferring $A_1$ and $E_{a,3}$ is unchanged, and
for computational efficiency we use only the $p=12$,
$n_{\mathrm{quad}}=10^6$ PC surrogate. The design space is now
four-dimensional, with $\bd =
[T_{0,1},\phi_1,T_{0,2},\phi_2]$. Stochastic optimization is used to
find the optimal experimental design, as a grid search is entirely
impractical. 

Coupling stochastic optimization schemes
(Section~\ref{s:stochasticOptimization}) with the estimator
$\hat{U}(\bd)$ of expected information gain introduces a few new
numerical tradeoffs. The number of samples in the outer loop of the
estimator controls the variance of $\hat{U}$, which dictates the noise
level of the objective function. Lower noise in the objective function
might imply fewer optimization iterations overall, while a noisier
objective may require many more iterations of either SPSA or NMNS to
make progress towards the optimum. On the other hand, noise should not
be reduced too much for SPSA, since the usefulness of its gradient
approximation relies on the existence of a non-negligible noise
level. In general, the task of balancing $n_{\mathrm{out}}$ against
the number of optimization iterations, in order to minimize the number
of model evaluations, is not trivial. We therefore test different
values of $n_{\mathrm{out}}$ to understand their impact on the SPSA
and NMNS optimization schemes. We fix $n_{\mathrm{in}}$ at $10^4$ in
order to maintain a low bias contribution from the inner loop.

Since the results of stochastic optimization are themselves random, we
use an ensemble of 100 independent optimization runs at any
given parameter setting to analyze performance.  Each optimization run
is capped at $n_{\mathrm{noisyObj}} = 10^4$ evaluations of the noisy
objective---i.e., of the summand in
Equation~(\ref{e:expectedUtilityMC})---where each noisy objective
evaluation itself involves $n_{\mathrm{in}}$
evaluations of the model $\bG$ or the surrogate $\bG^p$. We can thus
compare performance at a fixed computational cost.

\begin{figure}[htb]
  \centering 
  \subfigure[SPSA $n_{\mathrm{out}}=1$]
  {
    \includegraphics[width=0.47\textwidth]{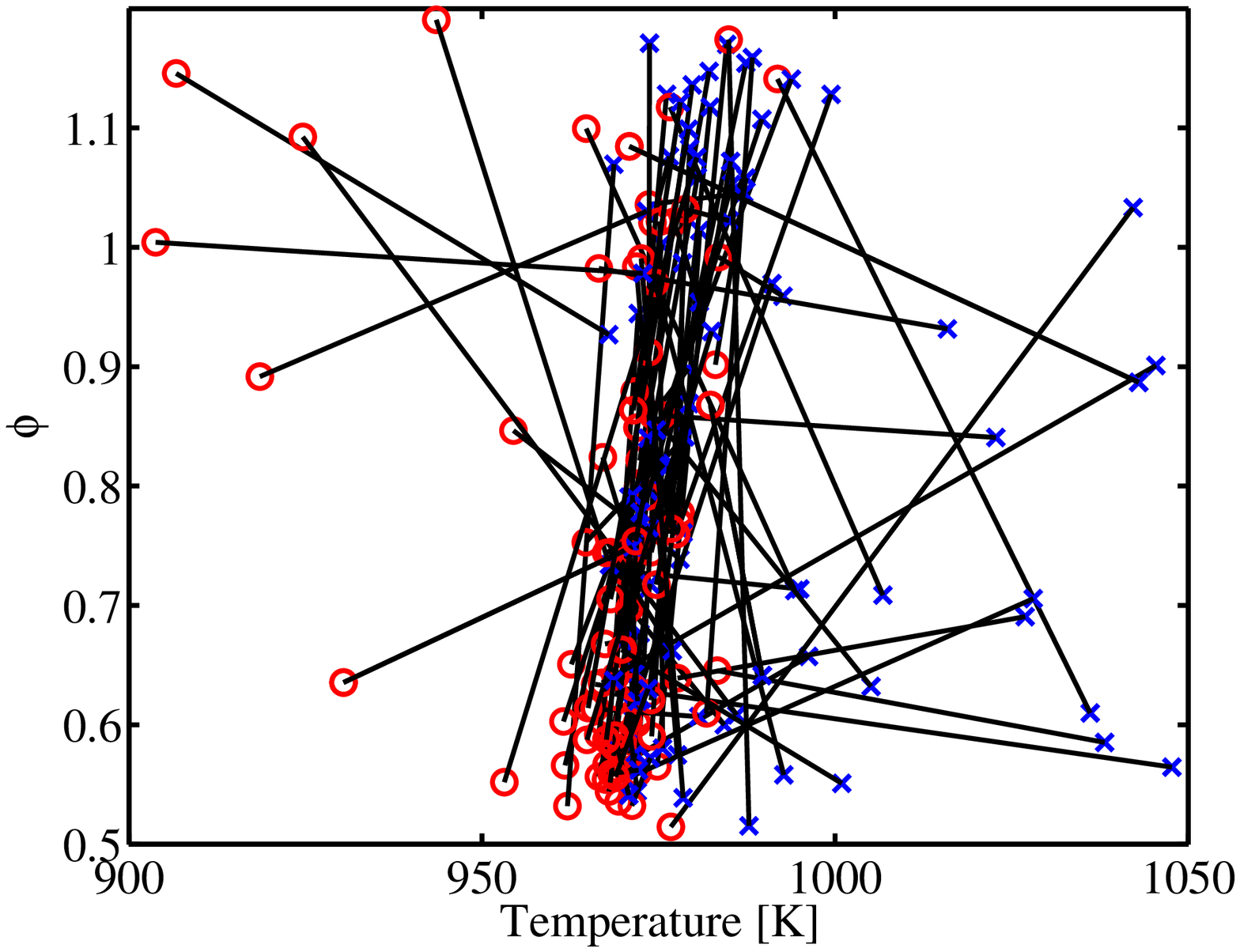}
  }
  \subfigure[NMNS $n_{\mathrm{out}}=1$]
  {
    \includegraphics[width=0.47\textwidth]{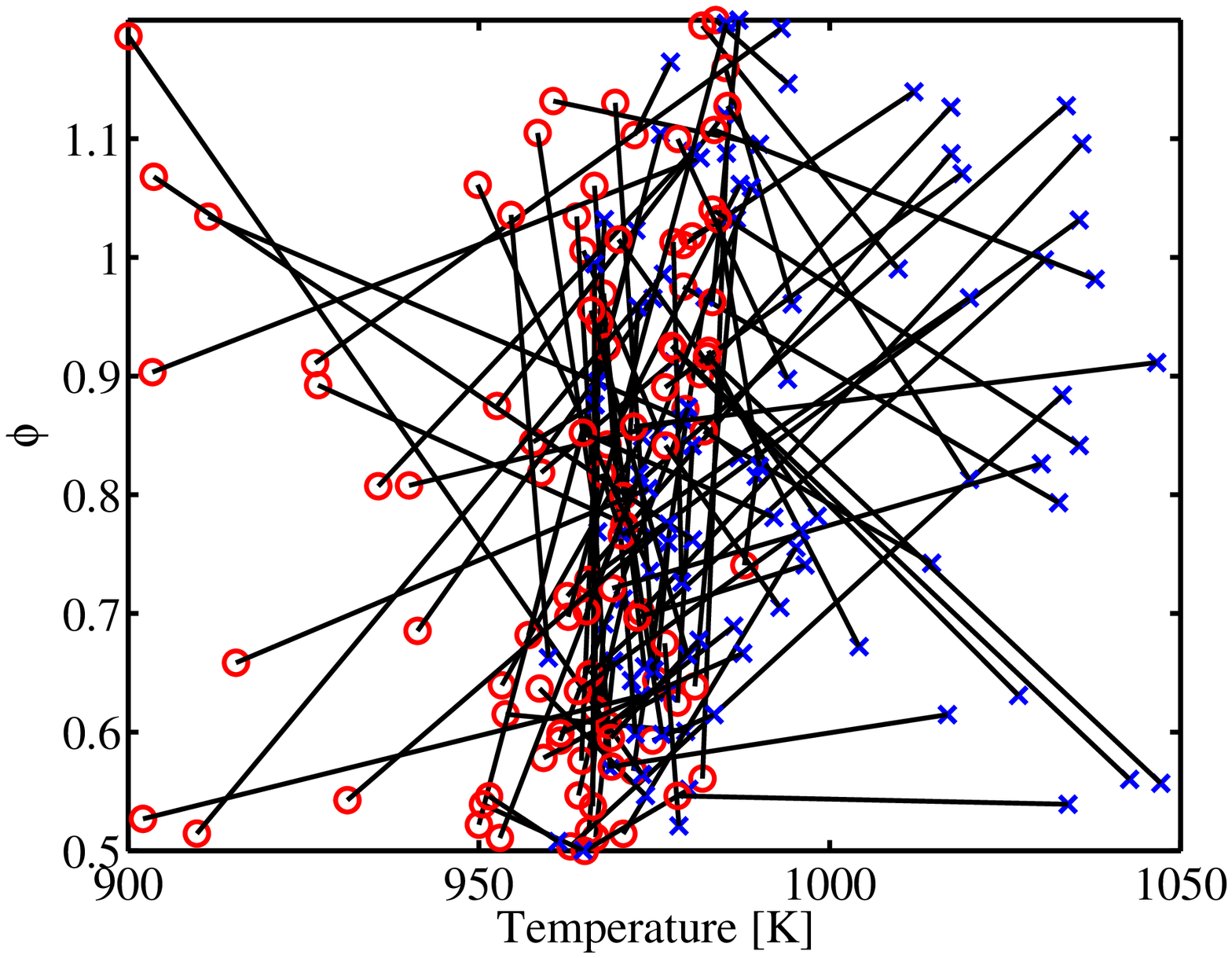}
  }
  \subfigure[SPSA $n_{\mathrm{out}}=10$]
  {
    \includegraphics[width=0.47\textwidth]{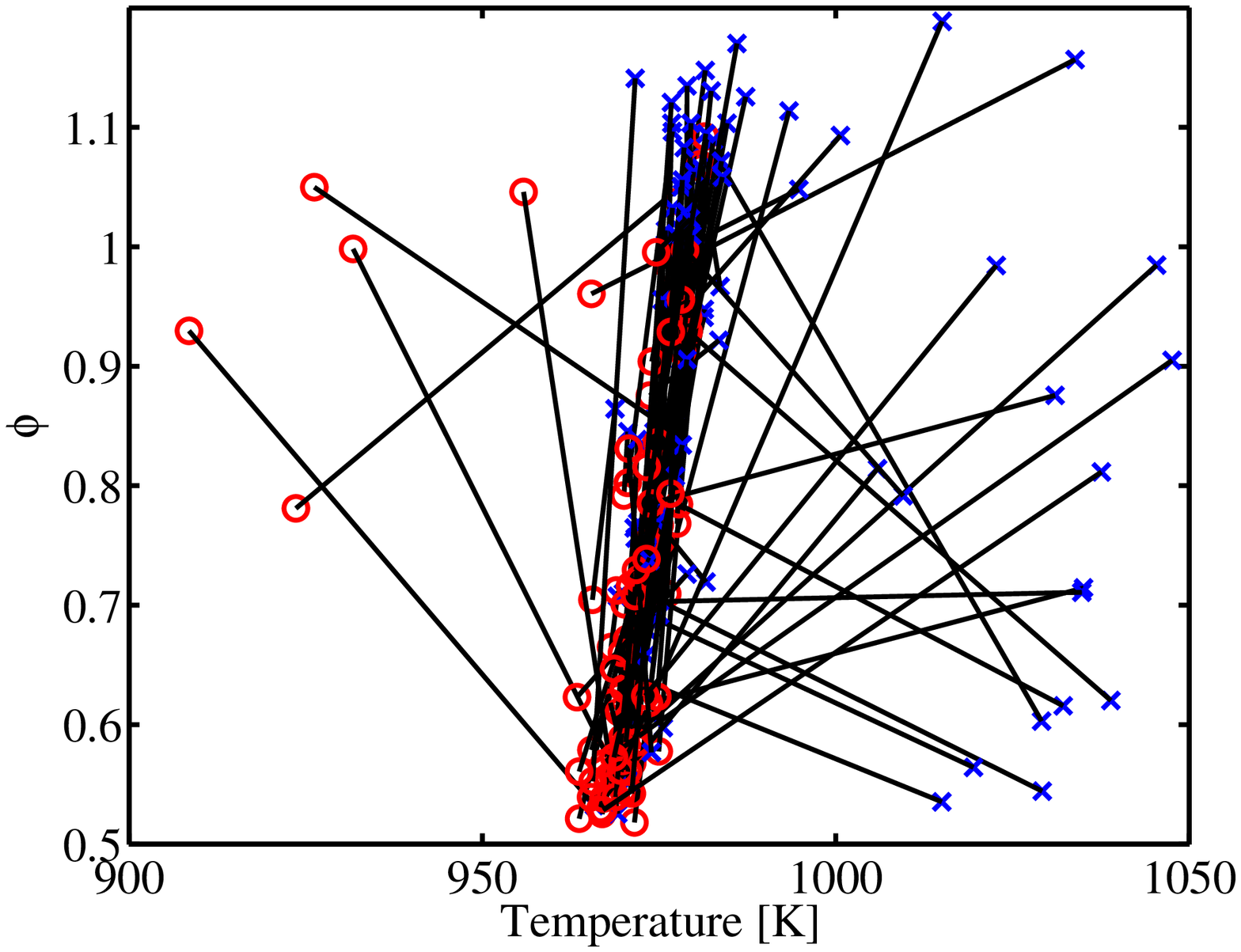}
  }
  \subfigure[NMNS $n_{\mathrm{out}}=10$]
  {
    \includegraphics[width=0.47\textwidth]{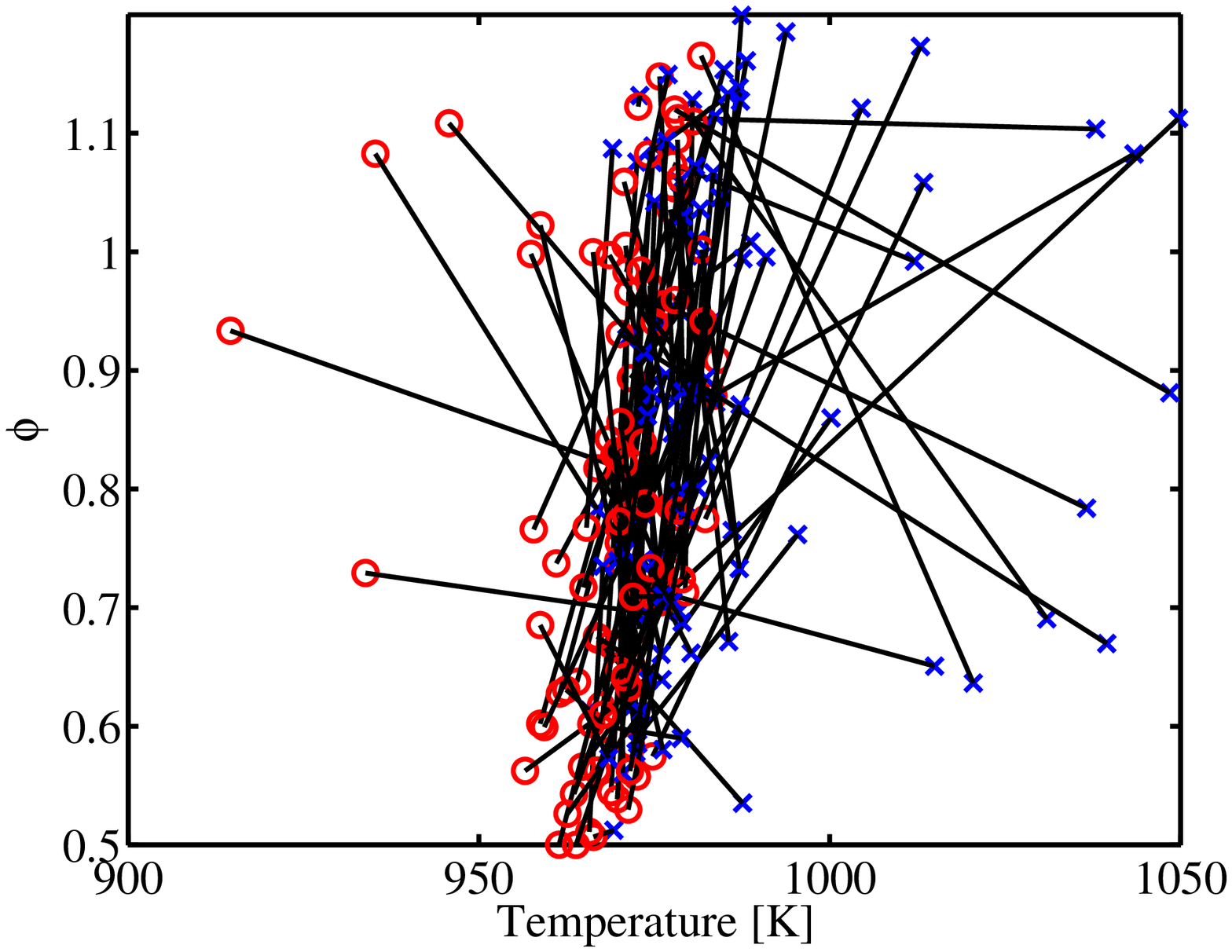}
  }
  \subfigure[SPSA $n_{\mathrm{out}}=100$]
  {
    \includegraphics[width=0.47\textwidth]{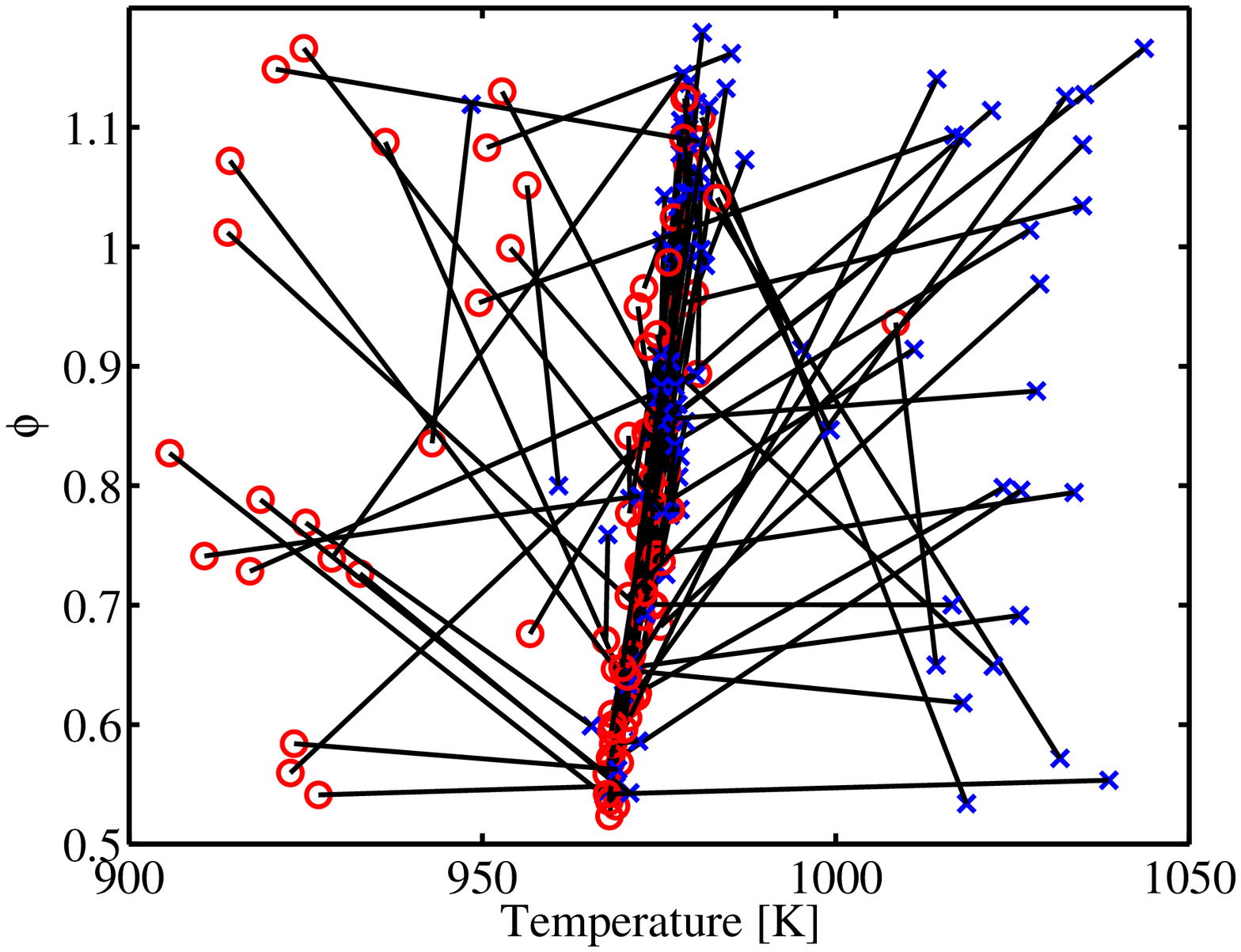}
  }
  \subfigure[NMNS $n_{\mathrm{out}}=100$]
  {
    \includegraphics[width=0.47\textwidth]{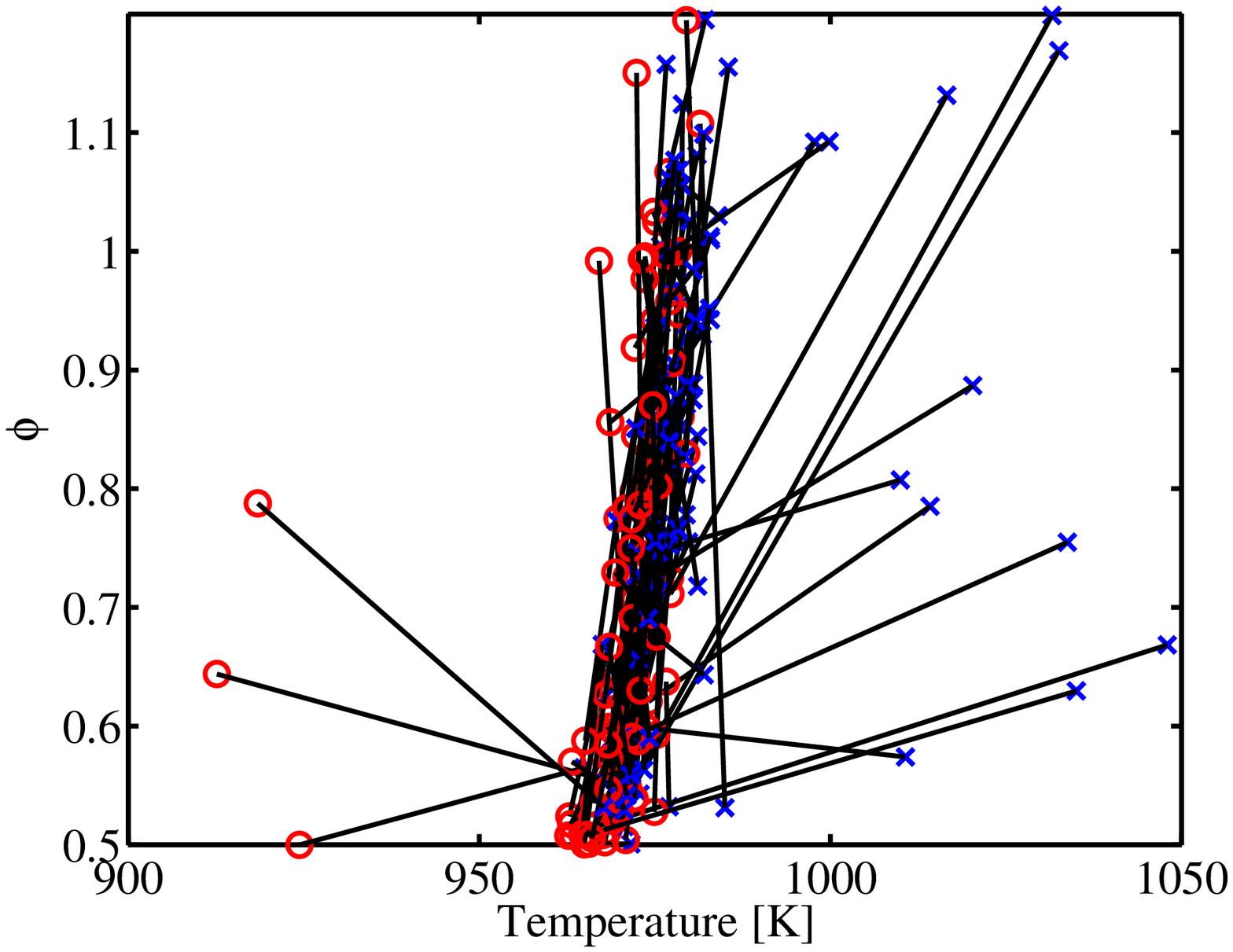}
  }
  \caption{Two-experiment combustion design problem: final outputs
    from 100 independent runs of stochastic optimization, using SPSA
    and NMNS with a limit of $n_{\mathrm{noisyObj}} = 10^4$, and with
    different numbers of outer Monte Carlo iterations, using the PC
    surrogate with $p=12$, $n_{\textrm{quad}}=10^6$.}
  \label{f:finalXp12}
\end{figure}

\begin{figure}[htb]
  \centering 
  \subfigure[SPSA $n_{\mathrm{out}}=1$]
  {
    \includegraphics[width=0.47\textwidth]{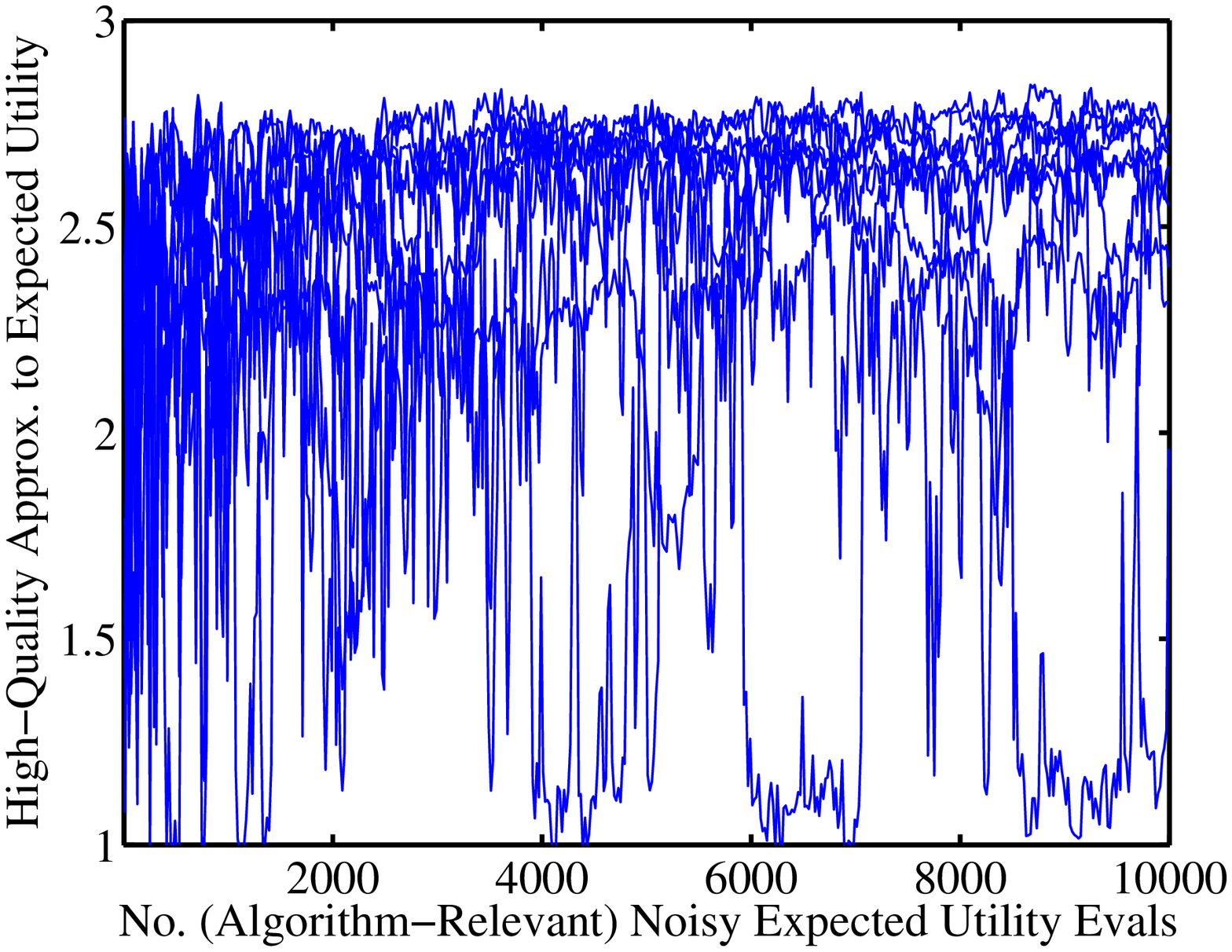}
  }
  \subfigure[NMNS $n_{\mathrm{out}}=1$]
  {
    \includegraphics[width=0.47\textwidth]{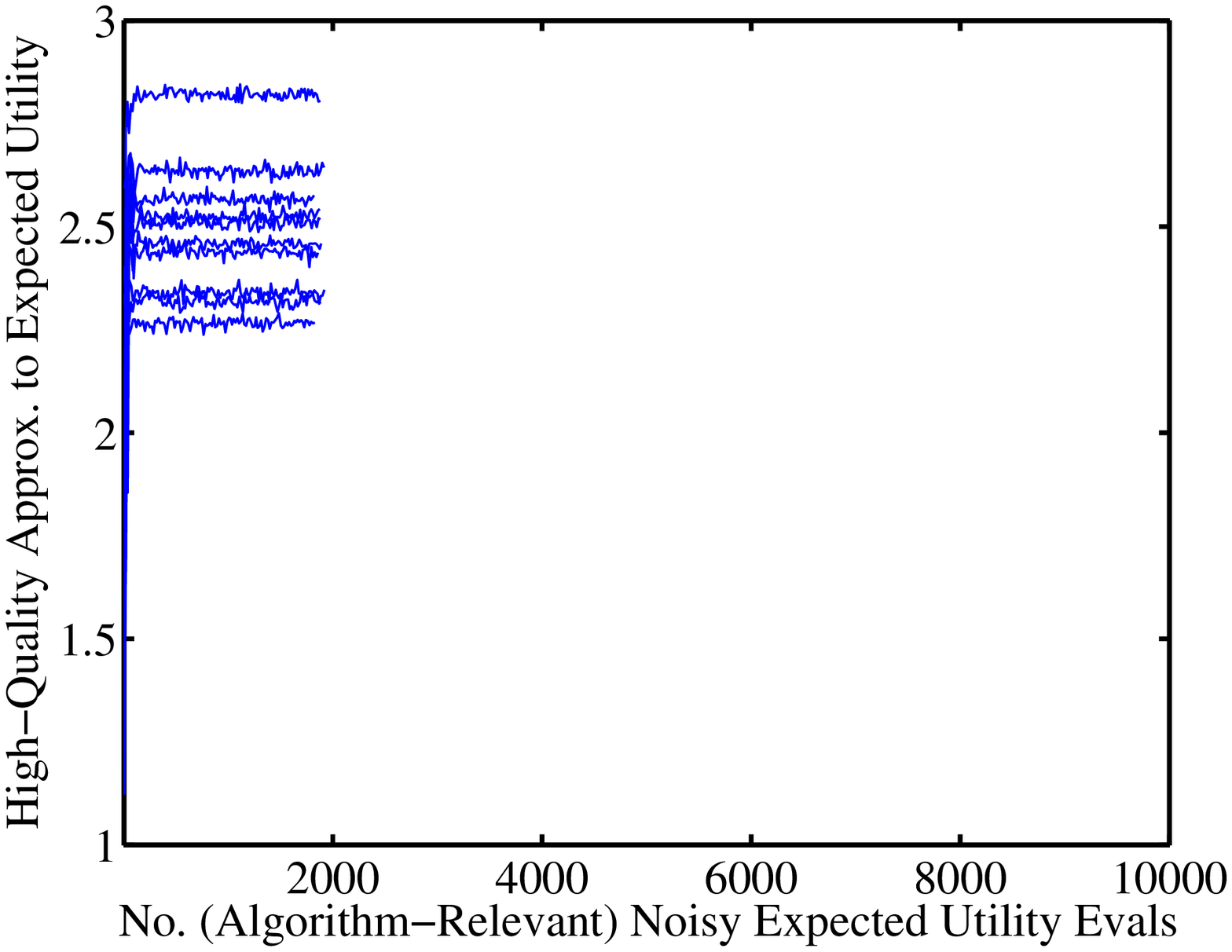}
  }
  \subfigure[SPSA $n_{\mathrm{out}}=10$]
  {
    \includegraphics[width=0.47\textwidth]{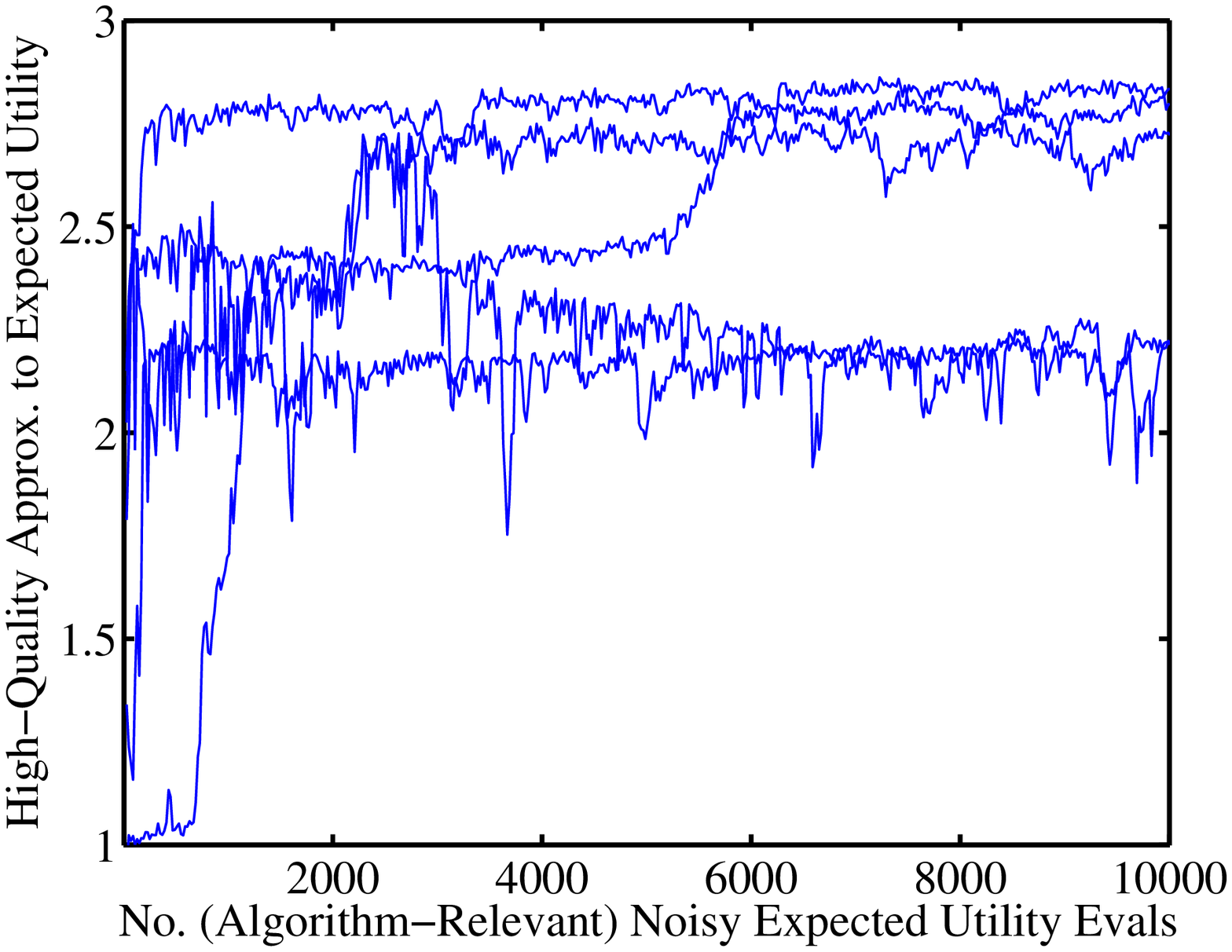}
  }
  \subfigure[NMNS $n_{\mathrm{out}}=10$]
  {
    \includegraphics[width=0.47\textwidth]{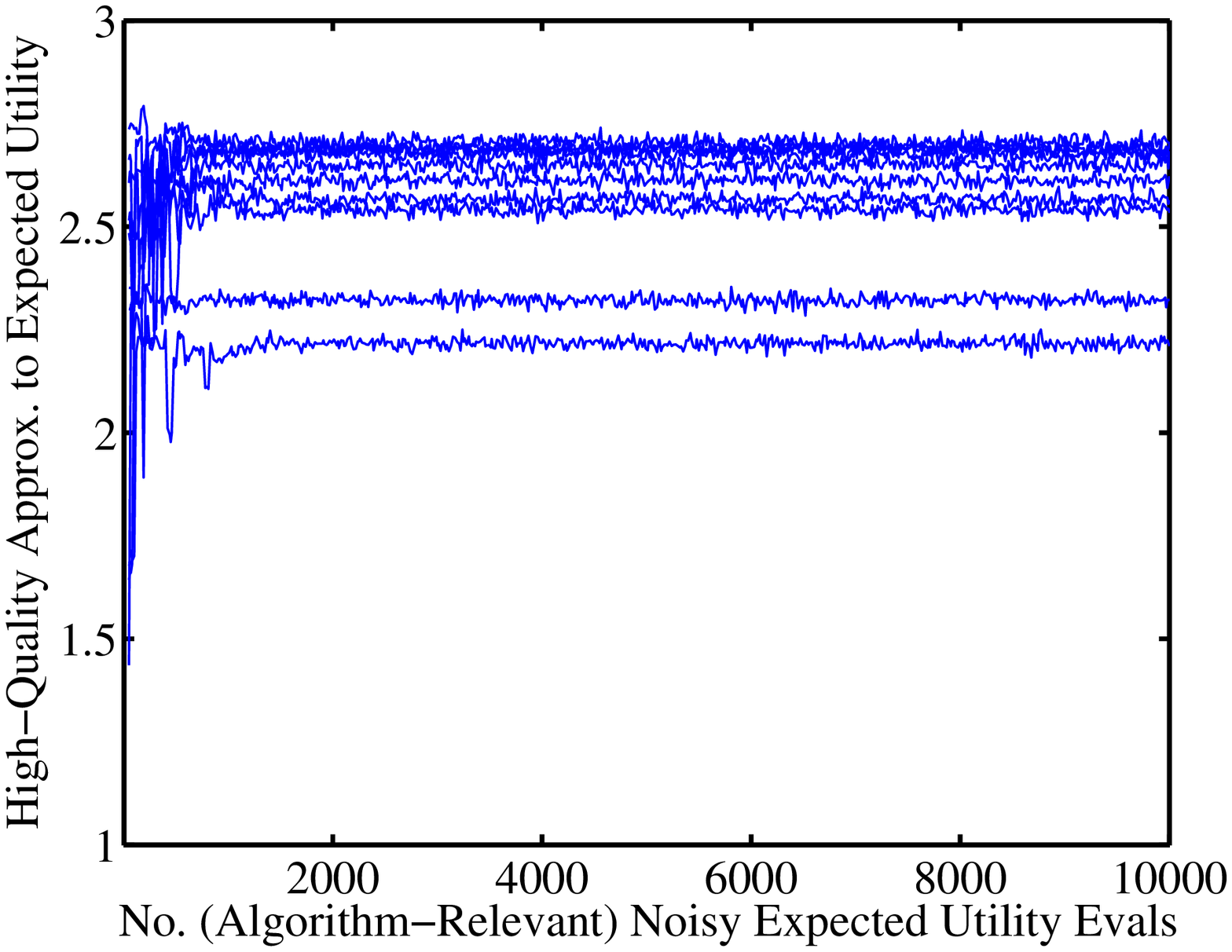}
  }
  \subfigure[SPSA $n_{\mathrm{out}}=100$]
  {
    \includegraphics[width=0.47\textwidth]{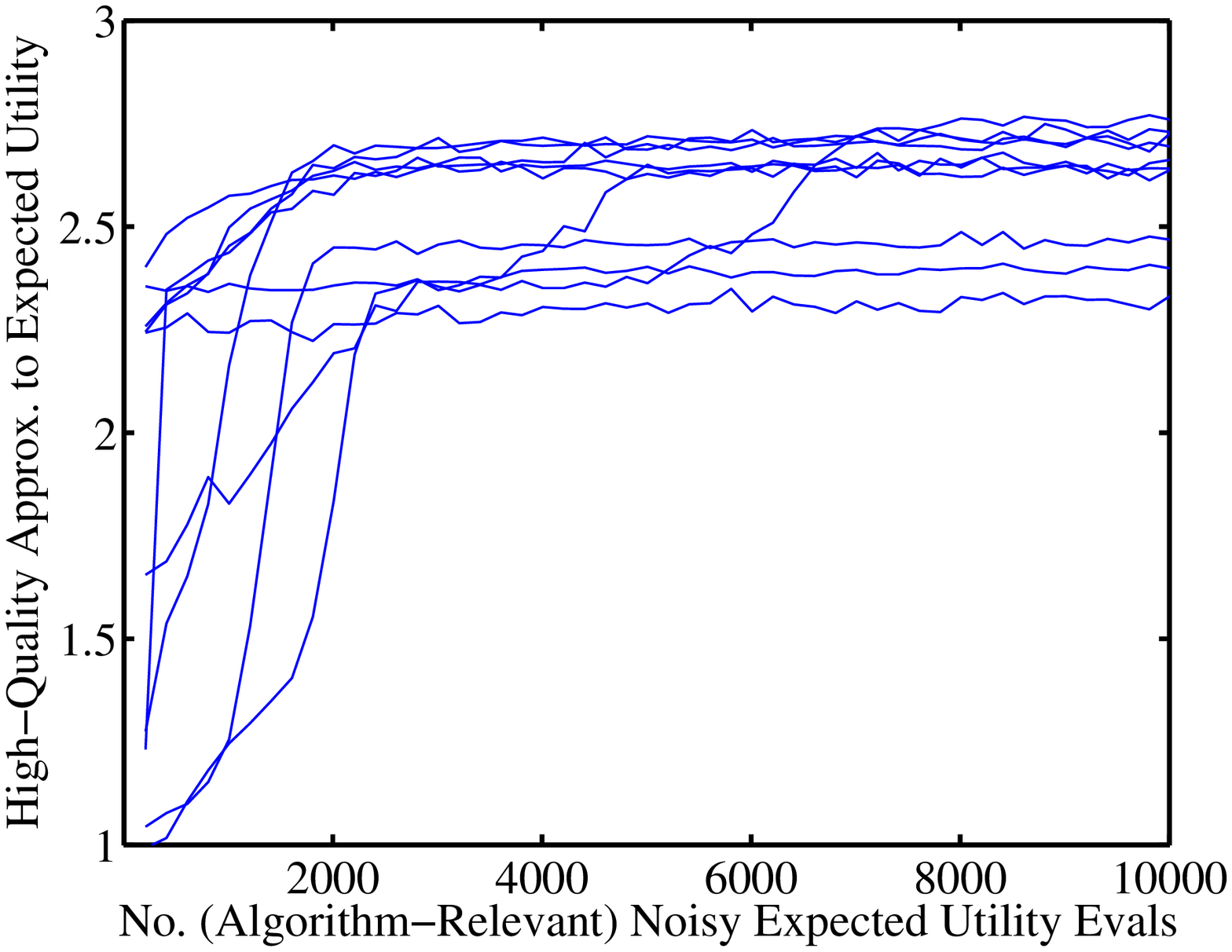}
  }
  \subfigure[NMNS $n_{\mathrm{out}}=100$]
  {
    \includegraphics[width=0.47\textwidth]{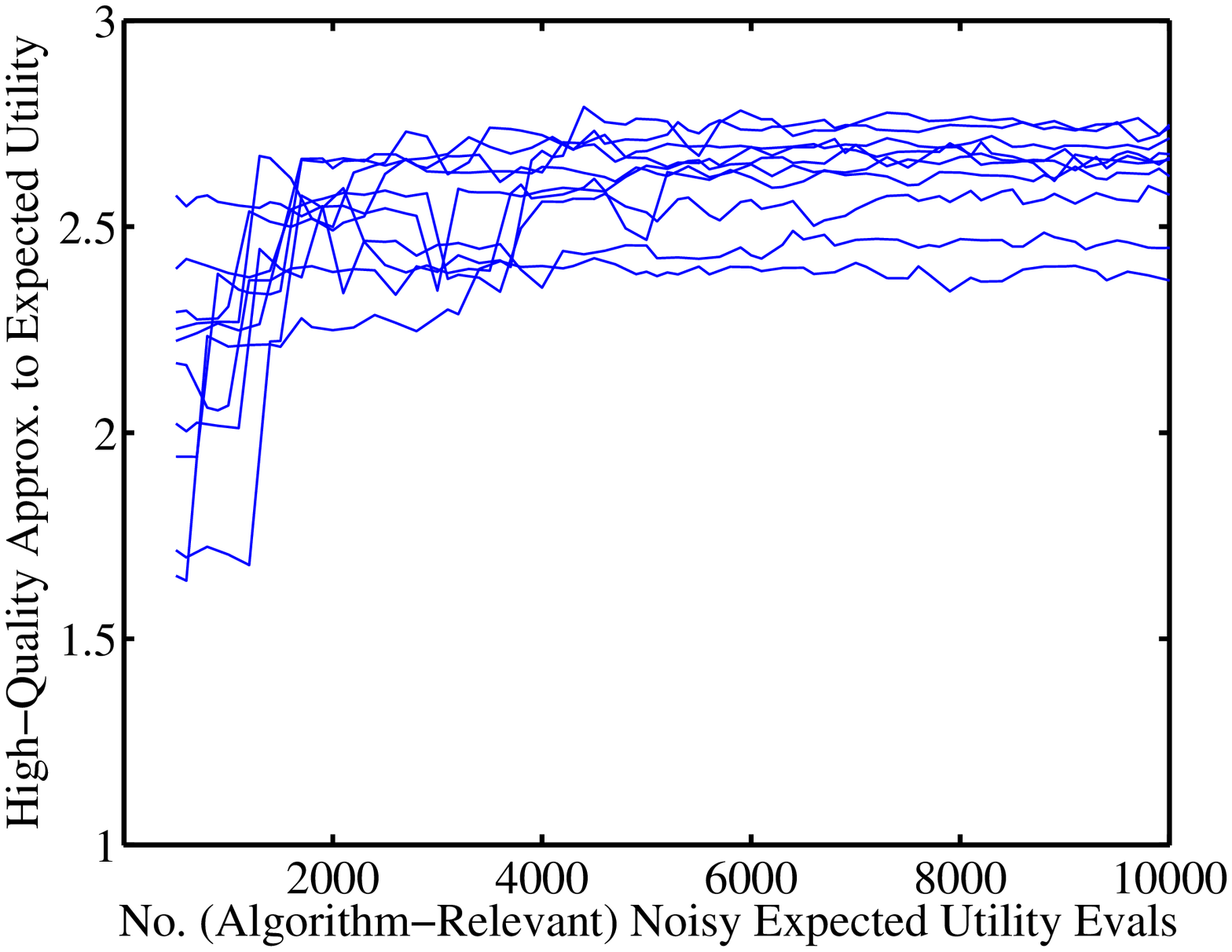}
  }
  \caption{\textit{High-quality} expected utility estimates, shown
    over the course of 10 independent stochastic optimization runs for
    the two-experiment combustion design. Each high-quality estimate is based on
    $n_{\mathrm{out}}=n_{\mathrm{in}}=10^4$ samples. Note that the
    output for NMNS $n_{\mathrm{out}}=1$ terminates earlier than the other
    cases because its simplices have already shrunk to sizes below
    machine precision. }
  \label{f:historyEUp12}
\end{figure}

Runs are performed for both SPSA and NMNS, with $n_{\mathrm{out}}$
ranging from 1 to 100.  The final design conditions and convergence
histories (the latter plotted for 10 runs only) are shown in
Figures~\ref{f:finalXp12} and~\ref{f:historyEUp12}, respectively. In
Figure~\ref{f:finalXp12}, each connected blue cross and red circle
represent a pair of final experimental designs, where the red circle
is arbitrarily chosen to represent the lower $T_0$ design. The
optimization results indicate that both experiments should be
performed near $T_0=975$ K, although the best $\phi$ is less precisely
determined and less influential. This pattern is similar to that of a
single-experiment optimal design.
Overall, a tighter clustering of the final design points is observed
as $n_{\mathrm{out}}$ is increased. SPSA groups the majority of the
final design points more tightly, but it also yields more outliers
than NMNS; in other words, it results in more of a ``hit-or-miss''
situation.  Figure~\ref{f:finalXp12} indicates that, for the NMNS
cases, a lower $n_{\mathrm{out}}$ lets the algorithm reach the
convergence ``plateau'' more quickly. This result is affected both by
the shrinkage rate of the simplex and by the fact that a higher
$n_{\mathrm{out}}$ simply requires more model evaluations even to
complete one iteration. The choice of $n_{\mathrm{out}}$ then should
take into consideration how many model evaluations are available. Even
then, the best choice may depend on the shape of the expected utility
surface and the variance of its estimator (which is not stationary in
$\bd$).

\begin{figure}[htb]
  \centering 
  \subfigure[SPSA $n_{\mathrm{out}}=1$]
  {
    \includegraphics[width=0.47\textwidth]{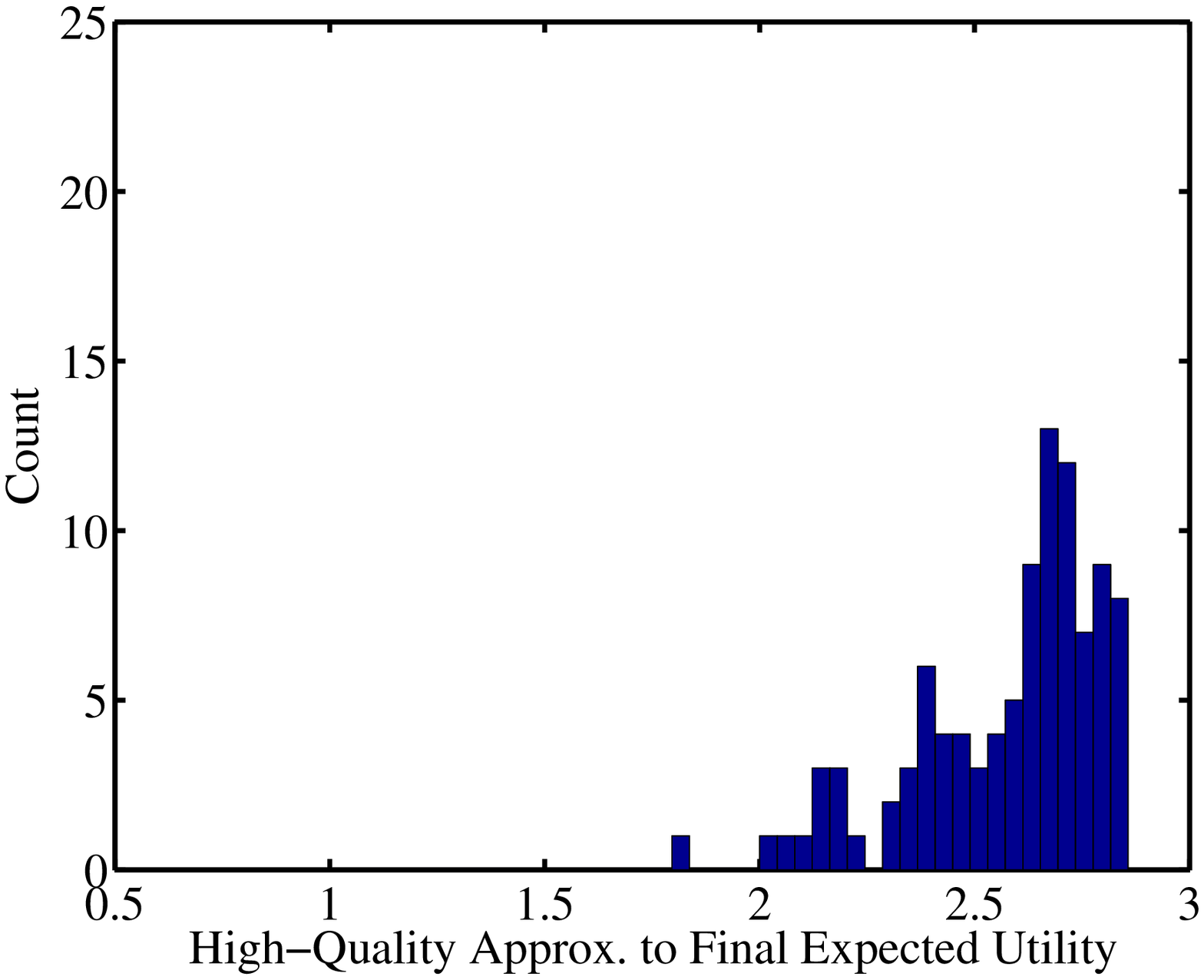}
  }
  \subfigure[NMNS $n_{\mathrm{out}}=1$]
  {
    \includegraphics[width=0.47\textwidth]{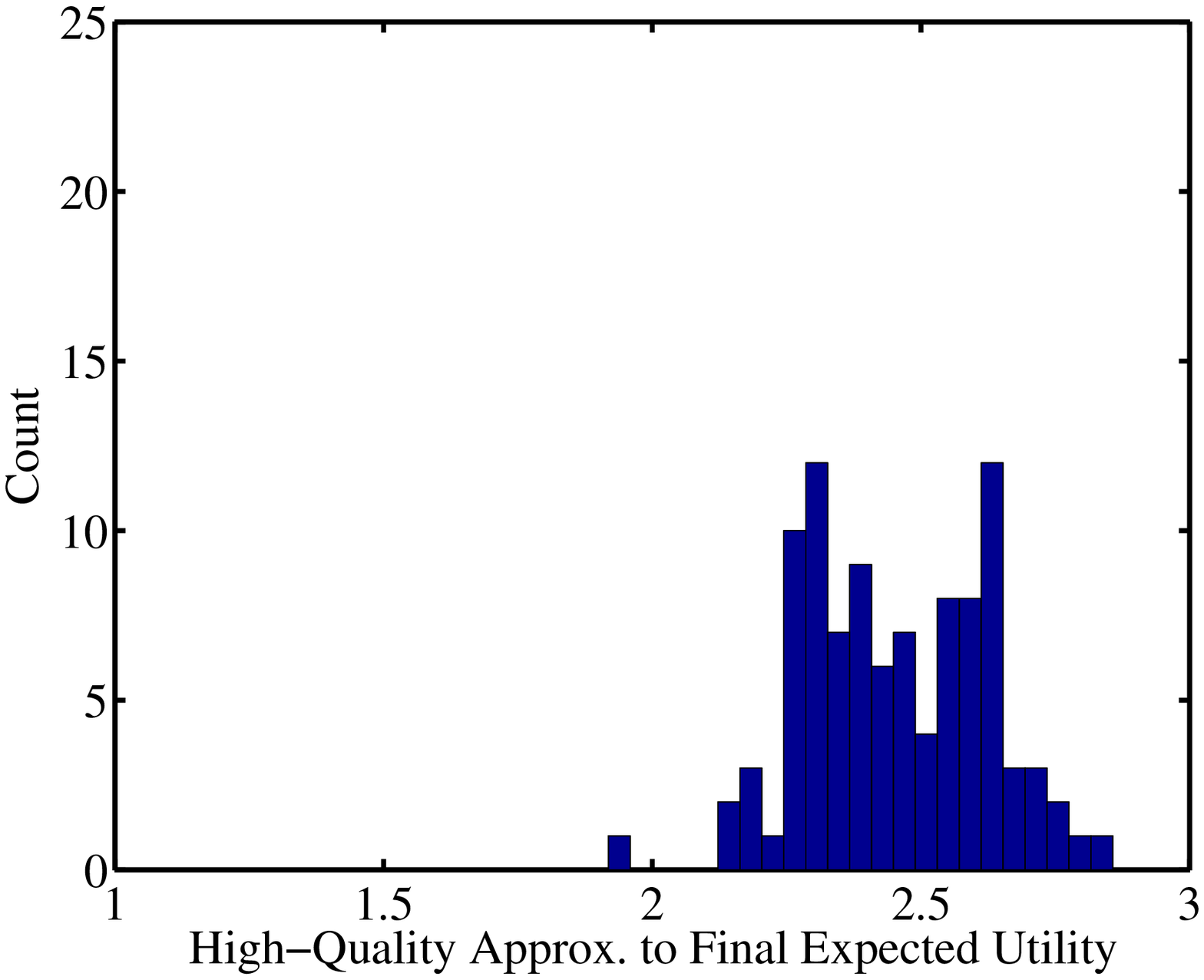}
  }
  \subfigure[SPSA $n_{\mathrm{out}}=10$]
  {
    \includegraphics[width=0.47\textwidth]{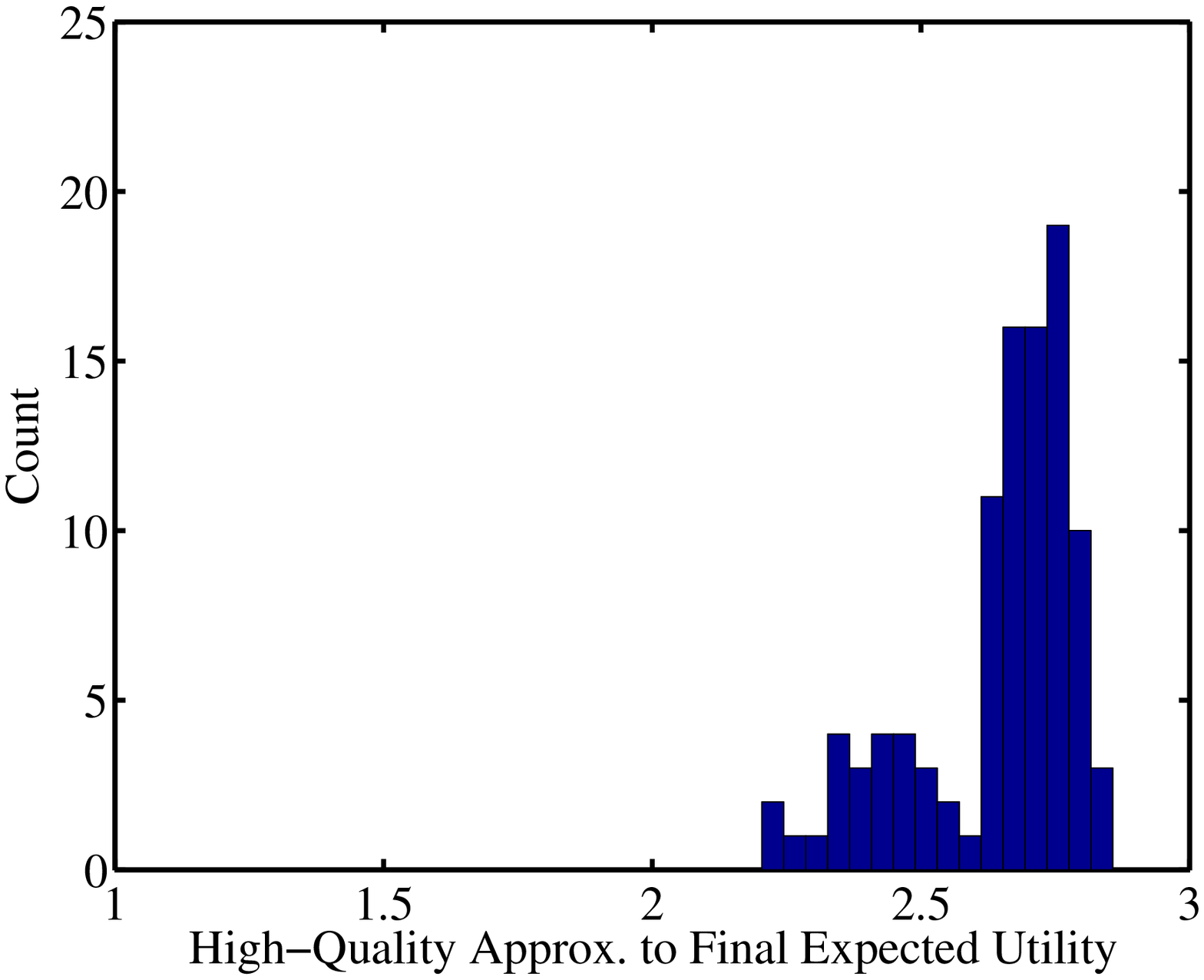}
  }
  \subfigure[NMNS $n_{\mathrm{out}}=10$]
  {
    \includegraphics[width=0.47\textwidth]{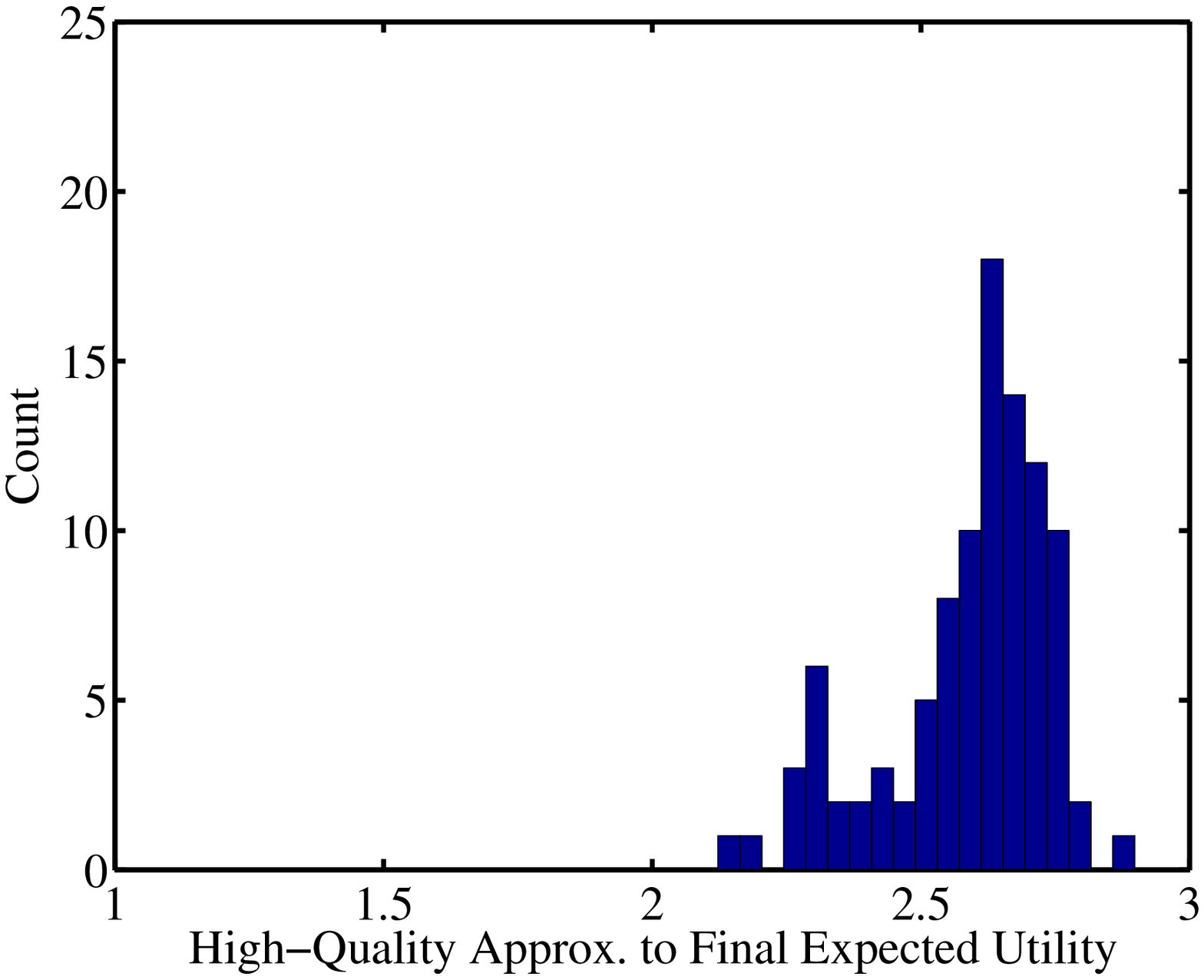}
  }
  \subfigure[SPSA $n_{\mathrm{out}}=100$]
  {
    \includegraphics[width=0.47\textwidth]{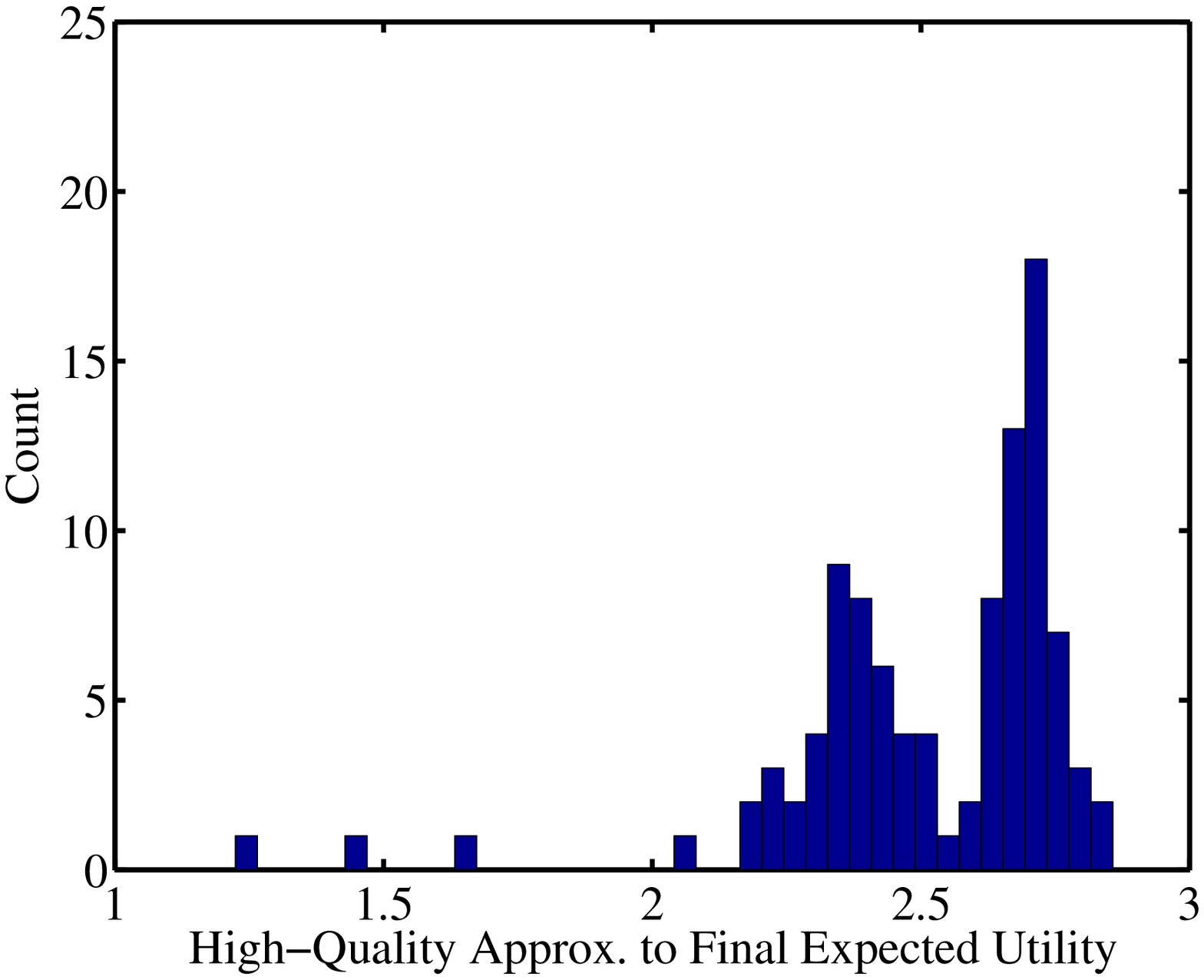}
  }
  \subfigure[NMNS $n_{\mathrm{out}}=100$]
  {
    \includegraphics[width=0.47\textwidth]{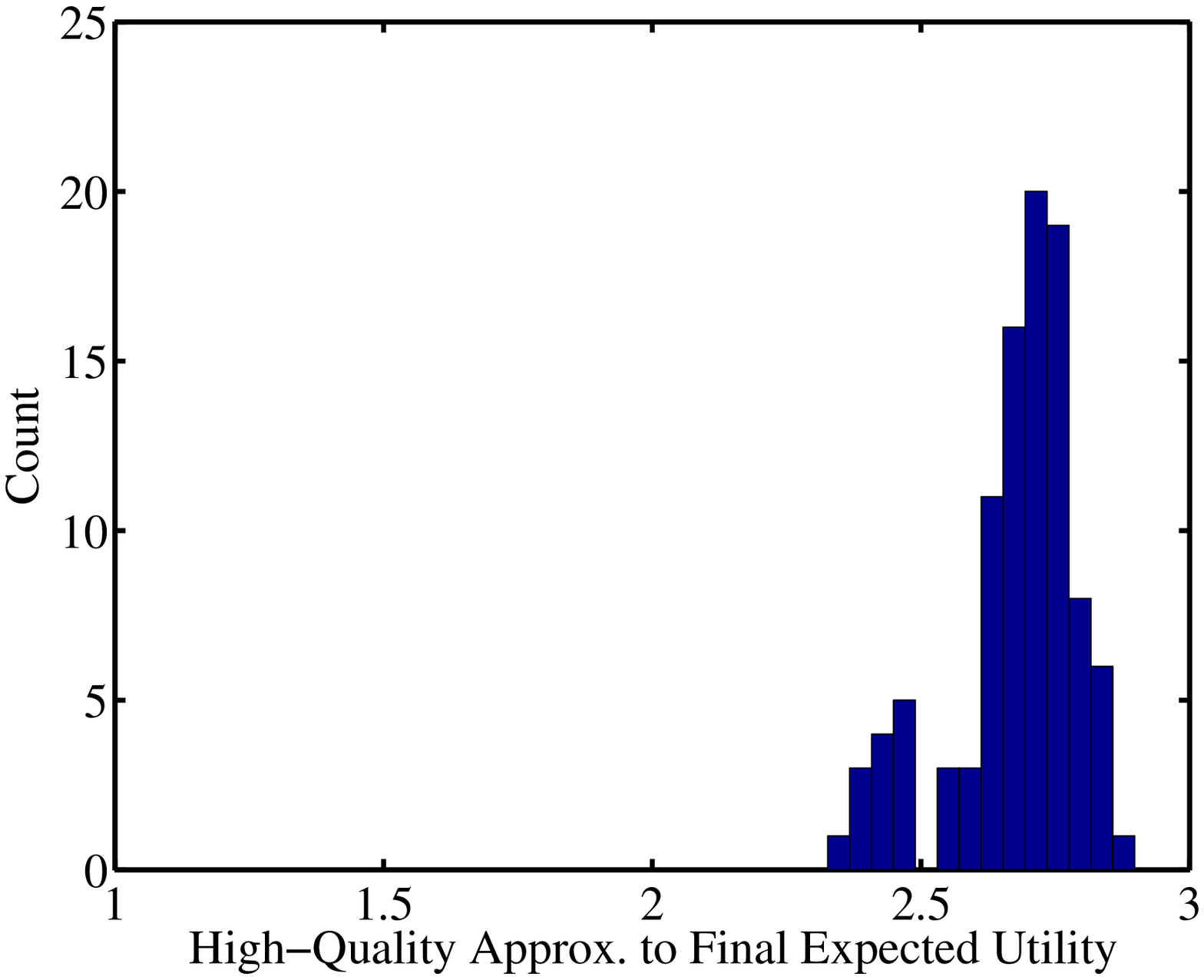}
  }
  \caption{\textit{High-quality} expected utility estimates
    corresponding to the optimization outputs in
    Figure~\ref{f:finalXp12}. Each high-quality estimate is based on
    $n_{\mathrm{out}}=n_{\mathrm{in}}=10^4$ samples.}
  \label{f:finalEUp12}
\end{figure}

Our observations so far are based on an assessment of the design
locations and convergence history. A more quantitative analysis should
focus on the expected utility value of the final designs, which is
what really matters in the end. Figure~\ref{f:finalEUp12} shows
histograms of expected utility for the 100 final design points
resulting from each optimization case. To compare the design points,
we want to make the error incurred in \textit{estimating}
$U(\bd^\ast)$ relatively negligible, and thus we employ a high-quality
estimator with $n_{\mathrm{out}}=n_{\mathrm{in}}=10^4$. This is not
the small-sample estimator used inside the optimization algorithms; it
is a more expensive estimator used afterwards, only for diagnostic
purposes. The histograms indicate that increasing $n_{\mathrm{out}}$
is actually not very effective for SPSA, as the persistence of
outliers creates a spread in the final values, supporting our
suspicion that too small a noise level may be bad for SPSA. On the other
hand, increasing $n_{\mathrm{out}}$ is effective for NMNS. The bimodal
structure of the histograms is due to the two groups: good designs on
the center ``ridge'' and outliers, with few designs having expected
utility values in between. Overall, NMNS performs better than SPSA in
this study, both in terms of the asymptotic distribution of design
parameters and how quickly the convergence plateau or ``knee'' is
reached.

To validate the results, the parameter inference problem is solved
with data from three two-experiment design points (labeled $D$, $E$,
and $F$) and with data from a four-experiment factorial design. All of
the experimental conditions are listed in
Table~\ref{t:inferenceConditions2}. Design $D$, a pair of experiments
lying on the ridge of ``good designs,'' is expected to have the
tightest posterior among the two-experiment designs. The posteriors
are shown in Figure~\ref{f:PCInference2}, using the PC surrogate
only. Indeed, design $D$ has the tightest posterior, and is much
better than the four-experiment factorial design even though it uses
fewer experiments! The factorial design blindly picks all the corners
in the design space, which are in general not good design points. (The
number of experiments in a factorial design would also increase
exponentially
with the number of design parameters, becoming impractical very
quickly.) Model-based optimal experimental design is far more robust
and capable than this traditional method.

\subsection{Is using a PC surrogate worthwhile?}
\label{ss:PCWorthIt}

\begin{figure}[htb]
  \centering
  \includegraphics[width=0.47\textwidth]{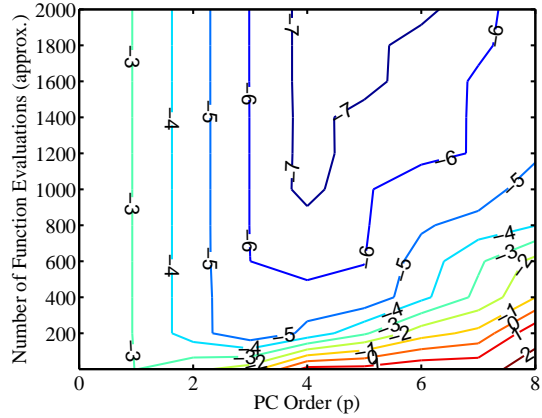}
  \caption{$\log_{10}$ of the $L^2$ error in the PC expansion for the
    peak heat release rate.}
  \label{f:L2ErrorsExample}
\end{figure}

In producing the previous results, we used a PC surrogate with $p=12$
and $n_{\mathrm{quad}}=10^6$, though analysis for one observable in
Figure~\ref{f:L2ErrorsExample} suggests that this polynomial
truncation and adapted quadrature rule are perhaps of higher quality
than necessary for our problem. To quantitatively determine whether a
polynomial surrogate offers efficiency gains in this study (or any
other), we must \textit{(i)} check if a lower quality PC surrogate may
be used while still achieving results of similar quality, and
\textit{(ii)} analyze the computational cost.

\begin{figure}[htb]
  \centering 
  \subfigure[Final positions]
  {
    \includegraphics[width=0.47\textwidth]{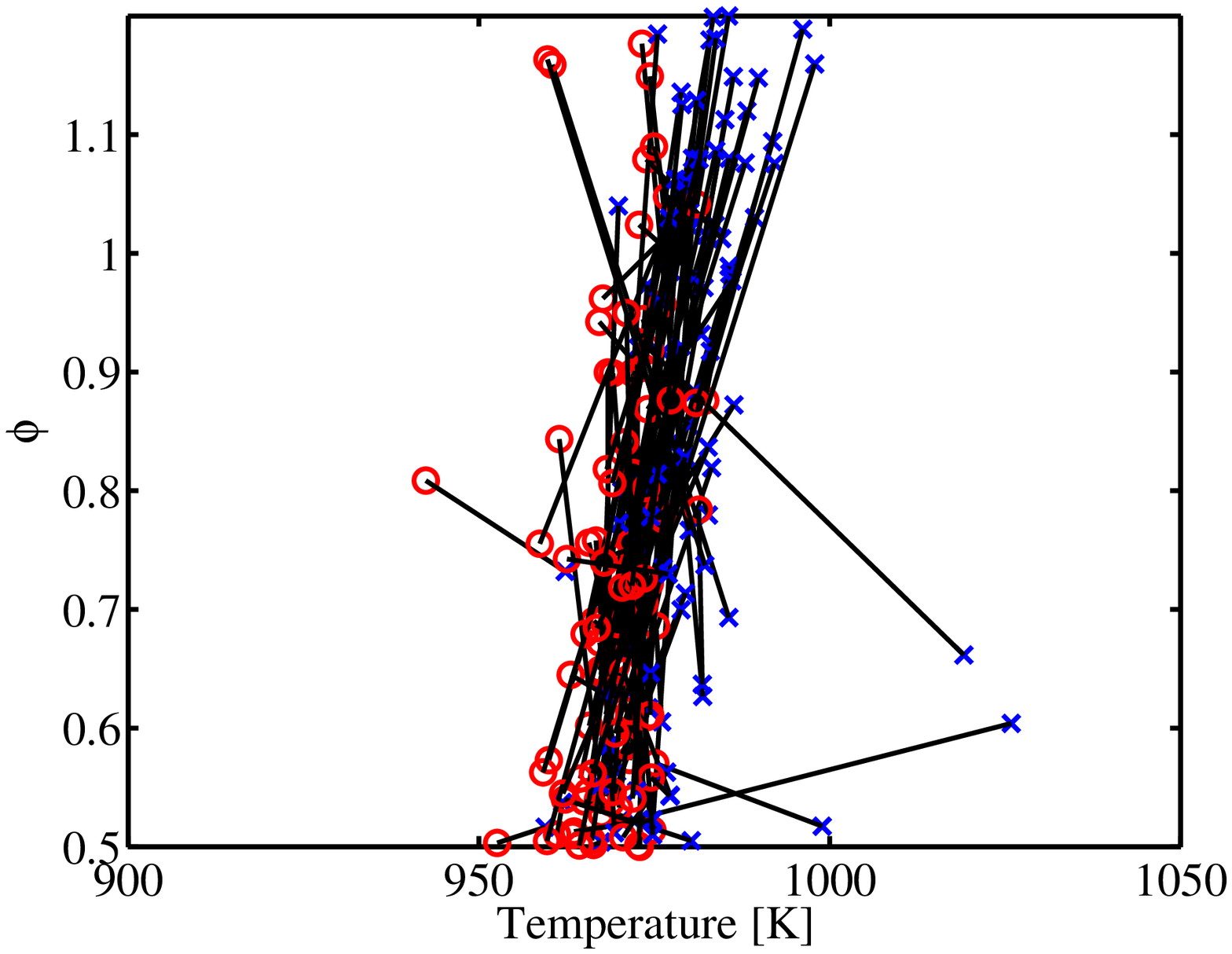}
  }
  \subfigure[Convergence history]
  {
    \includegraphics[width=0.47\textwidth]{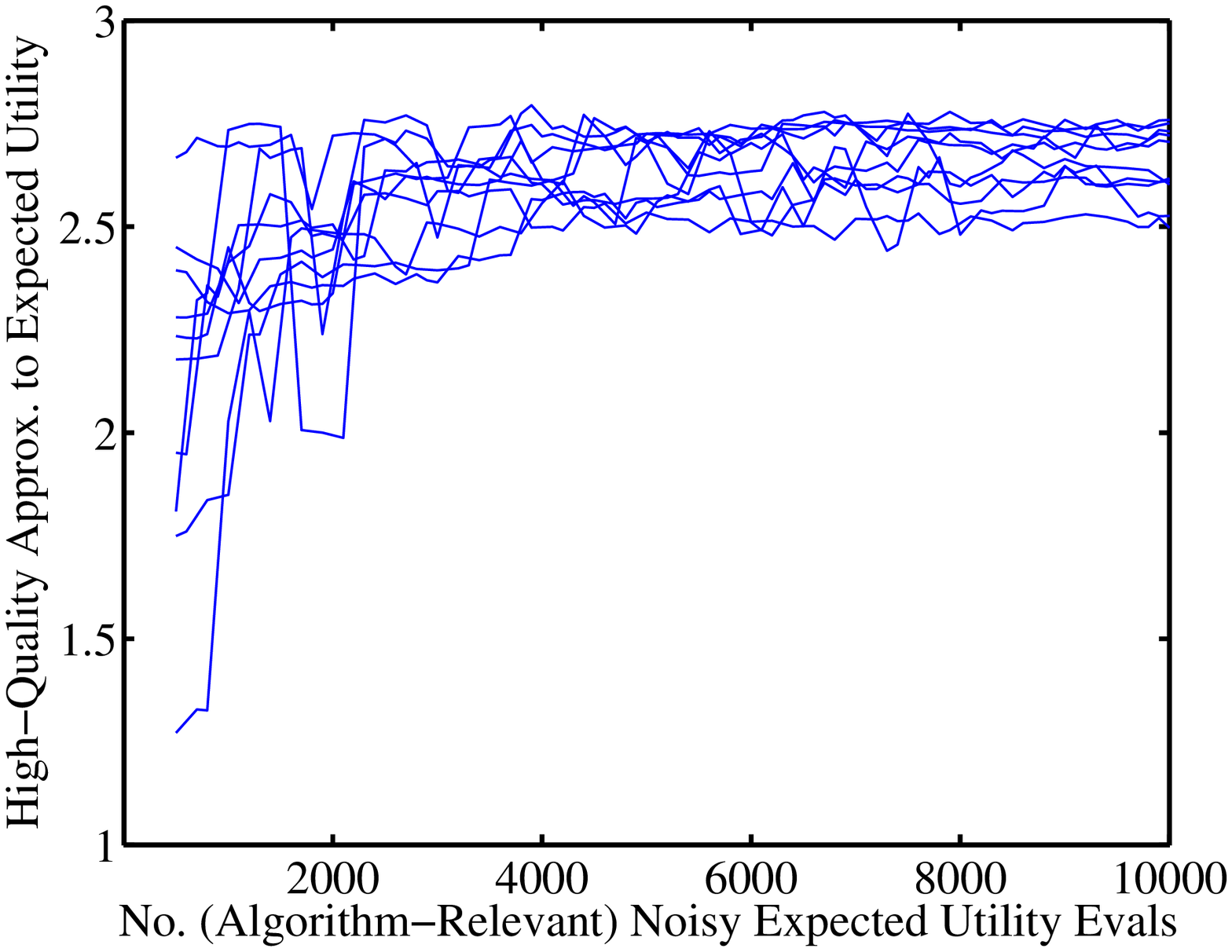}
  }
  \subfigure[Final expected utilities]
  {
    \includegraphics[width=0.47\textwidth]{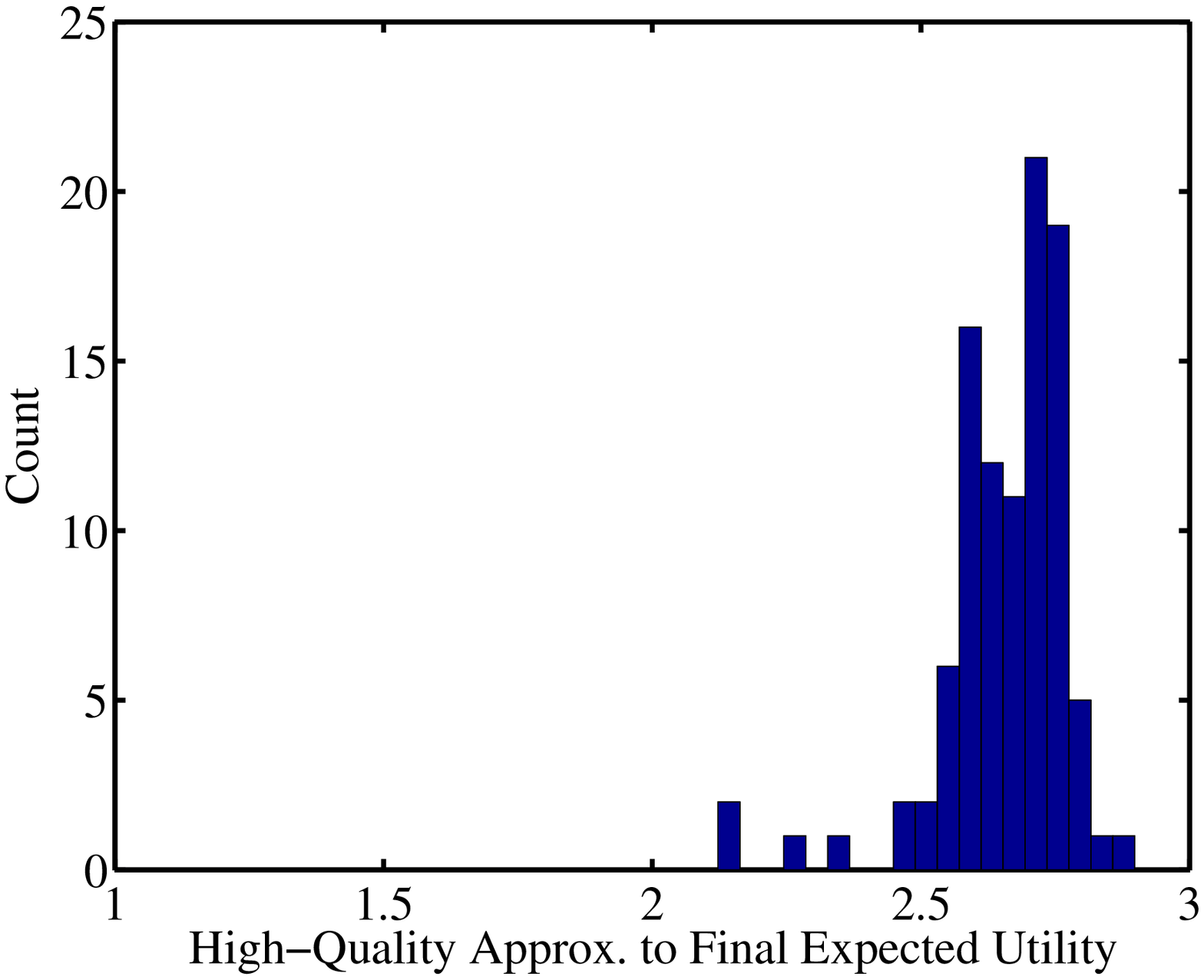}
  }
  \caption{Two-experiment combustion design problem: final outputs
    from 100 independent runs of stochastic optimization. Optimization
    is performed using only a \textit{lower-order} PC surrogate ($p=6$
    and $n_{\mathrm{quad}}=10^4$). To report performance, the expected
    utility estimates in subfigures (b) and (c) are computed using a
    \textit{higher-order} PC surrogate ($p=12$ and
    $n_{\mathrm{quad}}=10^6$) and
    $n_{\mathrm{out}}=n_{\mathrm{in}}=10^4$ samples. Only results with
    NMNS and $n_{\mathrm{out}}=100$ are shown. Compare to
    Figures~\ref{f:finalXp12}(f), \ref{f:historyEUp12}(f), and
    \ref{f:finalEUp12}(f).}
  \label{f:finalp6}
\end{figure}

To check if a lower-quality PC surrogate would suffice, the same
two-experiment optimal experimental design problem is solved but now
with $p=6$. We find that $n_{\mathrm{quad}}=10^4$ is roughly the
smallest $n_{\mathrm{quad}}$ that still yields reasonable results. The
final optimization positions (obtained via NMNS only), the convergence
history, and histogram of final expected utilities are shown in
Figure~\ref{f:finalp6}. In fact, the histogram appears to be even
tighter than that obtained with the $p=12$ surrogate.

To analyze the computational cost, we assume that optimization is
terminated after 5000 noisy objective function evaluations, as the
practical convergence plateau is reached by that point. If each noisy
objective function requires $10^4$ inner Monte Carlo samples and each
PC evaluation is negligible compared to full model, then using the
full ODE model requires $5000 \times 10^4 = 5 \times 10^7$ model
evaluations, whereas the construction of the PC surrogate requires
$10^4$ full ODE model evaluations. The surrogate thus provides a
speedup of roughly 3.5 orders of magnitude, saving 49,990,000 full
model evaluations (roughly 4 months of computational time for the
present problem, if run serially).

\begin{figure}[htb]
  \centering 
  \subfigure[Design $D$]
  {
    \includegraphics[width=0.47\textwidth]{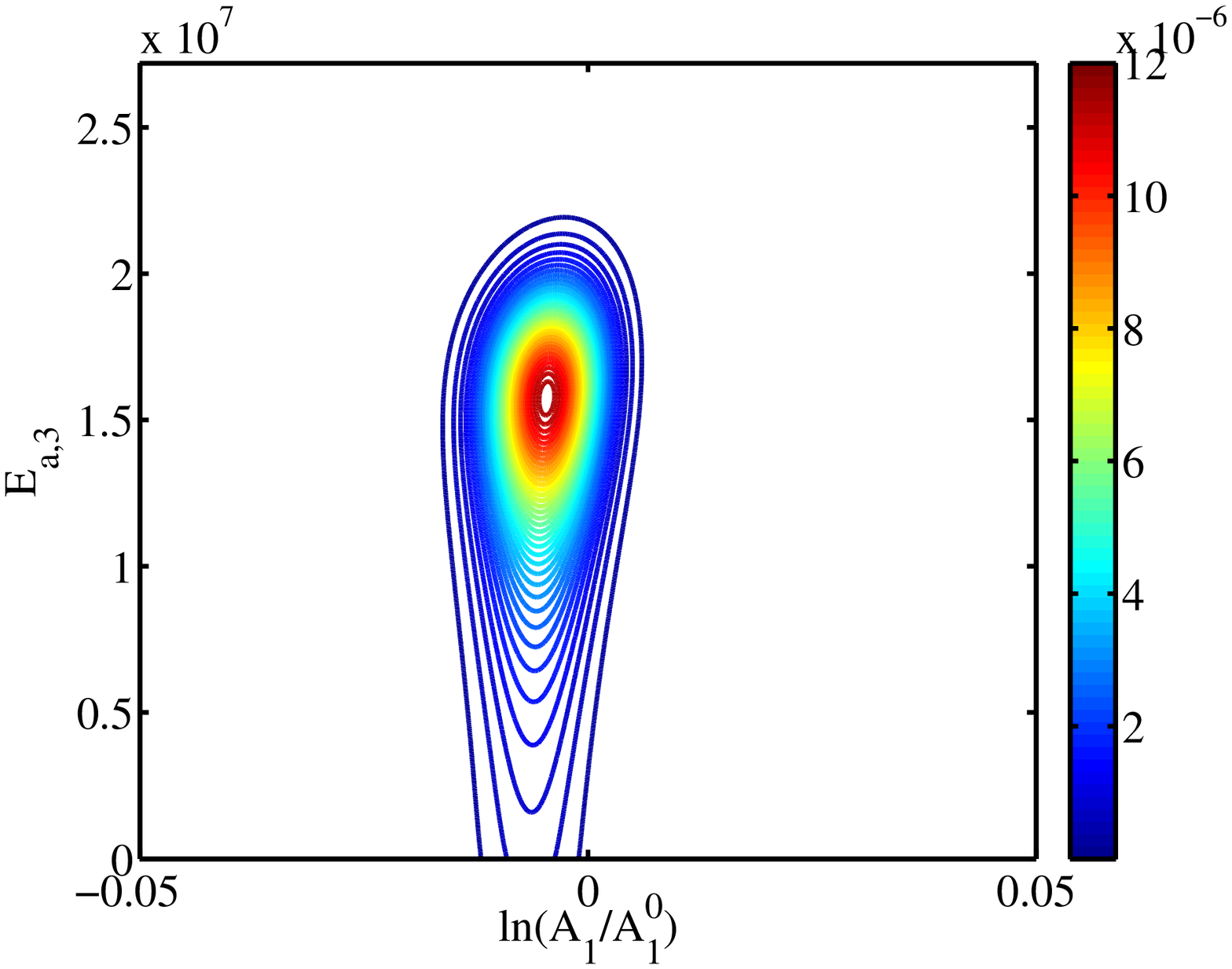}
  }
  \subfigure[Design $E$]
  {
    \includegraphics[width=0.47\textwidth]{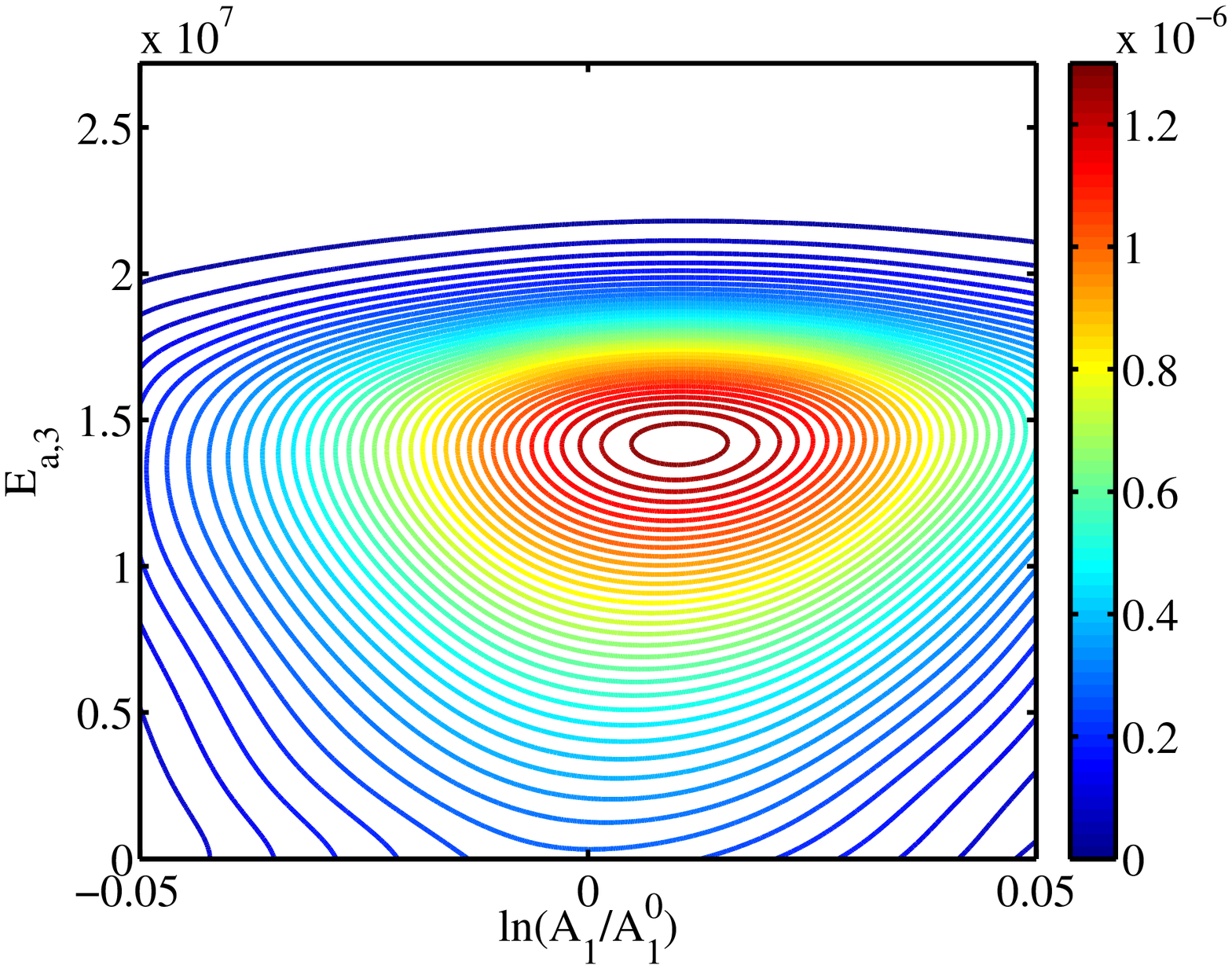}
  }
  \subfigure[Design $F$]
  {
    \includegraphics[width=0.47\textwidth]{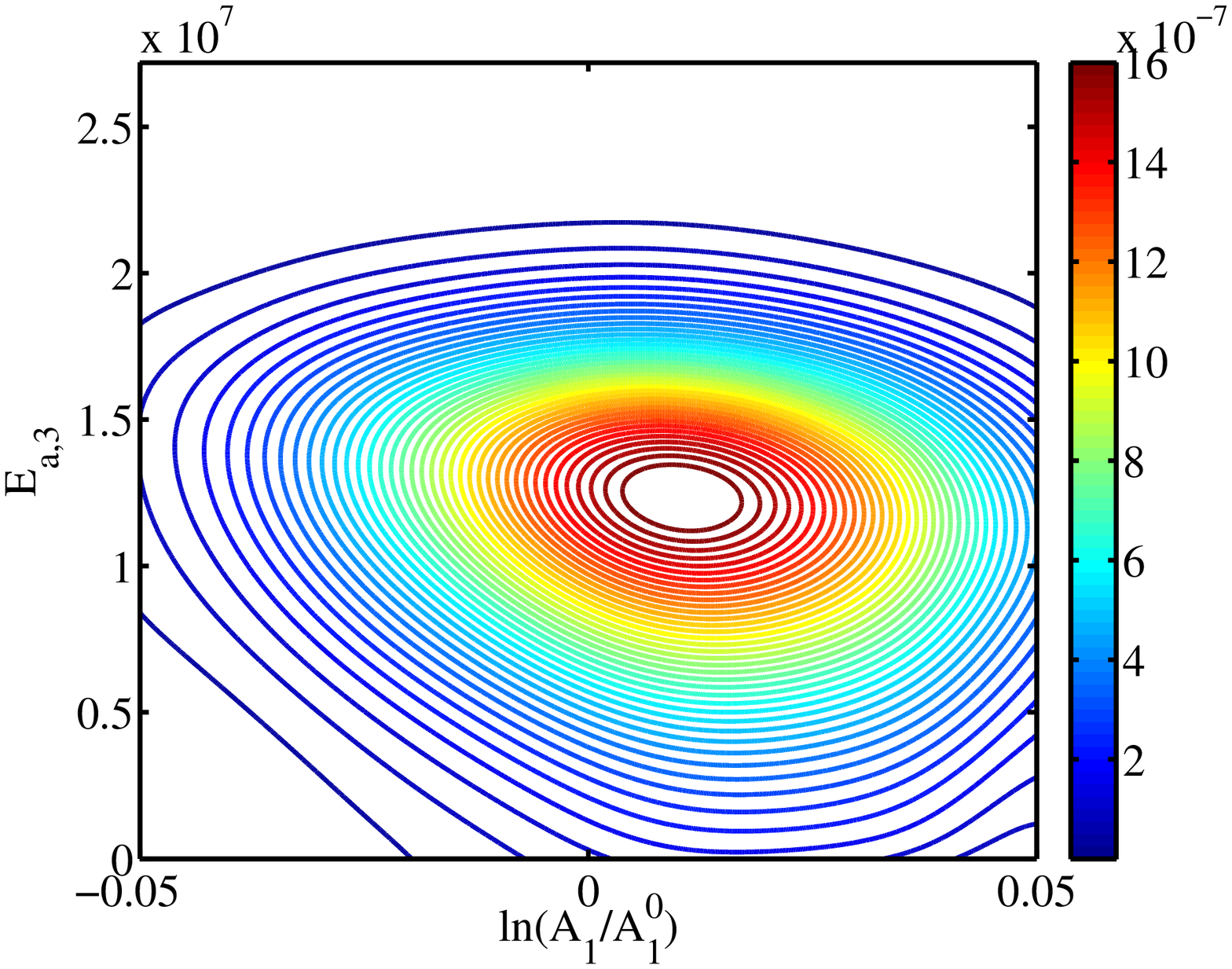}
  }
  \subfigure[Design ``factorial'']
  {
    \includegraphics[width=0.48\textwidth]{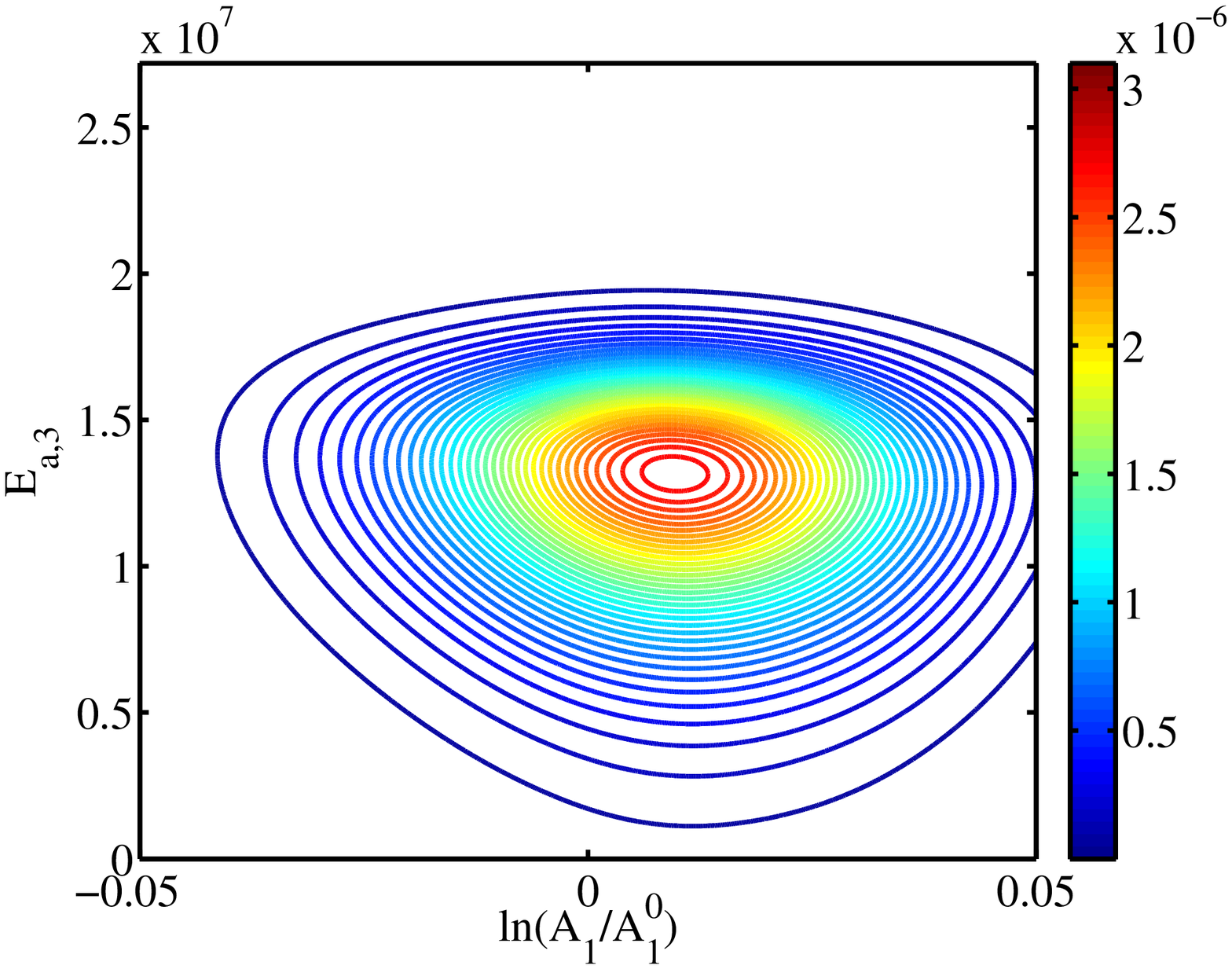}
  }
  \caption{Posterior densities resulting from inference at the
    experimental conditions listed in
    Table~\ref{t:inferenceConditions2}, using the PC surrogate with
    $p=12$, $n_{\textrm{quad}}=10^6$.}
  \label{f:PCInference2}
\end{figure}

\begin{figure}[htb]
  \centering 
  \subfigure[Design $D$]
  {
    \includegraphics[width=0.47\textwidth]{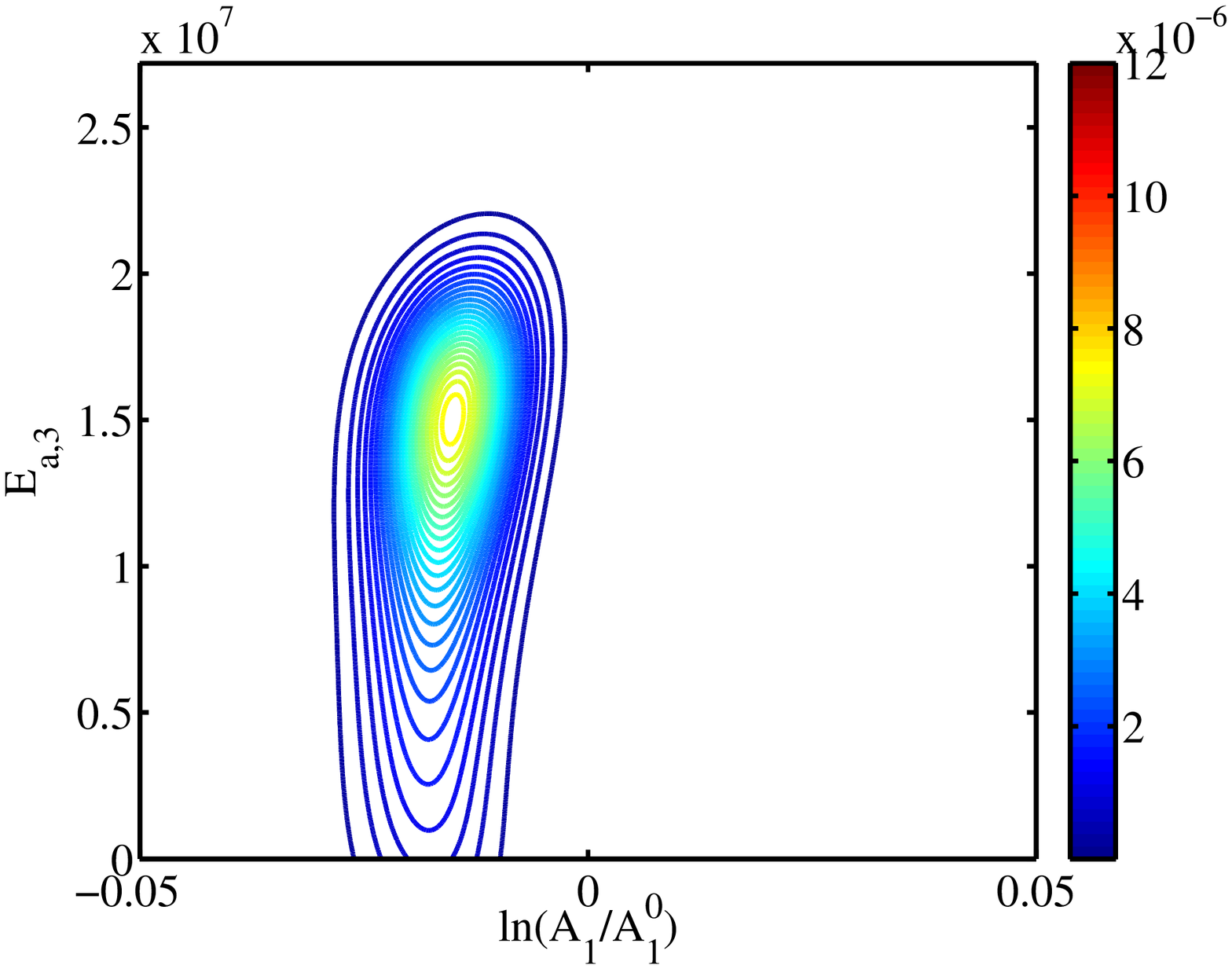}
  }
  \subfigure[Design $E$]
  {
    \includegraphics[width=0.47\textwidth]{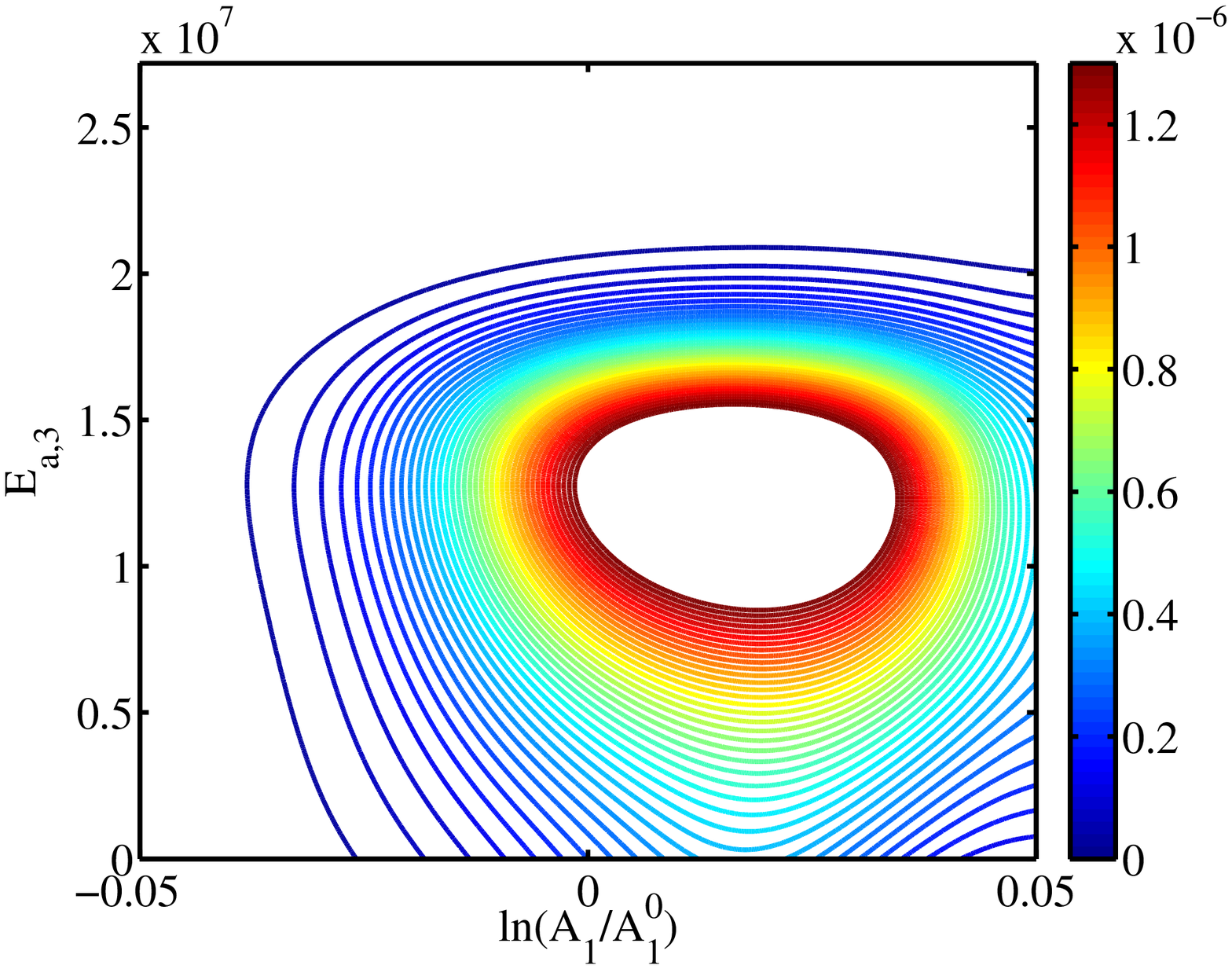}
  }
  \subfigure[Design $F$]
  {
    \includegraphics[width=0.47\textwidth]{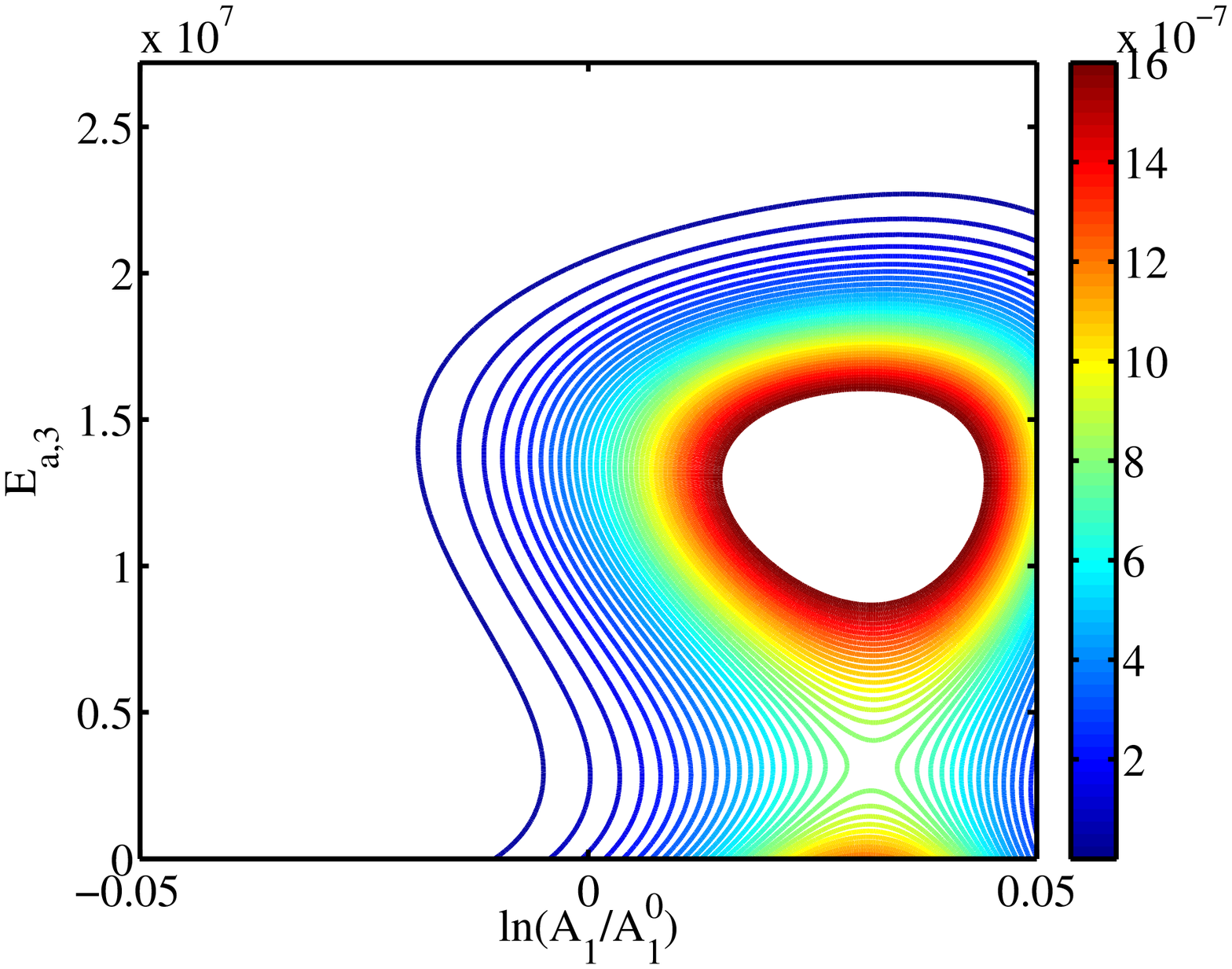}
  }
  \subfigure[Design ``factorial'']
  {
    \includegraphics[width=0.48\textwidth]{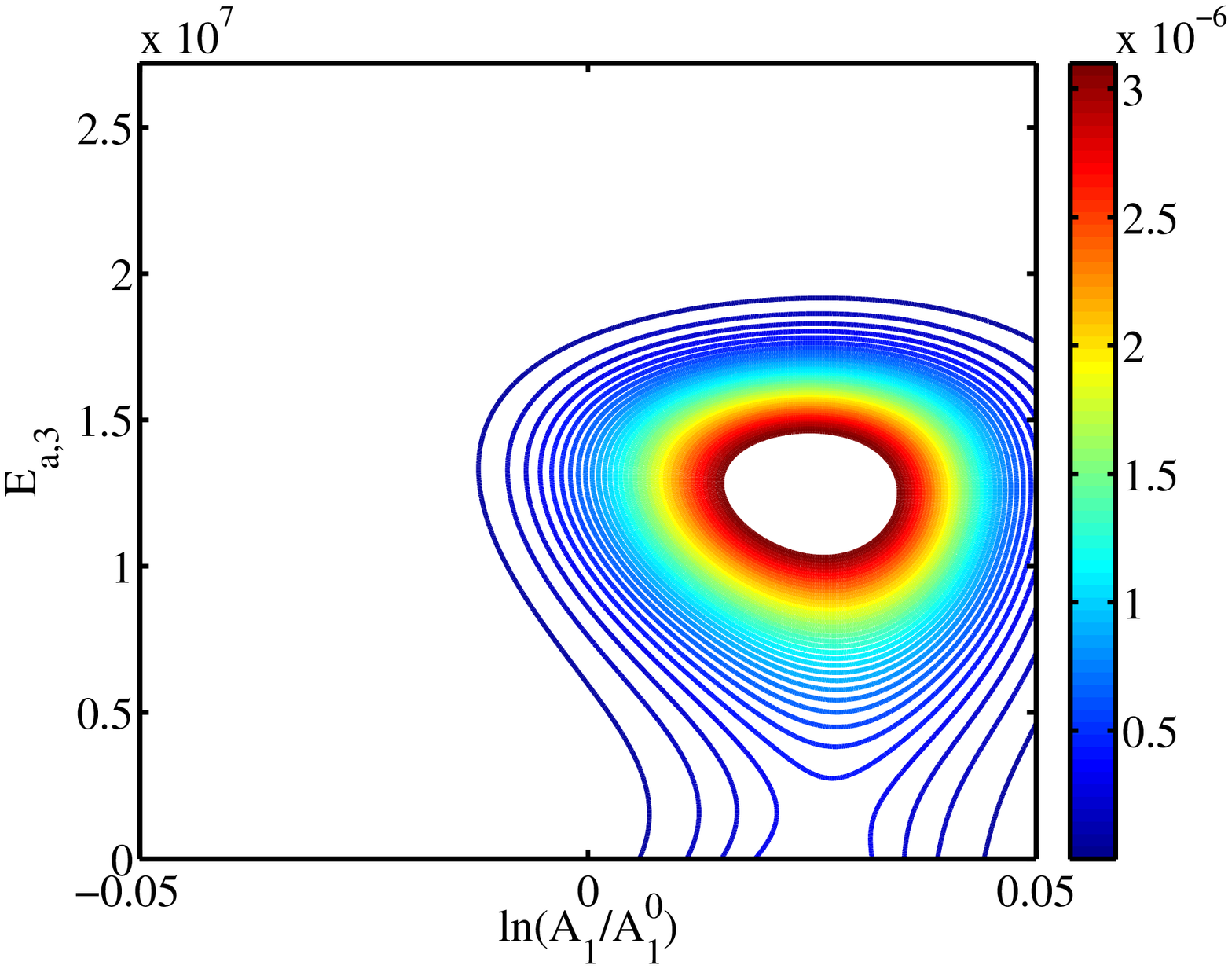}
  }
  \caption{Posterior densities resulting from inference at the
    experimental conditions listed in
    Table~\ref{t:inferenceConditions2}, using a \textit{lower-order PC
    surrogate} with $p=6$, $n_{\textrm{quad}}=10^4$. Compare to
    Figure~\ref{f:PCInference2}.}
  \label{f:PCInference2p6}
\end{figure}

Even though a low-quality surrogate may be sufficient for optimal
experimental design, it may not be sufficiently accurate for parameter
inference. For example, the two-experiment and four-experiment
posteriors obtained using the $p=6$, $n_{\mathrm{quad}}=10^4$
surrogate are shown in Figure~\ref{f:PCInference2p6}. These posterior
density contours show a substantial loss of accuracy compared to the
corresponding plots in Figure~\ref{f:PCInference2}. Because inference
does not involve averaging over the data space and broadly exploring
the design space, and because it generally favors a more restricted
range of the model parameters $\btheta$, it may be more sensitive to
local errors than the optimal experimental design formulation.
There are two possible solutions to this issue:
\begin{enumerate}
\item Build and use a high-order polynomial chaos surrogate at the
  outset of analysis, and use it for both optimal experimental design
  and inference. Because inference typically employs MCMC and thus
  requires many thousands or even millions of model evaluations, the
  combined computational savings will make such a surrogate almost
  certainly worth constructing.

\item A more efficient approach is to use a low-order polynomial chaos
  expansion to perform the optimal experimental design. Upon reaching
  the optimal design conditions, construct a new PC expansion for
  inference \textit{at that design only}. The new expansion does not
  need to capture dependence on the design variables, and thus it
  involves a smaller dimension and fewer interactions. This less
  expensive \textit{local} PC expansion can more easily be made
  sufficiently accurate for the inference problem.
\end{enumerate}

\section{Conclusions}
\label{s:conclusions}

\begin{figure}[htb]
  \centering
  \includegraphics[width=1.0\textwidth]{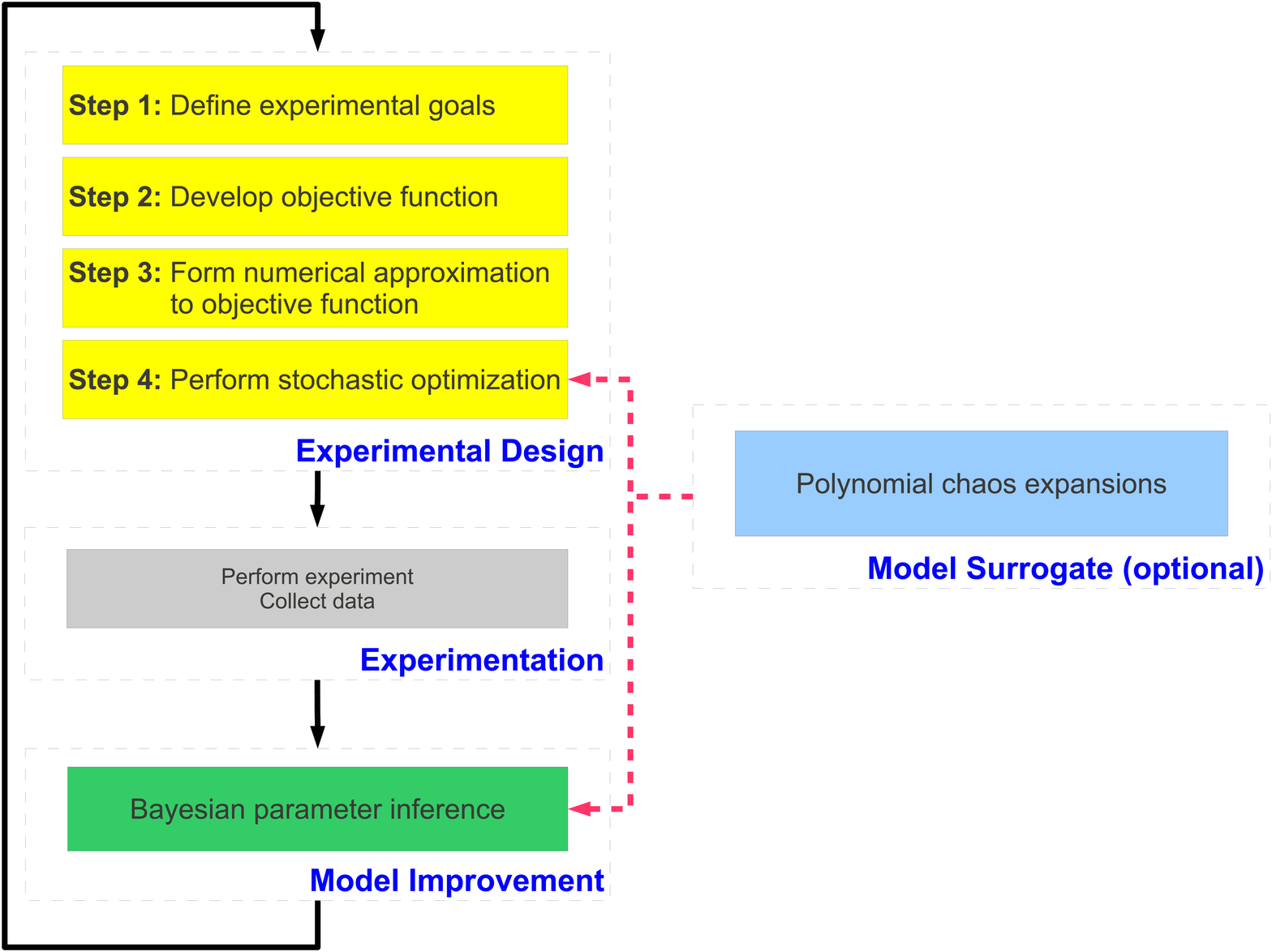}
  \caption{A flowchart summarizing the key steps of the
    design--experimentation--model improvement cycle.}
  \label{f:summary}
\end{figure}

This paper presents a systematic framework and a set of computational
tools for the optimal design of experiments. The framework can
incorporate nonlinear and computationally intensive models of the
experiment, linking them to rigorous information theoretic design
criteria and requiring essentially no simplifying assumptions. A
flowchart of the overall framework is given in Figure~\ref{f:summary},
showing the steps of optimal design and their role in a larger
design--experimentation--model improvement cycle. 

We focus on the experimental goal of parameter inference, in a
Bayesian statistical setting, where a good design criterion is shown
to maximize expected information gain from prior to posterior. A
two-stage Monte Carlo estimator of expected information gain is
coupled to algorithms for stochastic optimization.
The estimation of expected information gain, which would otherwise be
prohibitive with computationally intensive models, is substantially
accelerated through the use of flexible and accurate polynomial chaos
surrogate representations.

 
We demonstrate our method first on a simple nonlinear algebraic model,
then on shock tube ignition experiments described by a stiff system of
nonlinear ODEs. The latter system is challenging to approximate, as
certain model observables depend sharply on the combustion kinetic
parameters and design conditions, and ignition delays vary over
several orders of magnitude. In both these examples, we illustrate the
design of single and multiple experiments. We analyze the impact of
prior information on the optimal designs, and examine the selection of
observables according to their information value. We also
investigate numerical tradeoffs between variance in the estimator of
expected utility and performance of the stochastic optimization
schemes.

 
Overall we find that inference at optimal design conditions is indeed
very informative about the targeted parameters, and that model-based
optimal experiments are far more informative than those obtained with
simple heuristics. The use of surrogates offers significant
computational savings over stochastic optimization with the full
model, more than three orders of magnitude in the examples tested
here. Moreover, we find that the polynomial surrogate used in optimal
experimental design need not be extremely accurate in order to reveal
the correct design points; surrogate requirements for optimal design
are less stringent than for parameter inference.

 
Several promising avenues exist for future work. More efficient means
of constructing polynomial chaos expansions, adaptively and \textit{in
  conjunction with} stochastic optimization, may offer considerable
computational gains. Uncertainty in the design parameters themselves
can also be incorporated into the framework, as in real-world
experiments where the design conditions cannot be precisely imposed;
this additional uncertainty could be treated with a hierarchical
Bayesian approach. Structural inadequacy of the model $\bG$ is another
important issue; how successful is an ``optimal'' design (or indeed an
inference process) based on a forward model that cannot fully resolve
the physical processes at hand? Our current experience on design with
lower-order polynomial surrogates provides a glimpse into issues of
structural uncertainty, but a much more thorough exploration is
needed. Finally, the Bayesian optimal design methodology has a natural
extension to sequential experimental design, where one set of
experiments can be performed and analyzed before designing the next
set. A rigorous approach to sequential design, incorporating ideas
from dynamic programming and perhaps sequential Monte Carlo, may be
quite effective.


\section*{Acknowledgments}

The authors would like to acknowledge support from the KAUST Global
Research Partnership and from the US Department of Energy, Office of
Science, Office of Advanced Scientific Computing Research (ASCR) under
Grant No.\ DE-SC0003908.

\bibliographystyle{model1-num-names}
\bibliography{referencesAll}

\cleardoublepage

\begin{algorithm}[htb]
  $\bii = (1,\cdots,1)$\;
  $\mathcal{O} = \emptyset$\;
  $\mathcal{A} = \{\bii\}$\;
  \For{$r=1$ \KwTo $n_{\mathrm{coef}}$}{
    $v_r = \Delta_{\bii}f_r$\;
    Compute $h_{r,\bii}$\;
  }
  $\eta = \barh_{\bii}$. For example, $\barh_{\bii} = \max_r h_{r,\bii}$\;
  \While{$\eta>$ TOL}{
    select $\bii$ from $\mathcal{A}$ with the largest $\barh_{\bii}$\;
    $\mathcal{A} = \mathcal{A} \setminus \{\bii\}$\;
    $\mathcal{O} = \mathcal{O} \cup \{\bii\}$\;
    $\eta = \eta-\barh_{\bii}$\;
    \For{$p=1$ \KwTo $d$}{
      $\bj = \bii+\mb{e}_p$\;
      \If{$\mb{j}-\mb{e}_q\in \mathcal{O}$ for all $q=1,\cdots,d$}{
        $\mathcal{A} = \mathcal{A} \cup \{\bj\}$\;
        \For{$r=1$ \KwTo $n_{\mathrm{coef}}$}{
          $s_r = \Delta_{\bj}f_r$\;
          $v_r = v_r+s_r$\;
          Compute $h_{r,\bj}$\;
        }
        Compute $\barh_{\bj}$\;
        $\eta = \eta+\barh_{\bj}$\;
      }
    }
  }
  Symbols:\\
  $\mathcal{O}$---old index set\;
  $\mathcal{A}$---active index set\;
  $v_r$---computed integral value
  $\sum_{\bii\in\mathcal{O}\cup\mathcal{A}}\otimes_{p=1}^d
  \Delta_{i_p}f_r$ for the $r$th coefficient\;
  $\Delta_{\bii}f_r$---integral increment $\otimes_{p=1}^d
  \Delta_{i_p}f_r$ for the $r$th coefficient\;
  $h_{r,\bii}$---local error indicator for the $r$th coefficient\;
  $\barh_{\bii}$---total effect local error indicator\;
  $\eta$---global error estimate $\sum_{\bii\in\mathcal{A}}
  \barh_{\bii}$\;
  $TOL$---error tolerance\;
  $\mb{e}_p$---$p$th unit vector\;
  \caption{Modified dimension-adaptive sparse quadrature algorithm for
    non-intrusive spectral projection.}
  \label{a:DASQinNISP}
\end{algorithm}

\cleardoublepage

\begin{figure}[htb]
  \centering
  \subfigure[Prior $\theta\sim \CU(0,1)$]
  {
    \includegraphics[width=0.47\textwidth]{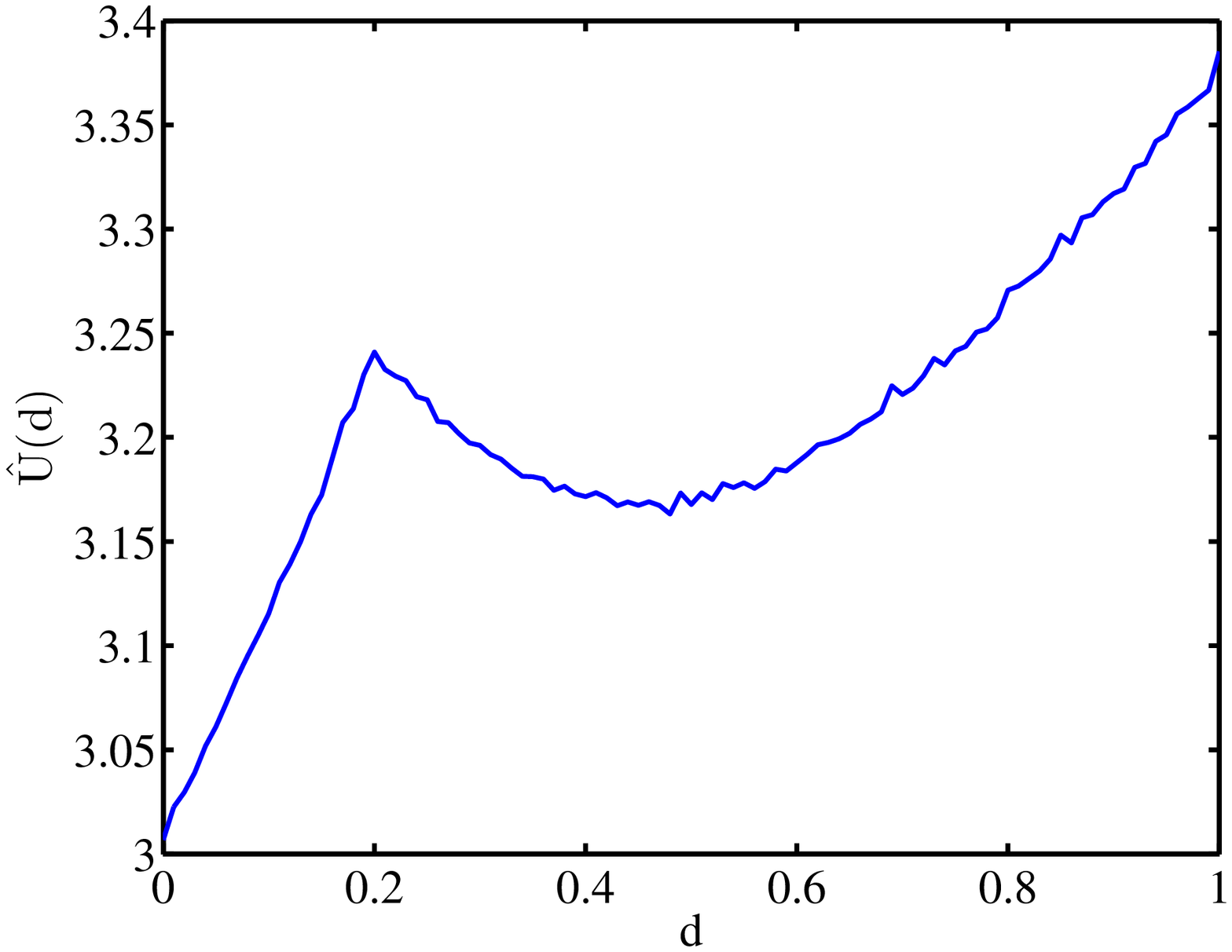}
    \label{f:simpleDesign1Exp_theta0to1}
  }\\
  \subfigure[Prior $\theta\sim \CU(0,\theta_e)$]
  {
    \includegraphics[width=0.47\textwidth]{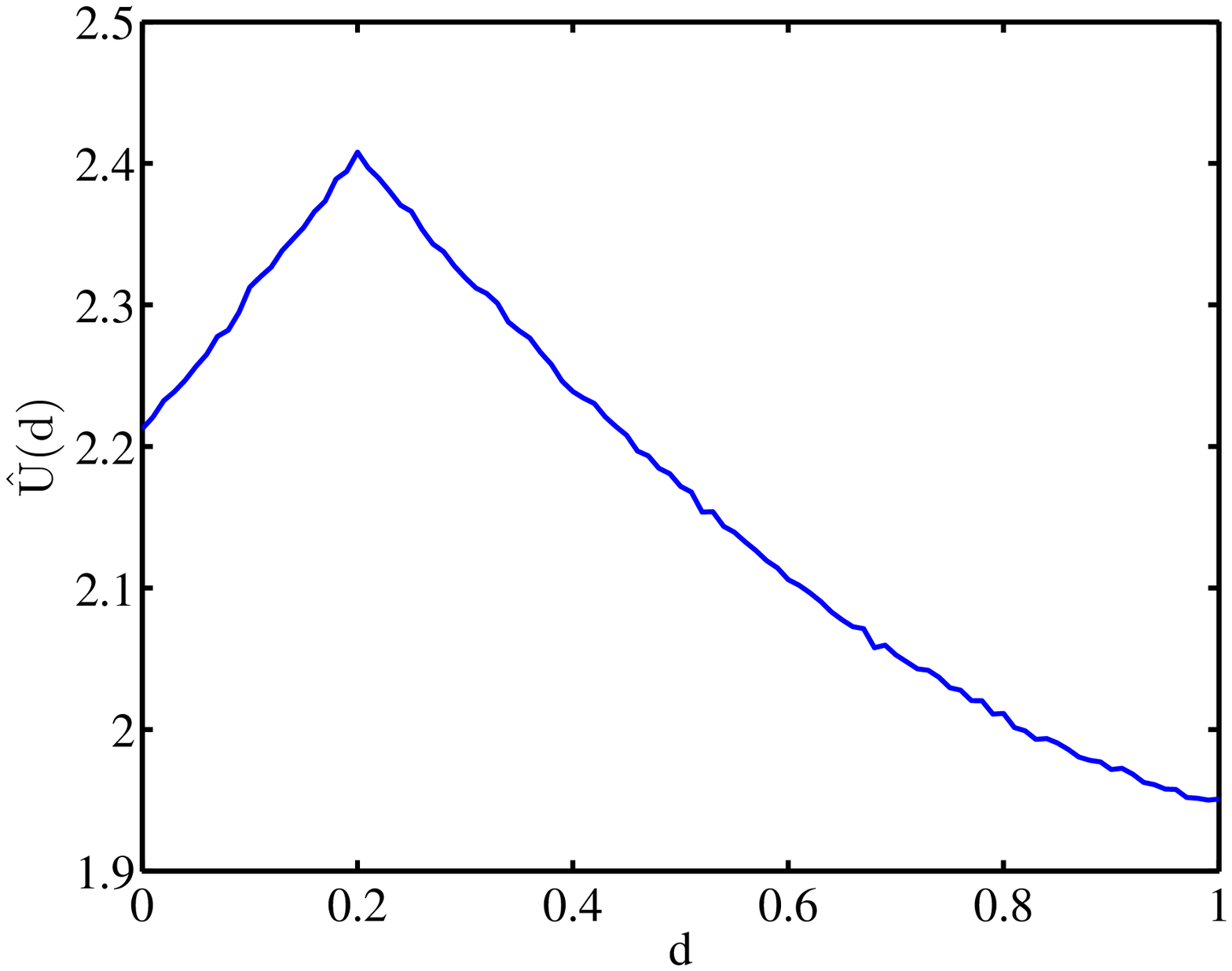}
    \label{f:simpleDesign1Exp_theta0to0_4373}
  }
  \subfigure[Prior $\theta\sim \CU(\theta_e,1)$]
  {
    \includegraphics[width=0.47\textwidth]{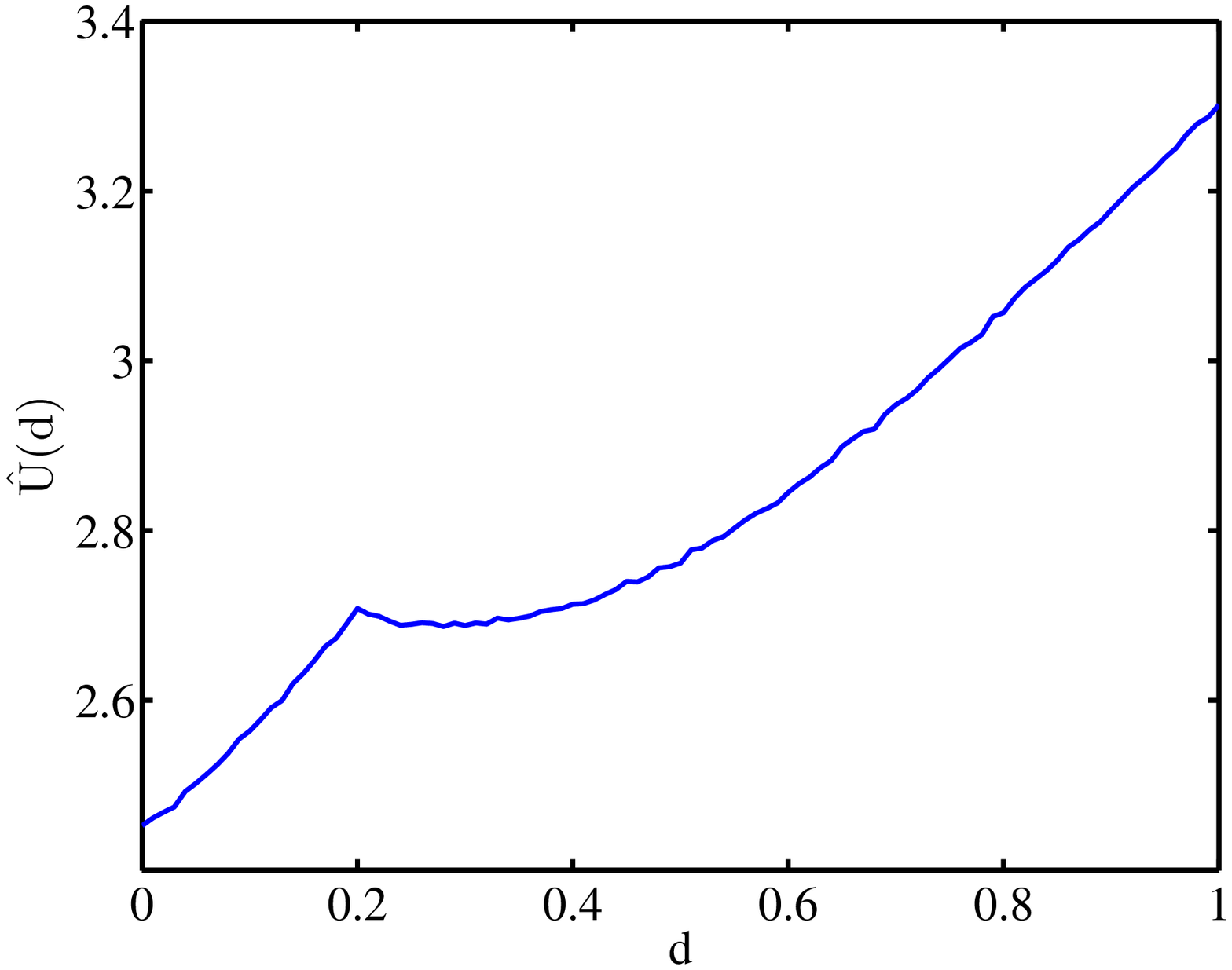}
    \label{f:simpleDesign1Exp_theta0_4373to1}
  }
  \caption{Estimated expected utility for the design of a single
    experiment with the simple nonlinear model under different priors,
    where $\theta_e=\sqrt{\frac{1-e^{-0.8}}{2.88}} \approx 0.4373$.}
  \label{f:simpleDesign1Exp}
\end{figure}

\begin{table}[htb]
  \centering
  \begin{tabular}{c|ccc}
    Reaction No. & \multicolumn{3}{c}{Elementary Reaction} \\\hline
    \textbf{R1} & $\mathbf{H + O_2}$ &$\mathbf{\Longleftrightarrow}$& $\mathbf{O + OH}$ \\
    R2 & $O + H_2$ &$\Longleftrightarrow$& $H + OH$ \\
    \textbf{R3} & $\mathbf{H_2 + OH}$ &$\mathbf{\Longleftrightarrow}$& $\mathbf{H_2O + H}$ \\
    R4 & $OH + OH$ &$\Longleftrightarrow$& $O + H_2O$ \\
    R5 & $H_2 + M$ &$\Longleftrightarrow$& $H + H + M$ \\
    R6 & $O + O + M$ &$\Longleftrightarrow$& $O_2 + M$ \\
    R7 & $O + H + M$ &$\Longleftrightarrow$& $OH + M$ \\
    R8 & $H + OH + M$ &$\Longleftrightarrow$& $H_2O + M$ \\
    R9 & $H + O_2 + M$ &$\Longleftrightarrow$& $HO_2 + M$ \\
    R10 & $HO_2 + H$ &$\Longleftrightarrow$& $H_2 + O_2$ \\
    R11 & $HO_2 + H$ &$\Longleftrightarrow$& $OH + OH$ \\
    R12 & $HO_2 + O$ &$\Longleftrightarrow$& $O_2 + OH$ \\
    R13 & $HO_2 + OH$ &$\Longleftrightarrow$& $H_2O + O_2$ \\
    R14 & $HO_2 + HO_2$ &$\Longleftrightarrow$& $H_2O_2 + O_2$ \\
    R15 & $H_2O_2 + M$ &$\Longleftrightarrow$& $OH + OH + M$ \\
    R16 & $H_2O_2 + H$ &$\Longleftrightarrow$& $H_2O + OH$ \\
    R17 & $H_2O_2 + H$ &$\Longleftrightarrow$& $HO_2 + H_2$ \\
    R18 & $H_2O_2 + O$ &$\Longleftrightarrow$& $OH + HO_2$ \\
    R19 & $H_2O_2 + OH$ &$\Longleftrightarrow$& $HO_2 + H_2O$
  \end{tabular}
  \caption{19-reaction hydrogen-oxygen
    mechanism~\cite{yetter:1991:acr}. Reactions involving the wildcard
    species $M$ are three-body interactions, with different
    efficiencies for different species. Kinetic parameters of the
    boldface reactions are targeted for inference.}
  \label{t:hoMech}
\end{table}

\begin{table}[htb]
  \centering
  \begin{tabular}{c|c}
    Observable & Explanation \\\hline
    $\tau_{ign}$ & Ignition delay, defined as the time of peak enthalpy
    release rate \\
    $\tau_{O}$ & Characteristic time in which peak $X_{O}$
    occurs \\
    $\tau_{H}$ & Characteristic time in which peak $X_{H}$
    occurs \\
    $\tau_{HO_2}$ & Characteristic time in which peak $X_{HO_2}$
    occurs \\
    $\tau_{H_2O_2}$ & Characteristic time in which peak $X_{H_2O_2}$
    occurs \\\hline
    $\left.\ddt{h}\right|_{\tau}$ & Peak value of enthalpy release rate \\
    $X_{O,\tau}$ & Peak value of $X_{O}$ \\
    $X_{H,\tau}$ & Peak value of $X_{H}$ \\
    $X_{HO_2,\tau}$ & Peak value of $X_{HO_2}$ \\
    $X_{H_2O_2,\tau}$ & Peak value of $X_{H_2O_2}$
  \end{tabular}
  \caption{Selected observables for the combustion problem. Note that
    $dh/dt<0$ when enthalpy is released or lost by the system. }
  \label{t:obs}
\end{table}

\begin{table}[htb]
\centering
\begin{tabular}{c|cc}
  Parameter & Lower Bound & Upper Bound \\ \hline
  $\ln(A_1/A_1^0)$ & $-0.05$ & 0.05 \\
  $E_{a,3}$ & 0 & $2.7196\times10^{7}$ \\
  $T_0$ & 900 & 1050 \\
  $\phi$ & 0.5 & 1.2 
\end{tabular}
\caption{Prior support of the uncertain kinetic parameters
  $\ln(A_1/A_1^0)$ and $E_{a,3}$, and ranges of the design variables
  $T_0$ and $\phi$. A uniform prior is assigned to the kinetic
  parameters. }
\label{t:priorSupport}
\end{table}

\begin{table}[htb]
\centering
\begin{tabular}{c|cc}
  Design Point & $T_0$ & $\phi$ \\ \hline
  $A$ & 975 & 0.5 \\
  $B$ & 925 & 0.85 \\
  $C$ & 1025 & 0.85
\end{tabular}
\caption{Experimental conditions at design points $A$, $B$, and $C$.}
\label{t:inferenceConditions}
\end{table}

\clearpage

\begin{table}[htb]
\centering
\begin{tabular}{c|cccccccc}
  Design Point & $T_{0,1}$ & $\phi_1$ & $T_{0,2}$ & $\phi_2$ & $T_{0,3}$ & $\phi_3$ & $T_{0,4}$ & $\phi_4$\\ \hline
  $D$ & 970 & 0.6 & 975 & 1.1 & -- & -- & -- & -- \\
  $E$ & 900 & 0.5 & 1050 & 1.2 & -- & -- & -- & -- \\
  $F$ & 900 & 1.2 & 1050 & 0.5 & -- & -- & -- & -- \\ \hline
  ``factorial'' & 900 & 0.5 & 900 & 1.2 & 1050 & 0.5 & 1050 & 1.2
\end{tabular}
\caption{Experimental conditions at design points $D$, $E$, $F$, and
  for a four-point factorial design.}
\label{t:inferenceConditions2}
\end{table}

\begin{figure}[htb]
  \centering
  \includegraphics[width=0.75\textwidth]{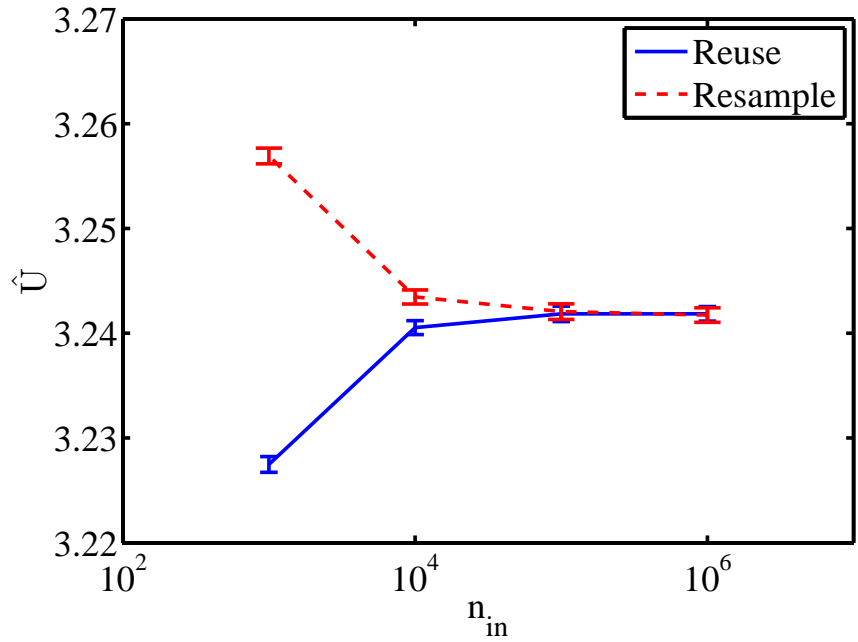}
  \caption{Expected utility estimates $\hU(d=0.2)$ for the simple
    nonlinear problem in
    Section~\ref{s:expectedUtilityNumericalMethods}, using different
    sample sizes $n_{\mathrm{in}}$. One estimator (blue solid line)
    reuses samples between the inner/outer Monte Carlo loops while the
    other (red dashed line) does not. Errors bars show the mean and
    standard deviation of the sampling distribution of each estimator.}
  \label{f:reuseBias}
\end{figure}

\cleardoublepage
\appendix
\section{Expected information gain from two experiments}
\label{app:infoGain}


Here we show that the expected information gain from two experiments
is not, in general, equal to the sum of the expected information gains
due to each experiment individually.

Consider two fixed experimental conditions $\bd_1$ and $\bd_2$, with
corresponding data $\by_1$ and $\by_2$. Following
Equation~(\ref{e:expectedUtility}), the expected information gain in the
model parameters $\btheta$ from performing both experiments is
\begin{eqnarray}
  U([\bd_1,\bd_2]) &=& \int_{\CY_1}\int_{\CY_2} \int_{\bTheta}
  p(\btheta|\by_1,\by_2,\bd_1,\bd_2)
  \ln\left[\frac{p(\btheta|\by_1,\by_2,\bd_1,\bd_2)}{p(\btheta)}\right]
  \,d\btheta \,p(\by_1,\by_2|\bd_1,\bd_2) \,d\by_2\,d\by_1 \nonumber\\ &=&
  \int_{\CY_1}\int_{\CY_2} \int_{\bTheta}
  \ln\left[\frac{p(\by_1,\by_2|\btheta,\bd_1,\bd_2)}{p(\by_1,\by_2|\bd_1,\bd_2)}
    \right]p(\by_1,\by_2|\btheta,\bd_1,\bd_2) p(\btheta)\,d\btheta
  \,d\by_2\,d\by_1 \nonumber\\ 
  &=& \int_{\CY_1}
  \int_{\bTheta} \ln\left[p(\by_1|\btheta,\bd_1)
    \right]p(\by_1|\btheta,\bd_1) p(\btheta) \,d\btheta
  \,d\by_1 \nonumber\\ && +\int_{\CY_2}
  \int_{\bTheta} \ln\left[p(\by_2|\btheta,\bd_2)
    \right]p(\by_2|\btheta,\bd_2) p(\btheta) \,d\btheta
  \,d\by_2  \nonumber\\ && -
  \int_{\CY_1}\int_{\CY_2} \int_{\bTheta} \ln\left[p(\by_1,\by_2|\bd_1,\bd_2)
    \right]p(\by_1,\by_2|\btheta,\bd_1,\bd_2) p(\btheta)
  \,d\btheta \,d\by_2\,d\by_1 \nonumber\\
  &=& \int_{\CY_1}
  \int_{\bTheta} \ln\left[p(\by_1|\btheta,\bd_1)
    \right]p(\by_1|\btheta,\bd_1) p(\btheta) \,d\btheta
  \,d\by_1 \nonumber\\ && +\int_{\CY_2}
  \int_{\bTheta} \ln\left[p(\by_2|\btheta,\bd_2)
    \right]p(\by_2|\btheta,\bd_2) p(\btheta) \,d\btheta
  \,d\by_2  \nonumber\\ && +
  h(\by_1,\by_2|\bd_1,\bd_2),
  \label{e:Uofd1d2}
\end{eqnarray}
where the second equality is due to application of Bayes' Theorem; the
third equality uses the fact that $p(\by_1,\by_2|\btheta,\bd_1,\bd_2)
= p(\by_1|\btheta,\bd_1)p(\by_2|\btheta,\bd_2)$, since the outputs are
conditionally independent given the designs \textit{and} parameters, and
data from each experiment depend only on its own design \textit{given
  the parameters}; and the last equality results from the marginalization
of $\btheta$ to obtain the differential entropy $h$.

Similarly, the sum of the expected information gain due to each
experiment individually is
\begin{eqnarray}
  U(\bd_1)+U(\bd_2) &=& \int_{\CY_1} \int_{\bTheta}
  p(\btheta|\by_1,\bd_1)
  \ln\left[\frac{p(\btheta|\by_1,\bd_1)}{p(\btheta)}\right] \,d\btheta
  \,p(\by_1|\bd_1) \,d\by_1 \nonumber\\ && + \int_{\CY_2}
  \int_{\bTheta} p(\btheta|\by_2,\bd_2)
  \ln\left[\frac{p(\btheta|\by_2,\bd_2)}{p(\btheta)}\right] \,d\btheta
  \,p(\by_2|\bd_2) \,d\by_2 \nonumber\\ 
  &=& \int_{\CY_1} \int_{\bTheta}
  \ln\left[\frac{p(\by_1|\btheta,\bd_1)}{p(\by_1|\bd_1)}
  \right]p(\by_1|\btheta,\bd_1) \, 
  p(\btheta)\,d\btheta \,d\by_1 \nonumber\\
  && + \int_{\CY_2} \int_{\bTheta}
  \ln\left[\frac{p(\by_2|\btheta,\bd_2)}{p(\by_2|\bd_2)}
  \right]p(\by_2|\btheta,\bd_2) \, 
  p(\btheta)\,d\btheta \,d\by_2 \nonumber\\
  &=& \int_{\CY_1} \int_{\bTheta}
  \ln\left[p(\by_1|\btheta,\bd_1) \right]p(\by_1|\btheta,\bd_1)
  p(\btheta) \,d\btheta \,d\by_1 \nonumber\\ && +\int_{\CY_2}
  \int_{\bTheta} \ln\left[p(\by_2|\btheta,\bd_2)
    \right]p(\by_2|\btheta,\bd_2) p(\btheta) \,d\btheta \,d\by_2
  \nonumber\\ && - \int_{\CY_1}\int_{\bTheta} \ln\left[p(\by_1|\bd_1)
    \right]p(\by_1|\btheta,\bd_1) p(\btheta) \,d\btheta \,d\by_1
  \nonumber\\ && - \int_{\CY_2}\int_{\bTheta} \ln\left[p(\by_2|\bd_2)
    \right]p(\by_2|\btheta,\bd_2) p(\btheta) \,d\btheta \,d\by_2
  \nonumber\\
  &=& \int_{\CY_1} \int_{\bTheta}
  \ln\left[p(\by_1|\btheta,\bd_1) \right]p(\by_1|\btheta,\bd_1)
  p(\btheta) \,d\btheta \,d\by_1 \nonumber\\ && +\int_{\CY_2}
  \int_{\bTheta} \ln\left[p(\by_2|\btheta,\bd_2)
    \right]p(\by_2|\btheta,\bd_2) p(\btheta) \,d\btheta \,d\by_2
  \nonumber\\ && + \, h(\by_1|\bd_1,\bd_2) + h(\by_2|\bd_1,\bd_2),
  \label{e:Uofd1Uofd2}
\end{eqnarray}
where we have also used the fact that $p(\by_1|\bd_1) =
p(\by_1|\bd_1,\bd_2)$ to arrive at the last equality.  Comparing
Equations~(\ref{e:Uofd1d2}) and~(\ref{e:Uofd1Uofd2}), the first two terms
are identical. The remaining terms follow the identity
\begin{eqnarray}
  h(\by_1,\by_2|\bd_1,\bd_2) \leq h(\by_1|\bd_1,\bd_2) +
  h(\by_2|\bd_1,\bd_2), 
\end{eqnarray}
and hence $U([\bd_1,\bd_2])\leq U(\bd_1)+U(\bd_2)$. Thus the total
expected information gain from two simultaneous experiments can never
be greater than the sum of the individual expected information
gains. Equality holds if and only if $\by_1$ and $\by_2$ are
conditionally independent given $\bd_1$ and $\bd_2$, or equivalently
$I(\by_1,\by_2|\bd_1,\bd_2)\equiv 0$. In other words, equality holds
when there is no mutual or ``overlapping'' information between the two
sets of data.

\section{Bias in the estimator of expected information gain}
\label{app:reuseBias}

The expected utility estimator described in
Section~\ref{s:expectedUtilityNumericalMethods}
(Equations~(\ref{e:expectedUtilityMC}) and~(\ref{e:evidenceMC})) is
biased. There are two sources of bias: (1) finite $n_{\mathrm{in}}$;
and (2) the reuse of prior samples between the outer and inner Monte
Carlo loops, in order to reduce the number of distinct evaluations of
the forward model $\bG\left ( \btheta, \bd \right )$. If we did not
reuse samples, the estimator employed in this paper would revert to
the estimator of~\cite{ryan:2003:eei}, which has bias proportional to
$n_{\mathrm{in}}^{-1}$. It is important to understand, then, how much
bias results from each source listed above and whether the reuse of
samples is worthwhile. We address these questions via a brief
numerical study.

Consider the simple nonlinear model described by
Equation~(\ref{e:simpleDesign}). The expected information gain for a
single experiment, with prior $\theta \sim \CU(0,1)$, is shown in
Figure~\ref{f:simpleDesign1Exp_theta0to1}. For simplicity, we now
consider numerical estimates of the expected information gain at
design $d=0.2$ only.
Figure~\ref{f:reuseBias} plots estimates of this value as a function
of $n_{\mathrm{in}}$ for two different estimators: one that reuses
samples between the outer and inner loops (in blue, marked `reuse'),
and another that does not reuse samples (in red, marked
`resample'). The value in the middle of each error bar is the mean of
many realizations of the corresponding estimator. The error bars
represent plus or minus one standard deviation of the sampling
distribution of each estimator,\footnote{Strictly speaking, in the
  estimator that reuses samples, we show the standard deviation of the
  estimator obtained by averaging $10^6/n_{\mathrm{out}}$ samples of
  the estimator $\hU(\bd)$ that itself uses $n_{\mathrm{out}} =
  n_{\mathrm{in}}$ samples.} for $n_{\mathrm{out}}$ fixed at $10^6$.


To assess the bias, we take the mean of the `resample' estimator with
$n_{\mathrm{in}}=10^6$ to be the ``true'' expected information gain;
certainly, since the bias of this estimator is inversely proportional
to $n_{\mathrm{in}}$, we expect it to be the least biased of all the
available numerical approximations. Fast convergence of the means with
respect to $n_{\mathrm{in}}$, as well as close agreement between the
`reuse' and `resample' cases at $n_{\mathrm{in}}=10^6$, further suggest
that bias is extremely small at this $n_{\mathrm{in}}$. We thus
evaluate the bias of the other estimators by taking the difference
between their means and this ``true'' value. As expected, the largest
bias occurs at $n_{\mathrm{in}}=10^3$, but it is only about $0.5\%$ of
the estimated value. Moreover, the biases of the `resample' and `reuse'
estimators are of the same order. Moving to the values of
$n_\mathrm{in}$ actually used in much of this paper ($10^4$ or
$10^5$), the bias falls by another order of magnitude or more.


These results suggest that sample reuse is a reasonable idea;
certainly it does not yield much more bias than the ordinary
estimator. But because it offers substantial gains in computational
efficiency via fewer evaluations of $\bG$, reuse can in turn allow
much larger values of $n_{\mathrm{in}}$ to be employed. Thus the
overall bias of the expected utility estimates can be dramatically
reduced, for the same computational effort.
Finally, we note that bias is a systematic error. The exact impact of
estimator bias on the results of stochastic optimization depends also on
how stationary the bias is with respect to $\bd$. As long as
bias-induced shifts do not compromise the \textit{relative} values of
the expected utility among different designs, the optimal design will not
change dramatically. Given the rather small levels of bias observed
here and the finite accuracy of any stochastic optimization
algorithm, we expect bias to have a small effect overall.

\section{Governing equations for homogeneous combustion}
\label{app:combustionGovEqns}

This appendix describes governing equations for the chemical system
analyzed in Section~\ref{s:applicationCombustionKinetics}. Consider
combustion in a spatially homogeneous (i.e., well-mixed) system at
constant pressure. Such ``zero-dimensional'' systems are frequently
used to model autoignition in shock tube
experiments~\cite{davidson:2004:ist}. Convective and diffusive
transport are neglected, leaving coupled ordinary differential
equations that represent conservation of individual species
(\ref{e:govEqnODE1}) and of energy (\ref{e:govEqnODE2}). The state of
the system is completely described by the species mass fractions
$Y_1, \ldots, Y_{n_s}$ (where $n_s$ is the total number of species)
and the temperature $T$. Governing equations are as follows:
\begin{eqnarray}
  \frac{dY_j}{dt} &=& \frac{\omegadot_j W_j}{\rho},\, \, j=1\ldots n_s \label{e:govEqnODE1} \\
  \frac{dT}{dt} &=& -\frac{1}{\rho c_p}\sum_{n=1}^{n_s} h_n
  \omegadot_n W_n \label{e:govEqnODE2}
\end{eqnarray}
with initial conditions
\begin{eqnarray}
      Y_j(t=0) = Y_{j,0}, \qquad T(t=0) = T_0 ,
  \label{e:govEqnIC}
\end{eqnarray}
where $\omegadot_j$ $\[\textrm{kmol}\cdot\textrm{m}^{-3}\cdot
\textrm{s}^{-1}\]$ is the molar production rate of the $j$th species,
$W_j$ $\[\textrm{kg}\cdot\textrm{kmol}^{-1}\]$ is the molecular weight
of the $j$th species, $\rho$ $\[\textrm{kg}\cdot\textrm{m}^{-3}\]$ is
the mixture density, $c_p$ $\[\textrm{J}\cdot
\textrm{K}^{-1}\cdot\textrm{kg}^{-1}\]$ is the mixture specific heat
capacity under constant pressure, and $h_n$
$\[\textrm{J}\cdot\textrm{kg}^{-1}\]$ is the specific enthalpy of the
$n$th species. The molar production rate is defined in terms of
elementary reaction rates as
\begin{eqnarray}
  \omegadot_j \equiv \frac{dC_j}{dt} = \sum_{m=1}^{n_r}(\nu_{mj}^{''}-\nu_{mj}^{'})
  \left(k_{f,m}\prod_{n=1}^{n_s}C_n^{\nu_{mn}^{'}}-
    k_{r,m}\prod_{n=1}^{n_s}C_n^{\nu_{mn}^{''}}\right), \label{e:omegadot}
\end{eqnarray}
where $C_j$ $\[\textrm{kmol}\cdot\textrm{m}^{-3}\]$ is the molar
concentration of the $j$th species, $n_r$ is the total number of
reactions, and $\nu_{mn}^{'}$ and $\nu_{mn}^{''}$ are the
(dimensionless) stoichiometric coefficients on the reactant and
product sides of the equation, respectively, for the $n$th species in
the $m$th reaction. Molar concentrations $C_j$ can be obtained from
mass fractions $Y_j$ as follows:
\begin{equation}
C_j = \rho \frac{Y_j}{W_j}.
\end{equation}

The forward and reverse rate constants of the $m$th reaction, denoted
by $k_{f,m}$ and $k_{r,m}$ respectively, are assumed to have the
modified Arrhenius form:
\begin{eqnarray}
  k_{f,m} &=&
  A_{m}T^{b_m}\exp\left(\frac{-E_{a,m}}{R_uT}\right) \\
  k_{r,m} &=& \frac{k_{f,m}}{K_{c,m}} =
  \frac{k_{f,m}}{\exp\left(\frac{-\Delta G^o_{T,m}}{R_uT}\right)},
\end{eqnarray}
where $A_m$
$\[\(\textrm{m}^3\cdot\textrm{kmol}^{-1}\)^{\(-1+\sum_{n=1}^{n_s}\nu_{mn}'\)}\cdot\textrm{s}^{-1}\cdot\textrm{K}^{-b_m}\]$
is the pre-exponential factor, $b_m$ is the exponent
of the temperature dependence, $E_{a,m}$
$\[\textrm{J}\cdot\textrm{kmol}^{-1}\]$ is the activation energy, $R_u
= 8314.472$
$\[\textrm{J}\cdot\textrm{kmol}^{-1}\cdot\textrm{K}^{-1}\]$ is the
universal gas constant, $K_{c,m}$
$\[\(\textrm{m}^3\cdot\textrm{kmol}^{-1}\)^{\(\sum_{n=1}^{n_s}\nu_{mn}'-\sum_{n=1}^{n_s}\nu_{mn}''\)}\]$
is the equilibrium constant, and $\Delta G^o_{T,m}$
$\[\textrm{J}\cdot\textrm{kmol}^{-1}\]$ is the change in Gibbs free
energy at standard pressure and temperature $T$. $A_m$, $b_m$, and
$E_{a,m}$ are collectively called the kinetic parameters of reaction
$m$.

The initial mass fractions $Y_{j,0}$ are expressed compactly using the
dimensionless equivalence ratio $\phi$:
\begin{eqnarray}
  \phi =
  \frac{\(Y_{O_2}/Y_{H_2}\)_{\textrm{stoic}}}{\(Y_{O_2}/Y_{H_2}\)} =
  \frac{\(X_{O_2}/X_{H_2}\)_{\textrm{stoic}}}{\(X_{O_2}/X_{H_2}\)}, 
\end{eqnarray}
where the subscript ``stoic'' refers to the stoichiometric ratios and
$X_j$ is the molar fraction of the $j$th species, related to the
mass fraction through
\begin{eqnarray}
  X_j = \frac{Y_j}{W_j \sum_{n=1}^{n_s}Y_n/W_n}.
\end{eqnarray}
In this paper, we use $X_j$ in place of the $Y_j$ as the species state
variables.
%
We assume a perfect gas mixture, thus closing the system with the
following equation of state:
\begin{eqnarray}
  \rho = \frac{p}{R_uT \sum_{n=1}^{n_s}Y_n/W_n},
\end{eqnarray}
where $p$ $\[\textrm{Pa}\]$ is the (assumed constant) pressure.

\end{document}